\documentclass[runningheads]{llncs}
\usepackage[T1]{fontenc}
\usepackage{graphicx,verbatim}
\usepackage{orcidlink}
\usepackage[dvipsnames]{xcolor}
\usepackage{booktabs}
\usepackage{algpseudocode}
\usepackage{algorithm}
\usepackage{multirow}
\usepackage{multicol}
\usepackage{mathtools}
\usepackage{xcolor}
\usepackage{ragged2e}
\usepackage{lmodern}
\usepackage[export]{adjustbox}
\usepackage{soul}
\usepackage{enumitem}
\usepackage[usestackEOL]{stackengine}
\usepackage[most]{tcolorbox}
\tcbuselibrary{listings}
\usepackage{amsfonts}

\newtcblisting{promptbox}[1][]{
  listing only,
  colback=blue!1,
  colframe=blue!60!black,
  title=#1,
  boxrule=0.8pt,
  arc=2pt,
  left=6pt,
  right=6pt,
  top=6pt,
  bottom=6pt,
  listing options={
    basicstyle=\ttfamily\small,
    breaklines=true,
    columns=fullflexible,
    keepspaces=true
  }
}

\usepackage[capitalize]{cleveref}
\crefname{section}{Sec.}{Secs.}
\Crefname{section}{Section}{Sections}
\Crefname{table}{Table}{Tables}
\crefname{table}{Tab.}{Tabs.}
\definecolor{cadmiumgreen}{rgb}{0.0, 0.42, 0.24}
\definecolor{custom}{cmyk}{0.1,0.48,0.49,0.2}
\definecolor{OliveGreen}{cmyk}{0.64,0,0.95,0.40}
\definecolor{new}{rgb}{0.81,0.05,0.9}
\definecolor{BrickRed}{rgb}{0.81,0.1,0.1}
\definecolor{RoyalBlue}{rgb}{0.2,0.2,0.75}

   \usepackage{xspace}
   \usepackage[normalem]{ulem}
   \usepackage{bm}
\makeatletter
\DeclareRobustCommand\onedot{\futurelet\@let@token\@onedot}
\def\@onedot{\ifx\@let@token.\else.\null\fi\xspace}

\makeatother

\def\clap#1{\hbox to 0pt{\hss #1\hss}}%
\def\initials#1{\protect\clap{\protect\smash{\protect\raisebox{1.4ex}{\protect\tiny{\protect\textsf{\protect\textit{#1}}}}}}}%
\makeatletter
\newcommand{\EDIT}[4][]{\protect\@ifundefined{hidecomments}{%
  \protect\strut{\color{#3}{\hspace{0pt}\initials{#2}\protect\sout{#1}{~#4}}}%
  }{#4}}
\newcommand{\NOTEboxed}[3]{\protect\@ifundefined{hidecomments}{%
  {\begin{center}\fbox{\parbox{0.97\linewidth}{\protect\EDIT{#1}{#2}{#3}}}\end{center}}
  }{}}
\newcommand{\COMM}[3]{\protect\@ifundefined{hidecomments}{%
  {\protect\EDIT{#1}{#2}{#3}}%
  }{}}
\newcommand{\DefAuthor}[2] % initials, color
{%
  \expandafter\newcommand\csname #1edit\endcsname[2][]{\protect\EDIT[##1]{#1}{#2}{##2}}
  \expandafter\newcommand\csname #1\endcsname[1]{\protect\COMM{#1}{#2}{[##1]}}
  \expandafter\newcommand\csname #1boxed\endcsname[1]{\protect\NOTEboxed{#1}{#2}{##1}}
}
\definecolor{dfltgreen}       {rgb}{0.0,0.5,0.0}
\definecolor{dfltred}         {rgb}{0.7,0.0,0.0}
\newcommand{\REVadd}[1]{\protect\@ifundefined{hidecomments}{%
  \strut{\color{dfltgreen}{#1}}}{#1}}
\newcommand{\REVedit}[2][]{\protect\@ifundefined{hidecomments}{%
  \strut{\color{dfltred}{\protect\sout{#1}}\color{dfltgreen}{~#2}}}%
  {#2}}
\makeatother

\definecolor{dkgreen}       {rgb}{0.0,0.5,0.0}
\definecolor{dkblue}        {rgb}{0.0,0.0,0.7}
\definecolor{dkcyan}        {rgb}{0.0,0.5,0.5}
\definecolor{dkmagenta}     {rgb}{0.5,0.0,0.5}
\DefAuthor{MK}{dkmagenta} % MARGRET'S COMMENTS
\DefAuthor{BS}{dkgreen} % Bernt's COMMENTS
\DefAuthor{SA}{dkblue} % Shreyash'S COMMENTS
\DefAuthor{Marcel}{dkcyan} % Marcel's COMMENTS
\DefAuthor{AG}{orange} % SHASHANK'S COMMENTS
\DefAuthor{todo}{red}

\usepackage{pifont}% http://ctan.org/pkg/pifont
\usepackage[english]{babel}
\usepackage[square,numbers]{natbib}
\usepackage{etoolbox}
\usepackage{hyperref}
\begin{document}
\title{Faithful, Interpretable Chest X-ray Diagnosis with Artifact-free B-cos Networks}
\titlerunning{Diagnosis with Artifact-free B-cos Networks}
\begin{comment}  %% Removed for anonymized MICCAI submission
\author{Anonymized Authors}  %% Added for anonymized MICCAI submission
\authorrunning{Anonymized Author et al.}
\institute{Anonymized Affiliations \\
    \email{email@anonymized.com}}

\end{comment}

\newcommand{\equalcontrib}{\textsuperscript{*}}

%\iffalse
\author{Shreyash Arya\inst{1}\equalcontrib\orcidlink{0000-0002-2108-4579} \and
Shashank Agnihotri\inst{2}\equalcontrib\orcidlink{0000-0001-6097-8551} \and
Marcel Kleinmann\inst{2}\orcidlink{0009-0001-6770-231X} \and \\
Bernt Schiele\inst{1}\orcidlink{0000-0001-9683-5237} \and
Margret Keuper\inst{2,1}\orcidlink{0000-0002-8437-7993}
}
\authorrunning{S. Arya et al.}
\institute{Max-Planck-Institute for Informatics, Saarland Informatics Campus, Germany
\email{\{sarya,schiele\}@mpi-inf.mpg.de}\\
 \and
Machine Learning Group, University of Mannheim, Mannheim, Germany\\
\email{\{shashank.agnihotri,keuper\}@uni-mannheim.de}}
%\fi
\iffalse
\author{Shreyash Arya\inst{1}\equalcontrib\orcidID{0000-0002-2108-4579} \and
Shashank Agnihotri\inst{2}\equalcontrib\orcidID{0000-0001-6097-8551} \and
Marcel Kleinmann\inst{2}\orcidID{0009-0001-6770-231X} \and \\
Bernt Schiele\inst{1}\orcidID{0000-0001-9683-5237} \and
Margret Keuper\inst{2,1}\orcidID{0000-0002-8437-7993}
}
%index{Arya, Shreyash}
%index{Agnihotri, Shashank}
%index{Kleinmann, Marcel}
%index{Schiele, Bernt}
%index{Keuper, Margret}
%
\authorrunning{S. Arya and S. Agnihotri et al.}
%
\institute{Max-Planck-Institute for Informatics, Saarland Informatics Campus, Germany
\email{\{sarya,schiele\}@mpi-inf.mpg.de}\\
 \and
Machine Learning Group, University of Mannheim, Mannheim, Germany\\
\email{\{shashank.agnihotri,keuper\}@uni-mannheim.de}}
\fi

\maketitle              % typeset the header of the contribution

\begingroup
\renewcommand\thefootnote{*}
\phantomsection
\footnotetext{Equal contribution.}
\endgroup

\begin{abstract}
Faithfulness and interpretability are essential for deploying deep neural networks (DNNs) in safety-critical domains such as medical image analysis. B-cos networks modify the parameterization of convolutional and classification layers to measure class evidence via feature-weight alignment, enabling built-in, class-specific contribution maps without post-hoc explanations. While maintaining diagnostic performance competitive with state-of-the-art DNNs, standard B-cos networks exhibit severe aliasing artifacts in their explanation maps, rendering them unsuitable for clinical use, where clarity is essential. 
In this work, we address this limitation by introducing anti-aliasing strategies using $\mathrm{ASAP}$ and $\mathrm{BlurPool}$ (BP) to significantly improve explanation quality. Our experiments on chest X-ray datasets demonstrate that the modified $\text{B-cos}_\mathrm{ASAP}$ and $\text{B-cos}_\mathrm{BP}$ preserve strong predictive performance while providing faithful and artifact-free explanations suitable for clinical application in multi-class and multi-label settings. Code is available at: \url{https://github.com/shrebox/Artifact-free-B-cos-Networks}.

% Authors must provide keywords and are not allowed to remove this Keyword section.

\end{abstract}

\begin{figure}

    \centering
    \includegraphics[width=1.0\linewidth]{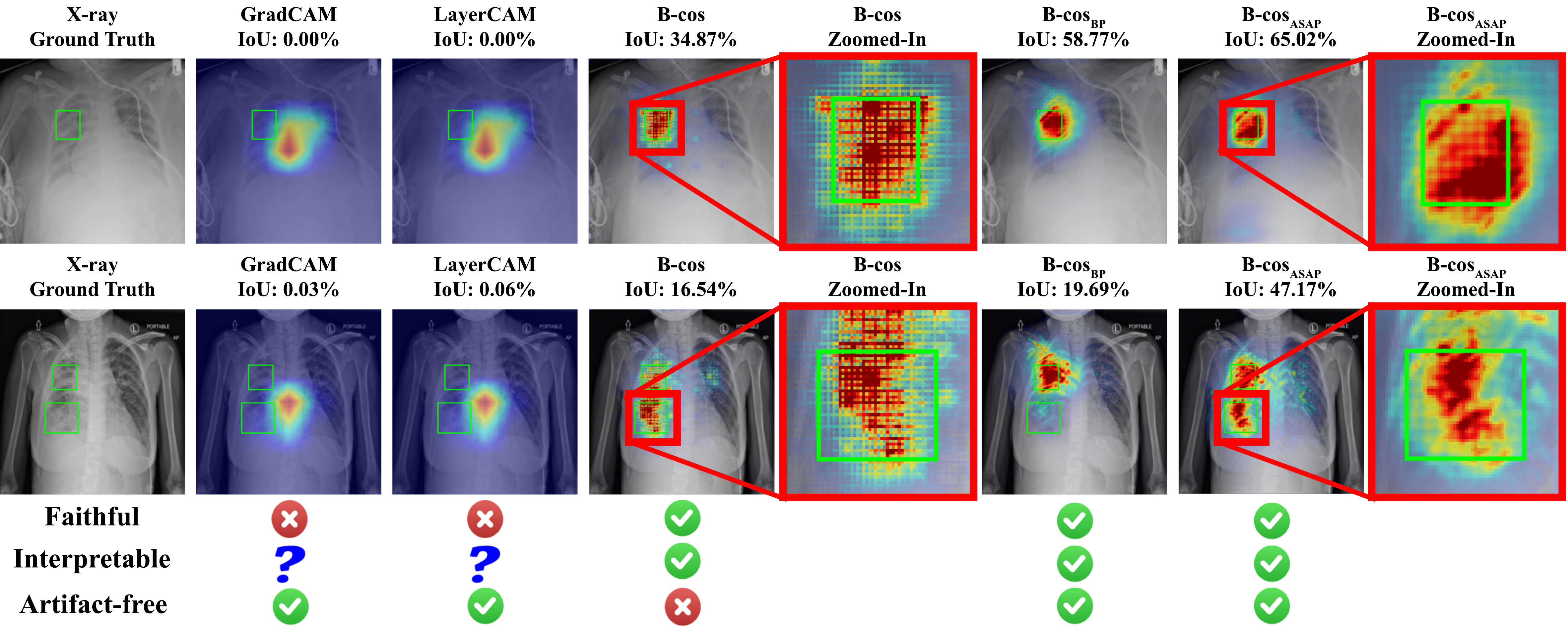}
   
    \caption{Comparison of standard B-cos with our anti-aliased variants, $\text{B-cos}_\mathrm{ASAP}$ and $\text{B-cos}_\mathrm{BP}$, against post-hoc methods (GradCAM~\cite{selvaraju_grad-cam_2020}, LayerCAM~\cite{jiang_layercam_2021}) on the Baseline (ResNet50). While B-cos explanations localize better than post-hoc maps, they exhibit severe grid artifacts that hinder clinical use; $\text{B-cos}_\mathrm{ASAP}$ and $\text{B-cos}_\mathrm{BP}$ remove these artifacts and improve localization (IoU with ground-truth boxes). Examples are from RSNA Pneumonia Detection; all methods correctly predict ``pneumonia''.}
    \label{fig:teaser}

\end{figure}
\section{Introduction}
\label{sec:intro}
Faithfulness and interpretability are critical prerequisites for deploying deep learning models in safety-critical domains such as healthcare, where decisions have direct implications for patient outcomes~\cite{current_challenges_2022, dlxraysurvey}. In particular, clinical adoption demands models whose reasoning processes can be understood and verified by medical professionals. However, most existing approaches in medical image processing rely on architectures that lack inherent interpretability, offering limited insight into the basis of their predictions~\cite{oakden-rayner_exploring_2019}. As emphasized in~\cite{dlxraysurvey}, knowing how a diagnosis was derived is highly valuable, especially in practice, where AI tools are used to support, not replace, radiologists. In this work, we explore B-cos networks~\cite{boehle_b-cos_2022}, which are inherently interpretable by design and generate class-specific contribution maps that visually indicate the evidence behind each prediction. Representative examples are shown in \cref{fig:teaser}, which compares our artifact-free, faithful, and interpretable explanations to other common approaches such as GradCAM and LayerCAM while displaying the need for our extensions when working with B-cos networks and chest X-rays simultaneously.

These models can directly show which part of the image caused network activations, enabling an understanding of why a classification was made~\cite{boehle_b-cos_2022}. Unlike post-hoc approaches such as GradCAM~\cite{selvaraju_grad-cam_2020} or LayerCAM~\cite{jiang_layercam_2021}, which provide coarse and sometimes misleading heatmaps that have insufﬁcient mechanistic explanation of the decision-making process~\cite{ghosh_bridging_2023, rudin2019stop_explaining}, B-cos networks offer inherently interpretable, class-specific contribution maps that yield clearer, more faithful visual explanations. 
This can help radiologists verify predictions, improve clinical workflows, and reduce diagnostic errors~\cite{sun2024attri_bcos_chest_xray,bcos_medical}. 
However, standard B-cos networks face a critical limitation: the generated explanation maps often exhibit grid artifacts, making them visually unreliable for clinical use.
%; second, the original formulation supports only multi-class classification, whereas chest X-rays commonly involve multiple co-occurring conditions.
In this work, we address these limitations by incorporating anti-aliasing techniques such as $\mathrm{ASAP}$ and $\mathrm{BlurPool}$ to improve visual clarity. %, and by extending the B-cos framework to support multi-label classification. 
Additionally, while the original B-cos networks support multi-label classification, they were proposed for multi-class classification.
Thus, we use the framework from \cite{rao2023studying_bcos_multilabel}, to obtain multi-label classification since chest X-rays commonly involve multiple co-occurring conditions~\cite{nguyen2020vindrcxr}.
We refer to the resulting aliasing-free model as $\text{B-cos}_\mathrm{ASAP}$ and $\text{B-cos}_\mathrm{BP}$.
While $\text{B-cos}_\mathrm{ASAP}$ combines B-cos networks with $\mathrm{ASAP}$~\cite{asap_2023}, $\text{B-cos}_\mathrm{BP}$ combines it with $\mathrm{BlurPool}$~\cite{blurpool}, both effective anti-aliasing techniques.
These anti-aliasing techniques help improve the explanation maps from the B-cos network, since they replace the artifact-producing downsampling operations with an artifact-free downsampling path.
Thus, removing the artifacts in the explanation maps of B-Cos networks.
Our main contributions are:

\begin{itemize}
\item Application-specific modifications to B-cos networks by integrating anti-aliasing methods ($\mathrm{ASAP}$ and $\mathrm{BlurPool}$) to produce spectral artifact-free explanations suitable for diagnostic use.
\item A practical evaluation of B-cos networks for interpretable and clinically relevant chest X-ray disease detection, showing that interpretability does not impede performance.
\item We show that our proposed framework can be adopted for both multi-class and multi-label classification, proving useful in critical medical applications.
%\item An extension of the B-cos framework to multi-label classification, addressing the clinical need to detect multiple co-existing conditions in a single chest radiograph.
\end{itemize}

\section{Related Work}
\label{sec:related}
We briefly review (i) clinically relevant evaluation criteria, (ii) deep learning for chest X-ray diagnosis and the datasets used in this work, and (iii) explainability and interpretability requirements for medical AI.

\textbf{Clinical Evaluation Metrics. }
While many medical imaging works report standard machine learning metrics~\cite{keuper2013blind}, clinical studies primarily assess diagnostic tools using sensitivity and specificity~\cite{pneumonia_ensemble_reference}. Sensitivity (recall) measures the fraction of true positives detected and is critical for early disease detection, because missed diagnoses can severely impact outcomes~\cite{juan_computer-assisted_2023}, with high reported mortality for childhood pneumonia~\cite{children_pneumonia_2016}. However, sensitivity often trades off against specificity, and low specificity can lead to unnecessary follow-up and treatment~\cite{children_pneumonia_2016}. In this work, we therefore report recall prominently while also tracking specificity-relevant behavior through complementary performance metrics, since reducing false positives is important for clinical trust.

\textbf{Deep learning for chest X-ray diagnosis. }
Automated abnormality detection in chest X-rays predates deep learning, with early rule-based and statistical systems reported already in the 1960s~\cite{dlxraysurvey}. Modern chest X-ray analysis is dominated by deep neural networks, enabled by large-scale datasets and architectural advances, and can approach expert-level performance in several settings~\cite{litjens_survey_2017,dlxraysurvey}. CNN backbones such as ResNet~\cite{resnet_original} are widely used on large datasets like CheXpert~\cite{irvin_chexpert_2019} and NIH ChestX-ray14~\cite{wang_chestx-ray8_2017}, with strong reported performance in chest X-ray classification~\cite{resnet50_medical}. DenseNet-based CheXNet~\cite{chexnet,densenet_original} is a prominent example for pneumonia detection.
%Efficient architectures such as MobileNet~\cite{howard_mobilenets_2017} and EfficientNet~\cite{efficientnet_lung,mobilenet_pneumonia} have also been explored for practical deployment. 
More recently, vision transformers have been applied to pneumonia detection~\cite{vit_pneumonia}, including medical variants such as IEViT~\cite{ievit_medical}. 
Most importantly for this paper, B-cos aligned transformers have been demonstrated in the medical domain, supporting the relevance of B-cos style inherent explanations beyond natural images~\cite{bcos_medical}.

\textbf{Datasets with localization supervision. }
Large-scale chest X-ray benchmarks such as NIH ChestX-ray14~\cite{wang_chestx-ray8_2017}, CheXpert~\cite{irvin_chexpert_2019}, and MIMIC-CXR~\cite{johnson_mimic-cxr_2019}
%, and PadChest~\cite{padchest} 
have driven progress in classification, but most do not provide region-level supervision for rigorous quantitative evaluation of explanation localization. Since our goal is to assess whether attribution maps align with pathology regions, we focus on datasets with bounding boxes. For binary pneumonia detection, we use the RSNA Pneumonia Detection Challenge dataset~\cite{rsna_pneumonia_dataset}, which provides image-level labels and bounding boxes. For multi-label diagnosis, we use the VinBigData Chest X-ray Abnormalities Detection dataset (VinDr-CXR based)~\cite{nguyen2020vindrcxr}, which offers bounding boxes for 14 thoracic findings with radiologist annotation, enabling evaluation under realistic co-occurring conditions.

\textbf{Explainability and inherent interpretability. }
Explainability aims to make model decisions understandable~\cite{gilpin_explaining_2019,alshami2025aim}, and is particularly important in medical imaging where clinicians must be able to verify and trust model outputs~\cite{chen_humancentered_2022}. Regulatory discussions further motivate decision traceability in medical AI, including references to European data protection frameworks such as GDPR~\cite{singh_explainable_2020}. A central distinction is between post-hoc explanations and inherently interpretable models~\cite{barredo_posthoc_2020}. Post-hoc heatmaps can be coarse and may not faithfully reflect the model decision process, which limits their reliability in high-stakes settings~\cite{ghosh_bridging_2023}. % removed citation turbe_evaluation_2023
This concern is echoed by arguments that high-stakes applications should prefer interpretable models over post-hoc explanations~\cite{rudin2019stop_explaining}. In this work, we follow this paradigm and evaluate B-cos networks~\cite{boehle_b-cos_2022,arya2024bcosification} for chest X-ray diagnosis, focusing on making their built-in contribution maps clinically usable by removing aliasing artifacts while preserving predictive performance.

\section{Methodology}
\label{sec:method}
\begin{figure}[t]
    \centering
    \includegraphics[width=1.0\linewidth]{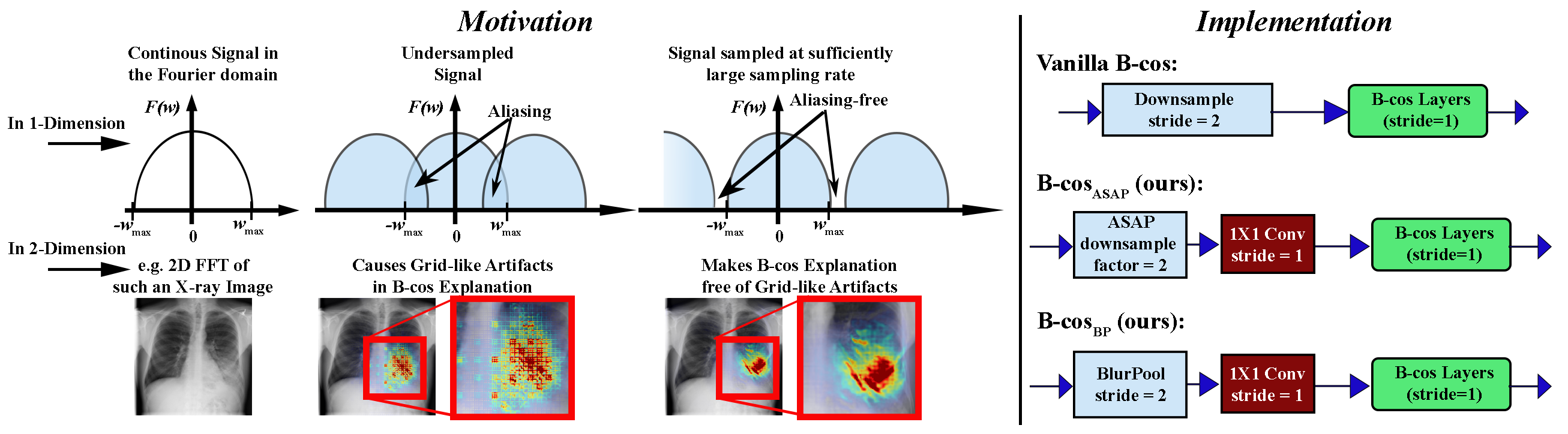}

    \caption{Method overview: Grid artifacts in B-cos explanations arise from aliasing while downsampling (undersampling). We mitigate them by replacing stride=2 convolutions with anti-aliasing operations like $\mathrm{ASAP}$ or $\mathrm{BlurPool}$.}
    \label{fig:method}
\end{figure}
%Following, we describe the proposed methodology to make B-cos explanations artifact-free. 

\subsection{Why do B-cos explanations have Grid-like Artifacts?}

It is essential to understand why strided downsampling introduces grid-like artifacts in B-cos explanations, and why $\mathrm{ASAP}$ mitigates them.
A B-cos neuron admits the scaled linear form,
\begin{equation}
\mathrm{B\text{-}cos}(x;w)\;=\;\hat w^{\top}x\cdot \bigl|c(x,\hat w)\bigr|^{\,B-1},
\qquad 
\hat w=\frac{w}{\|w\|},\;\;
c(x,\hat w)=\frac{\hat w^{\top}x}{\|x\|},
\label{eq:bcos_scaled_linear}
\end{equation}
which makes explicit that responses depend on both the linear projection $\hat w^\top x$ and an input-dependent gain $|c(x,\hat w)|^{B-1}$.
Crucially, B-cos explanations are defined via the induced linear map $W_{1\rightarrow l}(x)$ and the input contribution scores
\begin{equation}
s^{l}_{j}(x) \;=\; \bigl[W_{1\rightarrow l}(x)\bigr]^{\top}_{j}\ \odot\ x,
\label{eq:bcos_contrib}
\end{equation}
with pixelwise scores obtained by summing over channels at each spatial location.
Therefore, any structured distortion that propagates into $W_{1\rightarrow l}(x)$ appears directly and faithfully in the contribution map $s^{l}_{j}(x)$, as seen in \cref{fig:teaser}.
A key source of such distortions is improper sampling at resolution changes.
%Let $x[\mathbf{n}]$ be a 2D feature map, $h[\mathbf{n}]$ a kernel, and $v[\mathbf{n}]=(x*h)[\mathbf{n}]$.
Let $x$ be a 2D feature map, $h$ a kernel and $v=x*h$. 
A stride-$d$ convolution equals a convolution with a subsequent downsampling where every $d$th element is sampled
\begin{equation}
v_\downarrow[m,n] \;=\; (x*h)[m,n]\cdot \sum_{k,\ell} \delta[m-dk, n-d\ell] = v[dm,dn],
\label{eq:stride_decimate}
\end{equation}
where $\sum_{k,\ell} \delta[m-dk, n-d\ell]$ is a 2D dirac impulse comb, i.e. the sampling pattern.
%\begin{equation}
%v^\text{down}[m,n] \;=\; (x*h)[d\cdot n,d \cdot m],
%\label{eq:stride_decimate}
%\end{equation}
%where

\subsection{Anti-aliasing Using Nyquist Rate}
In the model forward pass, undersampling when downsampling can lead to aliasing, as shown by \cref{fig:method}~\cite{asap_2023,agnihotri2024beware,agnihotri2024improving}. 
Specifically, the pointwise multiplication in the spatial domain (c.f. Eq.~\eqref{eq:stride_decimate}) corresponds to a convolution in the Fourier domain, i.e., after applying the 2D DFT according to 
\begin{equation}
\small
V[\omega_x,\omega_y]=\frac{1}{MN}\sum_{m=0}^{M-1}\sum_{n=0}^{N-1} v[m,n]\,
e^{-j\left(\frac{\omega_x}{M}m+\frac{\omega_y}{N}n\right)},
\label{eq:asap_dft}
\end{equation}
for $v$ of size $M\times N$, the downsampling is given through~\cite{nyquist}
\begin{equation}\small
V_\downarrow[\omega_x,\omega_y]
\;=\;
V*\frac{1}{d^{2}}
\sum_{k,\ell}\delta\left[\omega_x - \frac{2\pi k}{d},\omega_y -\frac{2\pi\ell}{d}\right]=\frac{1}{d^{2}}
\sum_{k,\ell} V\left[\omega_x - \frac{2\pi k}{d},\omega_y- \frac{2\pi\ell}{d}\right].
%V\!\left(\frac{\boldsymbol{\omega}+2\pi\mathbf{k}}{s}\right).
\label{eq:alias_sum}
\end{equation}

%Specifically, the 
%With 2D DTFTs $V(\boldsymbol{\omega})$ and $Y(\boldsymbol{\omega})$, decimation by $s$ yields the alias-sum identity
%\begin{equation}
%Y(\boldsymbol{\omega})
%\;=\;
%\frac{1}{s^{2}}
%\sum_{\mathbf{k}\in\{0,\dots,s-1\}^{2}}
%V\!\left(\frac{\boldsymbol{\omega}+2\pi\mathbf{k}}{s}\right).
%\label{eq:alias_sum}
%\end{equation}
The $k,\ell\neq \mathbf{0}$ terms are alias replica that fold energy outside the reconstructable region %downsampled region $[-\pi/s,\pi/s)^2$ 
into the  baseband whenever $V[\omega_x,\omega_y]$ is not sufficiently low-pass, i.e. is sampled below the Nyquist rate.
Equivalently, alias-free, i.e. distortion-free downsampling requires an anti-aliasing pre-filter such that
\begin{equation}
V[\omega_x,\omega_y] \approx 0
\quad \text{for} \quad
\omega_x,\omega_y\notin[-\pi/d,\pi/d)^2.
\label{eq:nyquist_condition}
\end{equation}
When this condition is violated, periodic grid-like structure is injected into intermediate activations; in B-cos, such structure affects both the projection $\hat w^\top x$ and the gain term $|c(x,\hat w)|^{B-1}$ in \cref{eq:bcos_scaled_linear}, thereby modulating the induced map $W_{1\rightarrow l}(x)$ and appearing as spectral/grid artifacts in $s^l_j(x)$ via \cref{eq:bcos_contrib}.

$\mathrm{ASAP}$ replaces aliasing-prone downsampling by explicit bandlimited Fourier-domain downsampling.
%Given $f(m,n)\in \mathbb{R}^{M\times N}$, 
To this end, $\mathrm{ASAP}$ computes the 2D DFT
%\begin{equation}
%F(k,l)=\frac{1}{MN}\sum_{m=0}^{M-1}\sum_{n=0}^{N-1} f(m,n)\,
%e^{-2\pi j\left(\frac{k}{M}m+\frac{l}{N}n\right)}.
%\label{eq:asap_dft}
%\end{equation}
to downsample by factor $d$, $\mathrm{ASAP}$ enforces the new Nyquist limit by cropping (zeroing) coefficients outside the admissible band $|\omega_x|,|\omega_y|\le \tfrac{\pi}{d}$ (thus $\tfrac{\pi}{2}$ for $d=2$), which suppresses the alias-replica contributions in \cref{eq:alias_sum}.
To reduce ringing from a hard spectral cut, $\mathrm{ASAP}$ applies a Hamming window in the Fourier domain (2D via outer product) and uses stabilization variants (transpose and/or asymmetric padding for even sizes), before inverse transforming to obtain the $\tfrac{M}{d}\times\tfrac{N}{d}$ output.
Such a low-pass filter in the frequency domain corresponds to blurring in the spatial domain.
Thus, we also implement $\mathrm{BlurPool}$~\cite{blurpool}, which blurs in the spatial domain while downsampling.

\subsection{Making B-cos Aliasing-Free}
In our B-cos backbones, we therefore replace each stride=2 downsampling convolution with anti-aliasing operations while downsampling with $d$=2, followed by a stride=1 convolution:
\begin{equation}
\mathrm{Conv}(\text{stride}=2)
\;\Longrightarrow\;
\mathrm{ASAP}_{d=2}\circ \mathrm{Conv}(\text{stride}=1).
\label{eq:replace_stride2_asap}
\end{equation}
\iffalse
or, 
\begin{equation}
\mathrm{Conv}(\text{stride}=2)
\;\Longrightarrow\;
\mathrm{BlurPool}_{d=2}\circ \mathrm{Conv}(\text{stride}=1),
\label{eq:replace_stride2_blurpooling}
\end{equation}
\fi
We additionally use $\mathrm{BlurPool}_{d=2}$, which executes a spatial blurring operation before downsampling, as an anti-aliasing operation in place of $\mathrm{ASAP}_{d=2}$.
Both mitigate aliasing source at resolution transitions, yielding markedly cleaner, grid-like artifacts-free B-cos contribution maps as shown by \cref{fig:method}.

\section{Experiments}
\label{sec:experiments}
We run all experiments with three fixed random seeds. We evaluate on two chest X-ray datasets with bounding box annotations: RSNA Pneumonia Detection (binary: pneumonia vs.\ no pneumonia)~\cite{rsna_pneumonia_dataset} and VinBigData (multi-label classification of 14 thoracic findings)~\cite{nguyen2020vindrcxr}. Bounding boxes enable quantitative evaluation of explanation localization in addition to classification performance.

Following established protocols~\cite{pneumonia_ensemble_reference}, we use 5-fold cross-validation~\cite{kohavi_study_1995} with a 20\% held-out test split to capture clinical variability in pneumonia presentation~\cite{chestxrayinterpretation2014}. All baselines and B-cos variants are initialized from ImageNet-pretrained weights and fine-tuned on the target datasets. As commonly used architectures, we study ResNet-50~\cite{resnet_original}, ConvNeXt-tiny, and ConvNeXt-base~\cite{convnext}.

\textbf{Explanation metrics. }
We use Energy-based Pointing Game (EPG) scores inspired by ScoreCAM~\cite{wang_score-cam_2020}. Let $L^{c}(i,j)$ denote the attribution at pixel $(i,j)$ for class (label) $c$, with positive and negative contributions $L^{c}_{p}=\max(L^{c},0)$ and $L^{c}_{n}=\min(L^{c},0)$. For a ground-truth bounding box set $\text{bbox}$,
\begin{equation}
\label{eq:epg_base_compact}
\mathrm{EPG}(A,B)=\frac{\sum_{(i,j)\in\text{bbox}} A(i,j)}{\sum_{(i,j)\in\text{bbox}} A(i,j)+\sum_{(i,j)\notin\text{bbox}} B(i,j)}.
\end{equation}
Standard EPG is $\mathrm{EPG}(L^{c},L^{c})$. We report $\mathrm{EPG}_{\mathrm{Precision}}=\mathrm{EPG}(L^{c}_{p},L^{c}_{p})$, which focuses on how much positive evidence is attributed to within $\text{bbox}$ (only B-cos explanations allow analysis with this metric). In contrast, $\mathrm{EPG}_{\mathrm{Recall}}=\mathrm{EPG}(L^{c}_{p},|L^{c}_{n}|)$ measures negative evidence inside the box. We average EPG over classes; for VinBigData, we additionally report $\mathrm{EPG}^{\mathrm{micro}}$ (mean is weighted by per-label sample counts), mAP, Recall, F1, and AUROC.

\subsection{Quantitative Results}
We observe in \cref{fig:teaser} that modifying the strided convolution operations in B-cos with $\mathrm{ASAP}$ or $\mathrm{BlurPool}$ provides significantly better and aliasing-free explanations than the inherent B-cos explanations.
We also observe that these explanations are significantly better localized than the post-hoc explanations like GradCAM and LayerCAM on the Baseline architectures. 
It is paramount that interpretability does not come at the cost of performance.
Thus, in \cref{tab:RSNA_Pneumonia_Results} and \cref{tab:VinBigXray_Results}, we report the empirical performance over 5-fold and 3 random seeds on the RSNA Pneumonia Detection Challenge Dataset~\cite{rsna_pneumonia_dataset} and the VinBigXray Dataset~\cite{nguyen2020vindrcxr} using prominent architectures like ResNet50, ConvNeXt-tiny, and ConvNeXt-base, trained on the respective datasets.
We observe that the performance of B-cos, $\text{B-cos}_\mathrm{ASAP}$, and $\text{B-cos}_\mathrm{BP}$ is comparable to the baseline architectures.
This is reflected in the significantly higher than baseline Recall for the B-cos variants on the RSNA Pneumonia Detection Challenge Dataset.
This is desirable as a higher recall is often warranted for medical diagnosis.

\begin{table}[t]
    \centering
    
    \caption{Empirical Results on the RSNA Pneumonia Detection Challenge Dataset using ResNet50 and ConvNeXt-base.}
    %\resizebox{0.9\linewidth}{!}{
    \resizebox{1.0\linewidth}{!}{
    \scriptsize
    \setlength{\tabcolsep}{3pt}
    \begin{tabular}{llccccccc}
     \toprule
       Architecture & Model & Accuracy (\%) & AUROC & F1 Score & $\text{EPG}_\text{Recall}$ & $\text{EPG}_\text{Precision}$ & Recall & mIoU  \\
       \midrule
       \multirow{4}{*}{ResNet50} & Baseline & $83.5\pm0.4$ & $83.3\pm0.6$ & $58.0\pm1.3$ & -- & -- & $50.6\pm1.7$ & --  \\
       & $\text{B-cos}$ & $80.5\pm0.8$ & $86.7\pm0.7$ & $63.3\pm1.1$ & $100.0\pm0.0$ & $24.1\pm0.8$ & $74.8\pm2.2$ & $4.7\pm0.5$  \\
       & $\text{B-cos}_\text{BP}$ & $79.2\pm0.9$ & $87.1\pm0.7$ & $63.2\pm0.9$ & $98.7\pm1.4$ & $20.1\pm1.9$ & $79.2\pm1.7$ & $10.6\pm2.0$  \\
       & $\text{B-cos}_\text{ASAP}$ & $79.9\pm0.6$ & $87.1\pm0.5$ & $63.5\pm0.6$ & $98.6\pm0.9$ & $21.1\pm1.3$ & $77.5\pm1.1$ & $9.6\pm1.8$  \\
       \midrule
       \multirow{4}{*}{ConvNeXt-base} & Baseline & $81.0\pm0.5$ & $81.7\pm0.6$ & $51.5\pm1.7$ & -- & -- & $44.8\pm2.3$ & --  \\
       & $\text{B-cos}$ & $79.4\pm0.6$ & $88.2\pm0.5$ & $63.6\pm0.9$ & $97.8\pm0.6$ & $18.6\pm1.1$ & $79.9\pm1.6$ & $1.4\pm0.5$  \\
       & $\text{B-cos}_\text{BP}$ & $79.1\pm0.5$ & $88.1\pm0.4$ & $63.5\pm0.7$ & $93.9\pm1.5$ & $20.1\pm1.6$ & $80.7\pm1.4$ & $2.1\pm0.6$  \\
       & $\text{B-cos}_\text{ASAP}$ & $79.3\pm1.0$ & $88.0\pm0.4$ & $63.5\pm0.8$ & $94.4\pm1.7$ & $21.9\pm2.8$ & $80.2\pm2.3$ & $2.0\pm0.3$  \\
     \bottomrule
    \end{tabular}
    }
    \label{tab:RSNA_Pneumonia_Results}
\end{table}

%====================================================================================================
%VinBigXray LaTeX table
%====================================================================================================
\begin{table}[t]
    \centering
    
    \caption{Empirical Results on the VinBig Dataset using ResNet50, ConvNeXt-tiny, and ConvNeXt-base as the backbones.}
    \resizebox{\linewidth}{!}{
    \scriptsize
    \setlength{\tabcolsep}{3pt}

    \begin{tabular}{llccccccc}
     \toprule
       Architecture & Model & mAP & AUROC & F1 Score & $\text{EPG}_\text{Recall}^\text{micro}$ & $\text{EPG}_\text{Precision}^\text{micro}$ & Recall & mIoU  \\
       \midrule
       \multirow{4}{*}{ResNet50} & Baseline & $50.6\pm1.4$ & $80.8\pm1.1$ & $46.2\pm1.9$ & -- & -- & $38.6\pm1.8$ & --  \\
       & $\text{B-cos}$ & $53.5\pm1.4$ & $91.9\pm0.6$ & $44.5\pm1.5$ & $97.7\pm0.4$ & $9.4\pm0.2$ & $38.6\pm1.5$ & $2.3\pm0.2$  \\
       & $\text{B-cos}_\text{BP}$ & $47.8\pm1.3$ & $87.7\pm0.9$ & $42.3\pm1.6$ & $92.3\pm1.2$ & $9.0\pm0.2$ & $37.3\pm1.6$ & $5.4\pm0.7$  \\
       & $\text{B-cos}_\text{ASAP}$ & $47.6\pm1.3$ & $87.3\pm0.8$ & $42.2\pm1.2$ & $92.4\pm1.3$ & $9.9\pm0.4$ & $37.0\pm1.1$ & $5.2\pm0.9$  \\
       \midrule
       \multirow{4}{*}{ConvNeXt-tiny} & Baseline & $52.0\pm1.0$ & $90.2\pm0.7$ & $45.1\pm1.1$ & -- & -- & $39.4\pm1.0$ & --  \\
       & $\text{B-cos}$ & $57.9\pm1.5$ & $90.1\pm1.1$ & $53.4\pm1.2$ & $94.9\pm0.6$ & $12.3\pm0.4$ & $47.6\pm1.3$ & $2.3\pm0.3$  \\
       & $\text{B-cos}_\text{BP}$ & $55.9\pm1.6$ & $90.0\pm0.7$ & $49.4\pm1.3$ & $93.7\pm0.5$ & $14.6\pm0.4$ & $42.1\pm1.3$ & $4.6\pm0.4$  \\
       & $\text{B-cos}_\text{ASAP}$ & $50.6\pm1.4$ & $90.9\pm0.6$ & $40.0\pm1.8$ & $93.6\pm1.1$ & $11.4\pm1.0$ & $33.4\pm1.7$ & $2.3\pm0.5$  \\
       \midrule
       \multirow{4}{*}{ConvNeXt-base} & Baseline & $51.6\pm0.9$ & $89.7\pm0.6$ & $44.7\pm0.8$ & -- & -- & $39.0\pm0.9$ & --  \\
       & $\text{B-cos}$ & $58.4\pm1.5$ & $92.2\pm0.5$ & $51.3\pm1.4$ & $95.4\pm0.6$ & $13.2\pm0.4$ & $44.8\pm1.4$ & $1.6\pm0.1$  \\
       & $\text{B-cos}_\text{BP}$ & $54.9\pm1.8$ & $90.3\pm1.1$ & $47.5\pm1.4$ & $90.7\pm1.0$ & $16.2\pm0.7$ & $40.7\pm1.6$ & $4.6\pm0.4$  \\
       & $\text{B-cos}_\text{ASAP}$ & $49.8\pm1.3$ & $91.7\pm0.5$ & $37.7\pm1.9$ & $93.1\pm1.0$ & $13.4\pm1.1$ & $31.5\pm1.7$ & $2.8\pm0.4$  \\
     \bottomrule
    \end{tabular}
    }
    \label{tab:VinBigXray_Results}
\end{table}

\section{Discussion}
\label{sec:dicussion}

\begin{figure}[t]
    \centering
    \includegraphics[
        %width=0.75\linewidth,
        width=1.0\linewidth,
        trim={0mm 223mm 0mm 30mm},
        clip
    ]{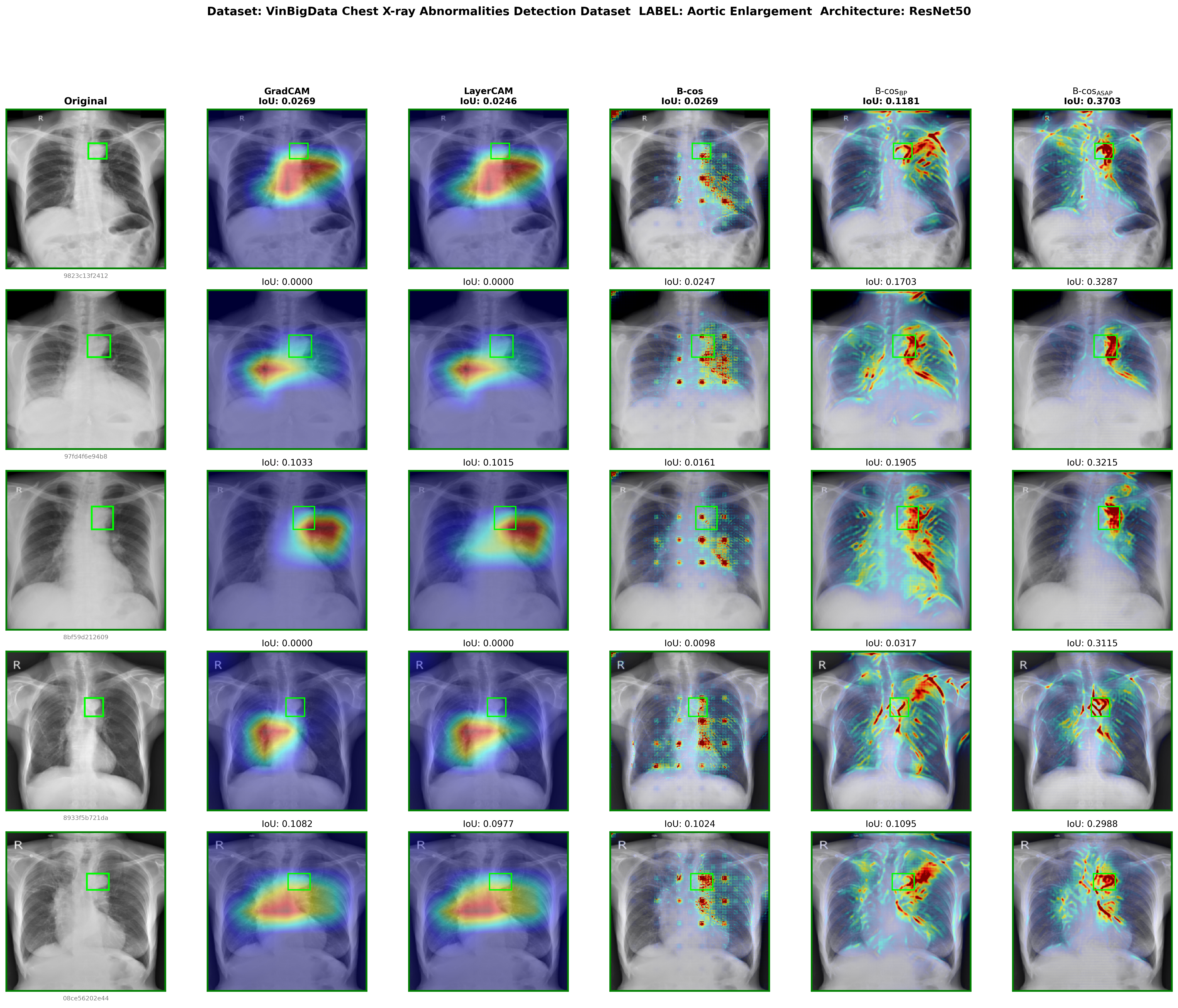}
    \caption{Visualizing explanations from Baseline and B-cos networks using post-hoc attribution methods and inherent explanations, respectively, for ResNet50 architecture on the VinBig Chest X-Ray dataset, for the ground truth label: Aortic Enlargement.}
    \label{fig:visualize_vinbig_resnet50}
\end{figure}

This work explores the potential of B-cos networks for chest X-ray classification. We show that they provide faithful, high-quality explanations grounded in the model architecture~\cite{boehle_b-cos_2022}, while achieving competitive classification performance.

Applying B-cos networks to the RSNA Pneumonia Detection Challenge~\cite{rsna_pneumonia_dataset} and VinBigData~\cite{nguyen2020vindrcxr} datasets revealed one critical limitation: explanation maps are impaired by aliasing artifacts from strided convolutions. Replacing these operations with anti-aliasing layers, namely $\mathrm{BlurPool}$~\cite{blurpool} and $\mathrm{ASAP}$~\cite{asap_2023}, removes these artifacts and substantially improves interpretability.

With these modifications, B-cos networks provide faithful, high-resolution explanations aligned with disease-relevant regions, while preserving performance competitive with baseline models. Our precision-based energy pointing game ($\mathrm{EPG}$) metrics show that over 50\% of high-confidence activations fall inside bounding boxes, although these cover only 11.65\% of the image area. This confirms that the models focus on clinically meaningful regions. Importantly, our goal is not to improve classifier performance, but to obtain faithful, aliasing-free B-cos explanations without sacrificing it. This is reflected in the results: on RSNA with ConvNeXt-B, $\text{B-cos}_\text{ASAP/BP}$ are only about 2\% lower in accuracy than the baseline, but improve recall from 45\% to $\approx$80\% and AUROC from 82\% to 88\%, while adding inherent explanations. On VinBigXray with ConvNeXt-B, $\text{B-cos}_\text{BP}$ is on par with or better than the baseline in accuracy, recall, precision, and AUROC, while $\text{B-cos}_\text{ASAP}$ shows a recall trade-off but maintains high AUROC and improves mIoU. A stronger encoder further improves metrics, as shown by ConvNeXt-B compared to ResNet50.

All metrics are reported as mean $\pm$ standard deviation over 3 seeds and 5 folds. Paired $t$-tests over matched seed/fold runs show that $\text{B-cos}_\text{ASAP}$ and $\text{B-cos}_\text{BP}$ significantly improve mIoU over B-cos in all dataset-backbone comparisons, with maximum $p=7.0\mathrm{e}{-3}$. On RSNA, neither variant is significantly worse than B-cos in Accuracy, Recall, Precision, or AUROC, indicating improved localization without loss of predictive performance. While $\text{B-cos}_\text{ASAP}$ and $\text{B-cos}_\text{BP}$ perform similarly overall, with different trade-offs, ASAP gives slightly better visual explanations, as shown in \cref{fig:visualize_vinbig_resnet50}.

Some limitations remain. We did not optimize for high recall alone, since this can increase false positives and reduce clinician trust~\cite{chen_humancentered_2022}; instead, we aimed for balanced performance across metrics. Moreover, although B-cos was evaluated on multiple architectures, interpretability for vision transformers~\cite{vision_transformer_original} was not analyzed in depth, and the effect of anti-aliasing in these models remains open.
%Future work could integrate spatial priors or bounding box information into B-cos networks to improve localization further. Applying B-cos networks to larger datasets without explicit annotations is another promising direction to study the trade-off between scalability and interpretability. 
While this work focuses on medical applications, the methodology is broadly applicable, for example, to industrial anomaly detection~\cite{poggi2025smart}.

%\noindent\textbf{Statistical Rigor.} 

\section{Conclusion}
\label{sec:conclusion}
This work is a comprehensive analysis of B-cos networks for chest X-ray classification, targeting the key limitation that has restricted their application in medical imaging: aliasing artifacts in explanation maps. We address them by integrating anti-aliasing techniques, $\mathrm{ASAP}$, and $\mathrm{BlurPool}$, which improve the clarity and reliability of the explanations. We demonstrate that the anti-aliasing methods work with both multi-class and multi-label B-cos networks to produce contribution maps, making the method suitable for realistic diagnostic tasks.
Our experiments on the RSNA Pneumonia Detection Challenge and VinBigData Chest X-ray Abnormalities Detection dataset show that B-cos networks provide competitive classification performance while offering inherently interpretable, high-fidelity explanations. These explanations align closely with disease-relevant regions, as demonstrated both qualitatively and through strong energy-based pointing game scores, outperforming post-hoc attribution methods like LayerCAM and GradCAM.
By enhancing the interpretability and practical applicability of B-cos networks, this work takes an important step toward transparent and trustworthy AI-assisted diagnosis in clinical settings~\cite{dlxraysurvey}, and sets the stage for future work on scalable, interpretable deep learning for medical imaging. 
%Code for the work will be released after acceptance.
% We share additional visualizations as a video in the supplementary material.

\section*{Acknowledgements. }S.A.\ and M.K.\ acknowledge support by DFG Research Unit 5336 -- Learning to Sense. Margret Keuper acknowledges funding by BMFTR project TrackOpt (01IS24074A-D).

\section*{Disclosure of Interests. }There are no conflicting interests.
{%\scriptsize
\bibliography{main}
}

\newpage
\appendix
{
    \centering
    \Large
    \textbf{Faithful, Interpretable Chest X-ray Diagnosis\\with Artifact-free B-cos Networks} \\
    \vspace{0.5em}Supplementary Material \\
    \vspace{1.0em}
}
\section{Scope and Reporting Conventions}
This appendix extends \emph{Faithful, Interpretable Chest X-ray Diagnosis with
Artifact-free B-cos Networks}. Medical comparisons use final completed
checkpoints and three seeds (5, 420 and 1337) across five folds. Each of the
15 seed/fold runs receives equal weight; reported dispersion is the sample
standard deviation across runs. All tabulated metrics are percentages.

Tables~\ref{tab:pneumonia-combined} and~\ref{tab:vinbigxray-combined}
combine classification and inherent-explanation scores across Augmentation
and sampling choices. Their marked blocks match the main paper ResNet50
configurations: heavy Augmentation with Oversampling on RSNA Pneumonia and
light Augmentation without sampling on VinBig.
Table~\ref{tab:training-configurations} in Section~\ref{sec:training-configurations}
lists the training configuration behind every reported model.

\subsection{How Localization is Averaged}
For RSNA Pneumonia, each run first averages the defined image-level localization
scores over evaluated images with annotated regions. EPG scores are averaged
over boxes within an image before this image average. For VinBig, each run
instead pools all evaluated image-disease pairs across disease labels and
averages their scores (the micro summary). Thus a disease with more evaluated
pairs contributes more observations than a rarer disease; disease classes are
not assigned equal weight. For both datasets, the final table then averages
the 15 run summaries, with sample SD across those same 15 summaries.

EPG recall and precision are evaluated at attribution threshold $\tau=0$.
The column named mIoU follows the main paper naming and denotes the mean
bounding-box localization IoU: the positive attribution map is min-max
normalized, binarized at 0.5, and compared with the union of ground-truth boxes.
This mask threshold differs from the EPG attribution threshold.

\clearpage
\section{Medical Classification and Inherent Explanations}
\begin{table}[!htbp]
\centering\small
\setlength{\tabcolsep}{3pt}
\renewcommand{\arraystretch}{1.08}
\caption{ResNet50 classification and inherent-explanation localization on RSNA Pneumonia.}\label{tab:pneumonia-combined}
\begin{adjustbox}{max width=\linewidth}\begin{tabular}{crrrrrrr@{}}
\toprule
Model & Accuracy (\%) & AUROC & F1 Score & $\mathrm{EPG}_R$ & $\mathrm{EPG}_P$ & Recall & mIoU \\
\midrule
\multicolumn{8}{c}{\textbf{No Augmentation / No Sampling}}\\[2pt]
Baseline & $83.4\pm0.5$ & $83.7\pm0.8$ & $60.5\pm1.3$ & -- & -- & $56.4\pm2.1$ & -- \\
$\text{B-cos}$ & $83.0\pm0.7$ & $83.5\pm0.8$ & $58.9\pm1.2$ & $99.5\pm0.2$ & $16.3\pm0.7$ & $54.1\pm2.0$ & $1.1\pm0.1$ \\
$\text{B-cos}_{\mathrm{BP}}$ & $81.5\pm0.8$ & $80.4\pm1.1$ & $56.8\pm1.7$ & $99.1\pm0.9$ & $14.1\pm1.2$ & $53.8\pm1.8$ & $4.8\pm0.9$ \\
$\text{B-cos}_{\mathrm{ASAP}}$ & $82.2\pm0.5$ & $81.2\pm0.9$ & $57.7\pm1.4$ & $99.8\pm0.3$ & $15.5\pm1.9$ & $53.8\pm2.2$ & $4.1\pm1.1$ \\
\midrule
\multicolumn{8}{c}{\textbf{No Augmentation / Oversampling}}\\[2pt]
Baseline & $83.3\pm0.4$ & $83.0\pm1.0$ & $57.3\pm1.9$ & -- & -- & $49.7\pm2.7$ & -- \\
$\text{B-cos}$ & $82.8\pm0.5$ & $83.4\pm0.8$ & $58.4\pm1.4$ & $99.5\pm0.2$ & $15.6\pm0.7$ & $53.8\pm1.9$ & $0.8\pm0.1$ \\
$\text{B-cos}_{\mathrm{BP}}$ & $81.8\pm0.5$ & $80.7\pm1.1$ & $55.9\pm1.1$ & $99.9\pm0.2$ & $16.1\pm1.6$ & $51.2\pm1.7$ & $6.2\pm1.3$ \\
$\text{B-cos}_{\mathrm{ASAP}}$ & $82.1\pm0.5$ & $80.3\pm1.7$ & $54.0\pm2.4$ & $99.9\pm0.1$ & $15.8\pm1.4$ & $46.9\pm4.1$ & $4.1\pm0.6$ \\
\midrule
\multicolumn{8}{c}{\textbf{Light Augmentation / No Sampling}}\\[2pt]
Baseline & $83.4\pm0.4$ & $83.3\pm0.8$ & $60.2\pm1.2$ & -- & -- & $55.6\pm1.8$ & -- \\
$\text{B-cos}$ & $82.2\pm0.7$ & $84.3\pm1.0$ & $58.6\pm1.6$ & $100.0\pm0.0$ & $22.8\pm0.6$ & $55.9\pm1.9$ & $3.6\pm0.4$ \\
$\text{B-cos}_{\mathrm{BP}}$ & $82.2\pm0.6$ & $83.5\pm1.0$ & $57.8\pm1.5$ & $98.0\pm2.1$ & $17.9\pm1.4$ & $54.0\pm2.4$ & $9.7\pm1.2$ \\
$\text{B-cos}_{\mathrm{ASAP}}$ & $82.6\pm0.5$ & $83.3\pm1.0$ & $58.2\pm1.5$ & $96.2\pm1.8$ & $18.2\pm0.9$ & $53.9\pm2.4$ & $8.0\pm1.2$ \\
\midrule
\multicolumn{8}{c}{\textbf{Light Augmentation / Oversampling}}\\[2pt]
Baseline & $83.3\pm0.4$ & $83.1\pm1.0$ & $57.3\pm1.4$ & -- & -- & $49.9\pm2.0$ & -- \\
$\text{B-cos}$ & $81.8\pm0.7$ & $84.6\pm0.8$ & $60.4\pm1.2$ & $100.0\pm0.0$ & $23.1\pm0.6$ & $61.8\pm2.4$ & $3.6\pm0.3$ \\
$\text{B-cos}_{\mathrm{BP}}$ & $81.1\pm0.7$ & $83.6\pm0.6$ & $59.9\pm1.2$ & $98.3\pm2.5$ & $17.7\pm1.4$ & $62.7\pm2.6$ & $10.2\pm1.6$ \\
$\text{B-cos}_{\mathrm{ASAP}}$ & $81.4\pm0.6$ & $84.0\pm0.9$ & $60.2\pm1.3$ & $98.3\pm2.1$ & $18.4\pm1.1$ & $62.3\pm2.3$ & $8.3\pm0.9$ \\
\midrule
\multicolumn{8}{c}{\textbf{Heavy Augmentation / No Sampling}}\\[2pt]
Baseline & $83.4\pm0.6$ & $83.5\pm1.0$ & $60.4\pm1.5$ & -- & -- & $56.1\pm1.9$ & -- \\
$\text{B-cos}$ & $84.1\pm0.4$ & $87.7\pm0.6$ & $61.2\pm1.2$ & $100.0\pm0.0$ & $24.0\pm0.7$ & $55.6\pm1.7$ & $4.8\pm0.5$ \\
$\text{B-cos}_{\mathrm{BP}}$ & $84.1\pm0.3$ & $87.6\pm0.5$ & $61.1\pm0.9$ & $97.8\pm1.5$ & $19.6\pm1.0$ & $55.3\pm1.6$ & $10.4\pm1.1$ \\
$\text{B-cos}_{\mathrm{ASAP}}$ & $84.1\pm0.5$ & $87.7\pm0.7$ & $61.4\pm1.1$ & $89.9\pm3.3$ & $18.9\pm1.2$ & $56.1\pm2.2$ & $9.2\pm1.3$ \\
\midrule
\multicolumn{8}{c}{\textbf{Heavy Augmentation / Oversampling $\dagger$}}\\[2pt]
Baseline & $83.5\pm0.4$ & $83.3\pm0.6$ & $58.0\pm1.3$ & -- & -- & $50.6\pm1.7$ & -- \\
$\text{B-cos}$ & $80.5\pm0.8$ & $86.7\pm0.7$ & $63.3\pm1.1$ & $100.0\pm0.0$ & $24.1\pm0.8$ & $74.8\pm2.2$ & $4.7\pm0.5$ \\
$\text{B-cos}_{\mathrm{BP}}$ & $79.2\pm0.9$ & $87.1\pm0.7$ & $63.2\pm0.9$ & $98.7\pm1.4$ & $20.1\pm1.9$ & $79.2\pm1.7$ & $10.6\pm2.0$ \\
$\text{B-cos}_{\mathrm{ASAP}}$ & $79.9\pm0.6$ & $87.1\pm0.5$ & $63.5\pm0.6$ & $98.6\pm0.9$ & $21.1\pm1.3$ & $77.5\pm1.1$ & $9.6\pm1.8$ \\
\bottomrule
\end{tabular}\end{adjustbox}
\par\smallskip
\begin{minipage}{0.98\linewidth}\footnotesize All metrics are percentages at the final checkpoint (epoch 29), reported as mean $\pm$ sample SD across $n=15$ equally weighted seed/fold runs. The $\dagger$ block is the main paper configuration. EPG and mIoU use inherent explanations; dashes indicate that these are undefined for Baseline.\end{minipage}
\end{table}
\clearpage
\begin{table}[!htbp]
\centering\small
\setlength{\tabcolsep}{3pt}
\renewcommand{\arraystretch}{1.08}
\caption{ResNet50 classification and inherent-explanation localization on VinBig.}\label{tab:vinbigxray-combined}
\begin{adjustbox}{max width=\linewidth}\begin{tabular}{crrrrrrr@{}}
\toprule
Model & mAP & AUROC & F1 Score & $\mathrm{EPG}_R^{\rm micro}$ & $\mathrm{EPG}_P^{\rm micro}$ & Recall & mIoU \\
\midrule
\multicolumn{8}{c}{\textbf{No Augmentation / No Sampling}}\\[2pt]
Baseline & $50.7\pm1.6$ & $80.0\pm1.2$ & $45.9\pm1.6$ & -- & -- & $38.2\pm1.3$ & -- \\
$\text{B-cos}$ & $46.3\pm1.8$ & $84.1\pm0.9$ & $43.7\pm1.7$ & $89.2\pm0.4$ & $7.7\pm0.2$ & $38.5\pm1.7$ & $1.0\pm0.1$ \\
$\text{B-cos}_{\mathrm{BP}}$ & $43.7\pm1.1$ & $81.6\pm0.9$ & $41.9\pm1.6$ & $89.1\pm0.5$ & $7.5\pm0.2$ & $36.6\pm1.6$ & $3.9\pm0.6$ \\
$\text{B-cos}_{\mathrm{ASAP}}$ & $43.3\pm1.2$ & $81.2\pm0.9$ & $40.4\pm1.6$ & $89.2\pm0.9$ & $8.3\pm0.3$ & $34.3\pm1.7$ & $3.6\pm1.0$ \\
\midrule
\multicolumn{8}{c}{\textbf{No Augmentation / Oversampling}}\\[2pt]
Baseline & $30.2\pm2.1$ & $73.8\pm1.3$ & $27.8\pm2.1$ & -- & -- & $30.8\pm1.8$ & -- \\
$\text{B-cos}$ & $26.8\pm1.4$ & $73.5\pm0.6$ & $28.4\pm1.3$ & $93.0\pm0.6$ & $7.2\pm0.1$ & $41.3\pm1.3$ & $1.1\pm0.1$ \\
$\text{B-cos}_{\mathrm{BP}}$ & $26.1\pm1.1$ & $72.2\pm1.1$ & $27.0\pm1.2$ & $92.8\pm0.5$ & $6.5\pm0.2$ & $33.5\pm1.1$ & $2.8\pm0.3$ \\
$\text{B-cos}_{\mathrm{ASAP}}$ & $25.9\pm1.3$ & $72.5\pm0.7$ & $25.2\pm1.6$ & $93.5\pm0.4$ & $7.4\pm0.2$ & $30.1\pm1.2$ & $2.7\pm0.6$ \\
\midrule
\multicolumn{8}{c}{\textbf{Light Augmentation / No Sampling $\dagger$}}\\[2pt]
Baseline & $50.6\pm1.4$ & $80.8\pm1.1$ & $46.2\pm1.9$ & -- & -- & $38.6\pm1.8$ & -- \\
$\text{B-cos}$ & $53.5\pm1.4$ & $91.9\pm0.6$ & $44.5\pm1.5$ & $97.7\pm0.4$ & $9.4\pm0.2$ & $38.6\pm1.5$ & $2.3\pm0.2$ \\
$\text{B-cos}_{\mathrm{BP}}$ & $47.8\pm1.3$ & $87.7\pm0.9$ & $42.3\pm1.6$ & $92.3\pm1.2$ & $9.0\pm0.2$ & $37.3\pm1.6$ & $5.4\pm0.7$ \\
$\text{B-cos}_{\mathrm{ASAP}}$ & $47.6\pm1.3$ & $87.3\pm0.8$ & $42.2\pm1.2$ & $92.4\pm1.3$ & $9.9\pm0.4$ & $37.0\pm1.1$ & $5.2\pm0.9$ \\
\midrule
\multicolumn{8}{c}{\textbf{Light Augmentation / Oversampling}}\\[2pt]
Baseline & $29.7\pm2.0$ & $73.5\pm1.1$ & $27.4\pm2.1$ & -- & -- & $30.5\pm1.7$ & -- \\
$\text{B-cos}$ & $31.8\pm1.5$ & $75.2\pm0.5$ & $31.1\pm1.1$ & $93.9\pm1.2$ & $9.5\pm0.1$ & $47.7\pm1.2$ & $2.5\pm0.3$ \\
$\text{B-cos}_{\mathrm{BP}}$ & $27.7\pm1.3$ & $72.8\pm0.7$ & $28.4\pm1.1$ & $90.4\pm0.7$ & $8.8\pm0.2$ & $40.8\pm1.6$ & $5.1\pm0.6$ \\
$\text{B-cos}_{\mathrm{ASAP}}$ & $27.7\pm1.3$ & $72.4\pm0.6$ & $28.6\pm1.2$ & $89.0\pm0.7$ & $9.2\pm0.2$ & $39.9\pm1.3$ & $4.3\pm1.1$ \\
\midrule
\multicolumn{8}{c}{\textbf{Heavy Augmentation / No Sampling}}\\[2pt]
Baseline & $50.7\pm1.9$ & $80.7\pm1.4$ & $46.0\pm2.0$ & -- & -- & $38.4\pm1.7$ & -- \\
$\text{B-cos}$ & $51.7\pm1.5$ & $92.3\pm0.5$ & $38.2\pm1.3$ & $98.3\pm0.2$ & $8.5\pm0.2$ & $31.3\pm1.3$ & $2.0\pm0.3$ \\
$\text{B-cos}_{\mathrm{BP}}$ & $49.6\pm1.2$ & $90.9\pm0.5$ & $38.9\pm1.6$ & $96.2\pm0.6$ & $8.4\pm0.3$ & $32.6\pm1.4$ & $5.6\pm0.7$ \\
$\text{B-cos}_{\mathrm{ASAP}}$ & $49.9\pm1.2$ & $91.1\pm0.5$ & $38.8\pm1.5$ & $97.7\pm0.4$ & $9.3\pm0.4$ & $32.6\pm1.3$ & $5.7\pm0.9$ \\
\midrule
\multicolumn{8}{c}{\textbf{Heavy Augmentation / Oversampling}}\\[2pt]
Baseline & $29.8\pm1.8$ & $73.5\pm0.8$ & $27.5\pm2.1$ & -- & -- & $30.6\pm1.6$ & -- \\
$\text{B-cos}$ & $32.1\pm1.6$ & $75.2\pm0.8$ & $31.0\pm1.1$ & $97.2\pm0.3$ & $9.4\pm0.2$ & $46.4\pm1.3$ & $3.0\pm0.3$ \\
$\text{B-cos}_{\mathrm{BP}}$ & $29.9\pm1.6$ & $74.1\pm0.6$ & $29.1\pm1.1$ & $93.2\pm0.6$ & $9.1\pm0.4$ & $45.4\pm1.6$ & $6.4\pm0.6$ \\
$\text{B-cos}_{\mathrm{ASAP}}$ & $29.7\pm1.7$ & $74.2\pm0.9$ & $29.5\pm1.3$ & $91.8\pm1.1$ & $9.5\pm0.3$ & $45.3\pm1.5$ & $5.7\pm0.7$ \\
\bottomrule
\end{tabular}\end{adjustbox}
\par\smallskip
\begin{minipage}{0.98\linewidth}\footnotesize All metrics are percentages at the final checkpoint (epoch 29), reported as mean $\pm$ sample SD across $n=15$ equally weighted seed/fold runs. The $\dagger$ block is the main paper configuration. EPG and mIoU use inherent explanations; dashes indicate that these are undefined for Baseline.\end{minipage}
\end{table}

\newpage
\subsection{Additional DenseNet architecture}
\begin{table}[!htbp]
\centering\small
\setlength{\tabcolsep}{3pt}
\renewcommand{\arraystretch}{1.08}
\caption{DenseNet-121 classification and inherent-explanation localization on VinBig.}\label{tab:densenet-combined}
\begin{adjustbox}{max width=\linewidth}\begin{tabular}{crrrrrrr@{}}
\toprule
Model & mAP & AUROC & F1 Score & $\mathrm{EPG}_R^{\rm micro}$ & $\mathrm{EPG}_P^{\rm micro}$ & Recall & mIoU \\
\midrule
\multicolumn{8}{c}{\textbf{Heavy Augmentation / No Sampling}}\\[2pt]
Baseline & $52.1\pm1.3$ & $85.0\pm1.0$ & $46.9\pm1.9$ & -- & -- & $39.8\pm1.8$ & -- \\
$\text{B-cos}$ & $49.0\pm0.9$ & $92.6\pm0.4$ & $33.7\pm1.0$ & $99.2\pm0.2$ & $6.9\pm0.2$ & $27.4\pm0.8$ & $1.9\pm0.4$ \\
$\text{B-cos}_{\mathrm{BP}}$ & $47.5\pm1.1$ & $92.5\pm0.4$ & $32.9\pm0.8$ & $99.2\pm0.2$ & $7.1\pm0.4$ & $26.9\pm0.8$ & $4.3\pm0.5$ \\
$\text{B-cos}_{\mathrm{ASAP}}$ & $47.7\pm0.9$ & $92.5\pm0.4$ & $33.1\pm0.8$ & $99.1\pm0.2$ & $7.2\pm0.4$ & $27.2\pm0.9$ & $3.1\pm0.5$ \\
\bottomrule
\end{tabular}\end{adjustbox}
\par\smallskip
\begin{minipage}{0.98\linewidth}\footnotesize All values are percentages, mean $\pm$ sample SD across $n=15$ runs at epoch 29. The recipe uses pretrained initialization, Adam, learning rate $10^{-4}$, batch size 64 and a cosine schedule with three warm-up epochs. EPG and mIoU use inherent explanations; they are undefined for Baseline.\end{minipage}
\end{table}
%\clearpage
\section{Comparison of explanation methods}
The ResNet50 and ConvNeXt comparisons below use the exact main paper training,
Augmentation and sampling configurations. VinBig ConvNeXt-tiny uses AdamW
with five warm-up epochs; ConvNeXt-base uses Adam with three warm-up epochs.
Both use learning rate $10^{-3}$, batch size 16 and 30 training epochs.
RSNA ConvNeXt-base uses Adam with five warm-up epochs, learning rate $10^{-3}$, batch size 16 and 30 training epochs.
DenseNet-121 uses the additional VinBig recipe in Table~\ref{tab:densenet-combined}.
Every displayed estimate requires the full three-seed, five-fold grid.
\begin{table}[!htbp]
\centering\small
\setlength{\tabcolsep}{8pt}
\renewcommand{\arraystretch}{1.06}
\caption{Explanation localization for RSNA Pneumonia, ResNet50, heavy Augmentation and Oversampling. All displayed rows use $n=15$ (three seeds and five folds).}\label{tab:explanation-rsna_resnet50}
\begin{adjustbox}{max width=\linewidth}\begin{tabular}{clrrr@{}}
\toprule
Model & Explanation & $\mathrm{EPG}_R$ & $\mathrm{EPG}_P$ & mIoU \\
\midrule
Baseline & FinerCAM & $100.0\pm0.0$ & $16.3\pm1.0$ & $4.4\pm1.3$ \\
 & GradCAM & $100.0\pm0.0$ & $16.3\pm1.0$ & $4.4\pm1.3$ \\
 & GradCAM++ & $100.0\pm0.0$ & $16.3\pm1.0$ & $4.4\pm1.3$ \\
 & Grad$\times$Input & $50.0\pm0.0$ & $15.4\pm0.5$ & $0.3\pm0.0$ \\
 & HiResCAM & $100.0\pm0.0$ & $16.3\pm1.0$ & $4.4\pm1.3$ \\
 & LayerCAM & $100.0\pm0.0$ & $16.2\pm1.0$ & $4.4\pm1.3$ \\
\midrule
$\text{B-cos}$ & FinerCAM & $100.0\pm0.0$ & $17.6\pm0.4$ & $17.3\pm1.4$ \\
 & GradCAM & $100.0\pm0.0$ & $17.6\pm0.4$ & $17.3\pm1.4$ \\
 & GradCAM++ & $100.0\pm0.0$ & $18.2\pm1.2$ & $13.7\pm1.6$ \\
 & Grad$\times$Input & $50.4\pm0.1$ & $16.1\pm0.8$ & $0.7\pm0.0$ \\
 & HiResCAM & $100.0\pm0.0$ & $20.5\pm0.5$ & $13.3\pm1.1$ \\
 & LayerCAM & $100.0\pm0.0$ & $20.6\pm0.5$ & $13.3\pm1.3$ \\
 & Inherent B-cos & $100.0\pm0.0$ & $24.1\pm0.8$ & $4.7\pm0.5$ \\
\midrule
$\text{B-cos}_{\mathrm{BP}}$ & FinerCAM & $100.0\pm0.0$ & $14.8\pm2.2$ & $15.7\pm3.8$ \\
 & GradCAM & $100.0\pm0.0$ & $14.8\pm2.2$ & $15.7\pm3.8$ \\
 & GradCAM++ & $100.0\pm0.0$ & $9.2\pm1.0$ & $15.4\pm3.2$ \\
 & Grad$\times$Input & $50.7\pm0.2$ & $17.9\pm0.8$ & $0.9\pm0.0$ \\
 & HiResCAM & $100.0\pm0.0$ & $18.4\pm3.0$ & $10.2\pm2.7$ \\
 & LayerCAM & $100.0\pm0.0$ & $18.0\pm3.2$ & $10.4\pm2.7$ \\
 & Inherent B-cos & $98.7\pm1.4$ & $20.1\pm1.9$ & $10.6\pm2.0$ \\
\midrule
$\text{B-cos}_{\mathrm{ASAP}}$ & FinerCAM & $100.0\pm0.0$ & $15.1\pm0.9$ & $22.9\pm2.9$ \\
 & GradCAM & $100.0\pm0.0$ & $15.2\pm0.9$ & $22.9\pm2.9$ \\
 & GradCAM++ & $100.0\pm0.0$ & $14.8\pm2.8$ & $21.6\pm2.3$ \\
 & Grad$\times$Input & $50.8\pm0.2$ & $17.1\pm0.6$ & $0.8\pm0.0$ \\
 & HiResCAM & $100.0\pm0.0$ & $19.5\pm1.3$ & $16.6\pm3.4$ \\
 & LayerCAM & $100.0\pm0.0$ & $19.8\pm1.5$ & $15.9\pm3.4$ \\
 & Inherent B-cos & $98.6\pm0.9$ & $21.1\pm1.3$ & $9.6\pm1.8$ \\
\bottomrule
\end{tabular}\end{adjustbox}
\par\smallskip
\begin{minipage}{0.97\linewidth}\footnotesize
Values are percentages (mean $\pm$ sample SD). Inherent B-cos is undefined for Baseline. Non-negative CAMs can yield $\mathrm{EPG}_R=100\%$ without perfect spatial localization.
\end{minipage}
\end{table}
\clearpage
\begin{table}[!htbp]
\centering\small
\setlength{\tabcolsep}{8pt}
\renewcommand{\arraystretch}{1.06}
\caption{Explanation localization for RSNA Pneumonia, ConvNeXt-base, heavy Augmentation and Oversampling. All displayed rows use $n=15$ (three seeds and five folds).}\label{tab:explanation-rsna_convnext_base}
\begin{adjustbox}{max width=\linewidth}\begin{tabular}{clrrr@{}}
\toprule
Model & Explanation & $\mathrm{EPG}_R$ & $\mathrm{EPG}_P$ & mIoU \\
\midrule
Baseline & GradCAM & $100.0\pm0.0$ & $15.3\pm0.8$ & $18.7\pm1.4$ \\
 & GradCAM++ & $100.0\pm0.0$ & $16.9\pm2.7$ & $17.1\pm5.2$ \\
 & Grad$\times$Input & $50.0\pm0.0$ & $17.0\pm0.6$ & $0.8\pm0.1$ \\
 & HiResCAM & $100.0\pm0.0$ & $15.3\pm0.8$ & $18.7\pm1.4$ \\
 & LayerCAM & $100.0\pm0.0$ & $15.6\pm2.1$ & $20.3\pm5.6$ \\
\midrule
$\text{B-cos}$ & GradCAM & $100.0\pm0.0$ & $16.4\pm2.0$ & $15.9\pm3.4$ \\
 & GradCAM++ & $100.0\pm0.0$ & $12.9\pm6.4$ & $13.9\pm6.7$ \\
 & Grad$\times$Input & $50.3\pm0.1$ & $21.9\pm0.7$ & $0.6\pm0.1$ \\
 & HiResCAM & $100.0\pm0.0$ & $16.4\pm2.0$ & $15.9\pm3.4$ \\
 & LayerCAM & $100.0\pm0.0$ & $12.3\pm6.3$ & $12.4\pm6.9$ \\
 & Inherent B-cos & $97.8\pm0.6$ & $18.6\pm1.1$ & $1.4\pm0.5$ \\
\midrule
$\text{B-cos}_{\mathrm{BP}}$ & GradCAM & $100.0\pm0.0$ & $17.7\pm2.3$ & $15.8\pm2.5$ \\
 & GradCAM++ & $100.0\pm0.0$ & $13.2\pm7.0$ & $9.1\pm4.3$ \\
 & Grad$\times$Input & $50.5\pm0.2$ & $22.4\pm0.9$ & $0.8\pm0.0$ \\
 & HiResCAM & $100.0\pm0.0$ & $17.7\pm2.3$ & $15.8\pm2.5$ \\
 & LayerCAM & $100.0\pm0.0$ & $13.3\pm7.1$ & $8.9\pm4.6$ \\
 & Inherent B-cos & $93.9\pm1.5$ & $20.1\pm1.6$ & $2.1\pm0.6$ \\
\midrule
$\text{B-cos}_{\mathrm{ASAP}}$ & GradCAM & $100.0\pm0.0$ & $16.2\pm3.0$ & $11.5\pm6.0$ \\
 & GradCAM++ & $100.0\pm0.0$ & $13.2\pm5.9$ & $8.8\pm7.3$ \\
 & Grad$\times$Input & $50.5\pm0.2$ & $23.2\pm1.5$ & $0.7\pm0.1$ \\
 & HiResCAM & $100.0\pm0.0$ & $16.2\pm3.0$ & $11.5\pm6.0$ \\
 & LayerCAM & $100.0\pm0.0$ & $13.2\pm5.9$ & $9.2\pm6.8$ \\
 & Inherent B-cos & $94.4\pm1.7$ & $21.9\pm2.8$ & $2.0\pm0.3$ \\
\bottomrule
\end{tabular}\end{adjustbox}
\par\smallskip
\begin{minipage}{0.97\linewidth}\footnotesize
Values are percentages (mean $\pm$ sample SD). Inherent B-cos is undefined for Baseline. Non-negative CAMs can yield $\mathrm{EPG}_R=100\%$ without perfect spatial localization.
 This comparison includes methods available for all four variants across the full 15-run grid. FinerCAM does not meet that coverage requirement for this configuration.
\end{minipage}
\end{table}
\clearpage
\begin{table}[!htbp]
\centering\small
\setlength{\tabcolsep}{8pt}
\renewcommand{\arraystretch}{1.06}
\caption{Explanation localization for VinBig, ResNet50, light Augmentation and No Sampling. All displayed rows use $n=15$ (three seeds and five folds).}\label{tab:explanation-vinbig_resnet50}
\begin{adjustbox}{max width=\linewidth}\begin{tabular}{clrrr@{}}
\toprule
Model & Explanation & $\mathrm{EPG}_R^{\rm micro}$ & $\mathrm{EPG}_P^{\rm micro}$ & mIoU \\
\midrule
Baseline & FinerCAM & $100.0\pm0.0$ & $9.4\pm0.5$ & $9.2\pm0.7$ \\
 & GradCAM & $100.0\pm0.0$ & $9.4\pm0.5$ & $9.2\pm0.7$ \\
 & GradCAM++ & $100.0\pm0.0$ & $9.5\pm0.4$ & $10.5\pm0.6$ \\
 & Grad$\times$Input & $50.4\pm0.1$ & $7.2\pm0.2$ & $0.3\pm0.0$ \\
 & HiResCAM & $100.0\pm0.0$ & $9.4\pm0.5$ & $9.2\pm0.7$ \\
 & LayerCAM & $100.0\pm0.0$ & $9.2\pm0.4$ & $10.3\pm0.6$ \\
\midrule
$\text{B-cos}$ & FinerCAM & $100.0\pm0.0$ & $8.0\pm0.1$ & $13.0\pm0.2$ \\
 & GradCAM & $100.0\pm0.0$ & $8.0\pm0.1$ & $13.0\pm0.2$ \\
 & GradCAM++ & $100.0\pm0.0$ & $7.9\pm0.3$ & $12.9\pm0.3$ \\
 & Grad$\times$Input & $52.6\pm0.2$ & $7.7\pm0.1$ & $0.4\pm0.0$ \\
 & HiResCAM & $100.0\pm0.0$ & $9.0\pm0.1$ & $12.6\pm0.3$ \\
 & LayerCAM & $100.0\pm0.0$ & $8.9\pm0.2$ & $12.5\pm0.3$ \\
 & Inherent B-cos & $97.7\pm0.4$ & $9.4\pm0.2$ & $2.3\pm0.2$ \\
\midrule
$\text{B-cos}_{\mathrm{BP}}$ & FinerCAM & $100.0\pm0.0$ & $8.5\pm0.2$ & $11.6\pm0.6$ \\
 & GradCAM & $100.0\pm0.0$ & $8.5\pm0.2$ & $11.6\pm0.6$ \\
 & GradCAM++ & $100.0\pm0.0$ & $8.4\pm0.2$ & $11.4\pm0.8$ \\
 & Grad$\times$Input & $53.4\pm0.5$ & $7.7\pm0.1$ & $0.6\pm0.1$ \\
 & HiResCAM & $100.0\pm0.0$ & $9.4\pm0.4$ & $10.6\pm0.7$ \\
 & LayerCAM & $100.0\pm0.0$ & $9.3\pm0.4$ & $10.5\pm0.8$ \\
 & Inherent B-cos & $92.3\pm1.2$ & $9.0\pm0.2$ & $5.4\pm0.7$ \\
\midrule
$\text{B-cos}_{\mathrm{ASAP}}$ & FinerCAM & $100.0\pm0.0$ & $8.9\pm0.4$ & $12.6\pm0.7$ \\
 & GradCAM & $100.0\pm0.0$ & $8.9\pm0.4$ & $12.6\pm0.7$ \\
 & GradCAM++ & $100.0\pm0.0$ & $9.1\pm0.5$ & $12.2\pm0.7$ \\
 & Grad$\times$Input & $52.7\pm0.4$ & $7.7\pm0.2$ & $0.6\pm0.1$ \\
 & HiResCAM & $100.0\pm0.0$ & $9.9\pm0.5$ & $11.5\pm0.7$ \\
 & LayerCAM & $100.0\pm0.0$ & $9.9\pm0.6$ & $11.3\pm0.7$ \\
 & Inherent B-cos & $92.4\pm1.3$ & $9.9\pm0.4$ & $5.2\pm0.9$ \\
\bottomrule
\end{tabular}\end{adjustbox}
\par\smallskip
\begin{minipage}{0.97\linewidth}\footnotesize
Values are percentages (mean $\pm$ sample SD). Inherent B-cos is undefined for Baseline. Non-negative CAMs can yield $\mathrm{EPG}_R=100\%$ without perfect spatial localization.
\end{minipage}
\end{table}
\clearpage
\begin{table}[!htbp]
\centering\small
\setlength{\tabcolsep}{8pt}
\renewcommand{\arraystretch}{1.06}
\caption{Explanation localization for VinBig, ConvNeXt-tiny, light Augmentation and No Sampling. All displayed rows use $n=15$ (three seeds and five folds).}\label{tab:explanation-vinbig_convnext_tiny}
\begin{adjustbox}{max width=\linewidth}\begin{tabular}{clrrr@{}}
\toprule
Model & Explanation & $\mathrm{EPG}_R^{\rm micro}$ & $\mathrm{EPG}_P^{\rm micro}$ & mIoU \\
\midrule
Baseline & FinerCAM & $100.0\pm0.0$ & $14.9\pm0.5$ & $17.3\pm0.6$ \\
 & GradCAM & $100.0\pm0.0$ & $14.9\pm0.5$ & $17.3\pm0.6$ \\
 & GradCAM++ & $100.0\pm0.0$ & $13.1\pm0.9$ & $17.0\pm0.6$ \\
 & Grad$\times$Input & $52.7\pm0.3$ & $8.5\pm0.1$ & $0.9\pm0.1$ \\
 & HiResCAM & $100.0\pm0.0$ & $14.9\pm0.5$ & $17.3\pm0.6$ \\
 & LayerCAM & $100.0\pm0.0$ & $14.2\pm0.7$ & $17.5\pm0.6$ \\
\midrule
$\text{B-cos}$ & FinerCAM & $100.0\pm0.0$ & $10.5\pm0.6$ & $12.6\pm1.3$ \\
 & GradCAM & $100.0\pm0.0$ & $10.5\pm0.6$ & $12.6\pm1.3$ \\
 & GradCAM++ & $100.0\pm0.0$ & $9.7\pm0.6$ & $11.7\pm1.3$ \\
 & Grad$\times$Input & $52.6\pm0.2$ & $10.3\pm0.1$ & $0.9\pm0.0$ \\
 & HiResCAM & $100.0\pm0.0$ & $10.5\pm0.6$ & $12.6\pm1.3$ \\
 & LayerCAM & $100.0\pm0.0$ & $9.8\pm0.5$ & $11.9\pm1.2$ \\
 & Inherent B-cos & $94.9\pm0.6$ & $12.3\pm0.4$ & $2.3\pm0.3$ \\
\midrule
$\text{B-cos}_{\mathrm{BP}}$ & FinerCAM & $100.0\pm0.0$ & $10.4\pm0.7$ & $10.9\pm1.9$ \\
 & GradCAM & $100.0\pm0.0$ & $10.4\pm0.7$ & $10.9\pm1.9$ \\
 & GradCAM++ & $100.0\pm0.0$ & $9.9\pm0.7$ & $10.0\pm1.9$ \\
 & Grad$\times$Input & $52.6\pm0.2$ & $12.1\pm0.3$ & $1.4\pm0.1$ \\
 & HiResCAM & $100.0\pm0.0$ & $10.4\pm0.7$ & $10.9\pm1.9$ \\
 & LayerCAM & $100.0\pm0.0$ & $9.9\pm0.7$ & $10.1\pm1.8$ \\
 & Inherent B-cos & $93.7\pm0.5$ & $14.6\pm0.4$ & $4.6\pm0.4$ \\
\midrule
$\text{B-cos}_{\mathrm{ASAP}}$ & FinerCAM & $100.0\pm0.0$ & $8.5\pm1.0$ & $9.9\pm1.9$ \\
 & GradCAM & $100.0\pm0.0$ & $8.5\pm1.0$ & $9.9\pm1.9$ \\
 & GradCAM++ & $100.0\pm0.0$ & $8.3\pm1.6$ & $9.4\pm2.4$ \\
 & Grad$\times$Input & $53.5\pm0.4$ & $11.0\pm0.4$ & $1.0\pm0.1$ \\
 & HiResCAM & $100.0\pm0.0$ & $8.5\pm1.0$ & $9.9\pm1.9$ \\
 & LayerCAM & $100.0\pm0.0$ & $8.3\pm1.6$ & $9.6\pm2.4$ \\
 & Inherent B-cos & $93.6\pm1.1$ & $11.4\pm1.0$ & $2.3\pm0.5$ \\
\bottomrule
\end{tabular}\end{adjustbox}
\par\smallskip
\begin{minipage}{0.97\linewidth}\footnotesize
Values are percentages (mean $\pm$ sample SD). Inherent B-cos is undefined for Baseline. Non-negative CAMs can yield $\mathrm{EPG}_R=100\%$ without perfect spatial localization.
\end{minipage}
\end{table}
\clearpage
\begin{table}[!htbp]
\centering\small
\setlength{\tabcolsep}{8pt}
\renewcommand{\arraystretch}{1.06}
\caption{Explanation localization for VinBig, ConvNeXt-base, light Augmentation and No Sampling. All displayed rows use $n=15$ (three seeds and five folds).}\label{tab:explanation-vinbig_convnext_base}
\begin{adjustbox}{max width=\linewidth}\begin{tabular}{clrrr@{}}
\toprule
Model & Explanation & $\mathrm{EPG}_R^{\rm micro}$ & $\mathrm{EPG}_P^{\rm micro}$ & mIoU \\
\midrule
Baseline & FinerCAM & $100.0\pm0.0$ & $14.1\pm0.5$ & $15.6\pm0.7$ \\
 & GradCAM & $100.0\pm0.0$ & $14.1\pm0.5$ & $15.6\pm0.7$ \\
 & GradCAM++ & $100.0\pm0.0$ & $12.7\pm0.9$ & $14.7\pm1.0$ \\
 & Grad$\times$Input & $53.2\pm0.3$ & $8.5\pm0.2$ & $0.9\pm0.1$ \\
 & HiResCAM & $100.0\pm0.0$ & $14.1\pm0.5$ & $15.6\pm0.7$ \\
 & LayerCAM & $100.0\pm0.0$ & $14.0\pm0.8$ & $15.4\pm0.7$ \\
\midrule
$\text{B-cos}$ & FinerCAM & $100.0\pm0.0$ & $10.6\pm0.5$ & $11.5\pm1.4$ \\
 & GradCAM & $100.0\pm0.0$ & $10.6\pm0.5$ & $11.5\pm1.4$ \\
 & GradCAM++ & $100.0\pm0.0$ & $10.2\pm0.7$ & $11.1\pm1.5$ \\
 & Grad$\times$Input & $52.6\pm0.2$ & $10.7\pm0.1$ & $0.8\pm0.1$ \\
 & HiResCAM & $100.0\pm0.0$ & $10.6\pm0.5$ & $11.5\pm1.4$ \\
 & LayerCAM & $100.0\pm0.0$ & $10.3\pm0.7$ & $11.1\pm1.5$ \\
 & Inherent B-cos & $95.4\pm0.6$ & $13.2\pm0.4$ & $1.6\pm0.1$ \\
\midrule
$\text{B-cos}_{\mathrm{BP}}$ & FinerCAM & $100.0\pm0.0$ & $10.7\pm0.8$ & $10.9\pm1.4$ \\
 & GradCAM & $100.0\pm0.0$ & $10.7\pm0.8$ & $10.9\pm1.4$ \\
 & GradCAM++ & $100.0\pm0.0$ & $10.7\pm1.0$ & $10.6\pm1.5$ \\
 & Grad$\times$Input & $52.6\pm0.2$ & $12.5\pm0.3$ & $1.4\pm0.1$ \\
 & HiResCAM & $100.0\pm0.0$ & $10.7\pm0.8$ & $10.9\pm1.4$ \\
 & LayerCAM & $100.0\pm0.0$ & $10.9\pm1.1$ & $10.6\pm1.5$ \\
 & Inherent B-cos & $90.7\pm1.0$ & $16.2\pm0.7$ & $4.6\pm0.4$ \\
\midrule
$\text{B-cos}_{\mathrm{ASAP}}$ & FinerCAM & $100.0\pm0.0$ & $9.1\pm1.2$ & $11.0\pm1.5$ \\
 & GradCAM & $100.0\pm0.0$ & $9.1\pm1.2$ & $11.0\pm1.5$ \\
 & GradCAM++ & $100.0\pm0.0$ & $9.1\pm1.3$ & $9.9\pm1.6$ \\
 & Grad$\times$Input & $53.3\pm0.5$ & $11.1\pm0.5$ & $1.0\pm0.1$ \\
 & HiResCAM & $100.0\pm0.0$ & $9.1\pm1.2$ & $11.0\pm1.5$ \\
 & LayerCAM & $100.0\pm0.0$ & $8.7\pm1.2$ & $10.2\pm1.6$ \\
 & Inherent B-cos & $93.1\pm1.0$ & $13.4\pm1.1$ & $2.8\pm0.4$ \\
\bottomrule
\end{tabular}\end{adjustbox}
\par\smallskip
\begin{minipage}{0.97\linewidth}\footnotesize
Values are percentages (mean $\pm$ sample SD). Inherent B-cos is undefined for Baseline. Non-negative CAMs can yield $\mathrm{EPG}_R=100\%$ without perfect spatial localization.
\end{minipage}
\end{table}
\clearpage
\begin{table}[!htbp]
\centering\small
\setlength{\tabcolsep}{8pt}
\renewcommand{\arraystretch}{1.06}
\caption{Explanation localization for VinBig, DenseNet-121, heavy Augmentation and No Sampling. All displayed rows use $n=15$ (three seeds and five folds).}\label{tab:explanation-vinbig_densenet121}
\begin{adjustbox}{max width=\linewidth}\begin{tabular}{clrrr@{}}
\toprule
Model & Explanation & $\mathrm{EPG}_R^{\rm micro}$ & $\mathrm{EPG}_P^{\rm micro}$ & mIoU \\
\midrule
Baseline & FinerCAM & $100.0\pm0.0$ & $5.3\pm0.2$ & $9.1\pm0.3$ \\
 & GradCAM & $100.0\pm0.0$ & $5.3\pm0.2$ & $9.1\pm0.3$ \\
 & GradCAM++ & $100.0\pm0.0$ & $5.5\pm0.1$ & $10.8\pm0.3$ \\
 & Grad$\times$Input & $50.8\pm0.1$ & $6.3\pm0.1$ & $0.4\pm0.1$ \\
 & HiResCAM & $100.0\pm0.0$ & $6.1\pm0.2$ & $9.8\pm0.3$ \\
 & LayerCAM & $100.0\pm0.0$ & $5.6\pm0.1$ & $10.9\pm0.3$ \\
\midrule
$\text{B-cos}$ & FinerCAM & $100.0\pm0.0$ & $6.6\pm0.2$ & $12.5\pm0.3$ \\
 & GradCAM & $100.0\pm0.0$ & $6.6\pm0.2$ & $12.5\pm0.3$ \\
 & GradCAM++ & $100.0\pm0.0$ & $6.8\pm0.6$ & $12.1\pm0.6$ \\
 & Grad$\times$Input & $52.7\pm0.2$ & $6.5\pm0.2$ & $0.6\pm0.0$ \\
 & HiResCAM & $100.0\pm0.0$ & $7.3\pm0.3$ & $12.9\pm0.3$ \\
 & LayerCAM & $100.0\pm0.0$ & $7.2\pm0.5$ & $12.2\pm0.5$ \\
 & Inherent B-cos & $99.2\pm0.2$ & $6.9\pm0.2$ & $1.9\pm0.4$ \\
\midrule
$\text{B-cos}_{\mathrm{BP}}$ & FinerCAM & $100.0\pm0.0$ & $6.6\pm0.3$ & $12.4\pm0.5$ \\
 & GradCAM & $100.0\pm0.0$ & $6.6\pm0.3$ & $12.4\pm0.5$ \\
 & GradCAM++ & $100.0\pm0.0$ & $6.6\pm0.4$ & $11.9\pm0.9$ \\
 & Grad$\times$Input & $54.2\pm0.3$ & $6.5\pm0.2$ & $1.0\pm0.1$ \\
 & HiResCAM & $100.0\pm0.0$ & $7.3\pm0.3$ & $12.7\pm0.6$ \\
 & LayerCAM & $100.0\pm0.0$ & $7.1\pm0.4$ & $12.1\pm1.0$ \\
 & Inherent B-cos & $99.2\pm0.2$ & $7.1\pm0.4$ & $4.3\pm0.5$ \\
\midrule
$\text{B-cos}_{\mathrm{ASAP}}$ & FinerCAM & $100.0\pm0.0$ & $6.6\pm0.3$ & $12.5\pm0.5$ \\
 & GradCAM & $100.0\pm0.0$ & $6.6\pm0.3$ & $12.5\pm0.5$ \\
 & GradCAM++ & $100.0\pm0.0$ & $6.6\pm0.5$ & $12.0\pm0.9$ \\
 & Grad$\times$Input & $54.0\pm0.3$ & $6.5\pm0.2$ & $0.9\pm0.0$ \\
 & HiResCAM & $100.0\pm0.0$ & $7.3\pm0.3$ & $12.8\pm0.6$ \\
 & LayerCAM & $100.0\pm0.0$ & $7.0\pm0.5$ & $12.2\pm1.0$ \\
 & Inherent B-cos & $99.1\pm0.2$ & $7.2\pm0.4$ & $3.1\pm0.5$ \\
\bottomrule
\end{tabular}\end{adjustbox}
\par\smallskip
\begin{minipage}{0.97\linewidth}\footnotesize
Values are percentages (mean $\pm$ sample SD). Inherent B-cos is undefined for Baseline. Non-negative CAMs can yield $\mathrm{EPG}_R=100\%$ without perfect spatial localization.
\end{minipage}
\end{table}
\clearpage
\section{ImageNet Classification from Random Initialization}
The following comparisons are for ImageNet~\cite{imagenet} Classification from Random Initialization.
All use Adam with learning rate $10^{-4}$ and zero weight decay, cosine
annealing for 90 epochs, batch size 64 per GPU (declared effective batch size
256), and adaptive gradient clipping. F1, precision, and recall are macro
averages over 1,000 classes; AUROC is macro one-vs-rest. The results are for one random seed `5''. 
Pretrained and partial weight-transfer variants are excluded.
\begin{table}[!htbp]\centering\footnotesize
\setlength{\tabcolsep}{7pt}
\caption{ImageNet classification from random initialization. All values are percentages from the single observed seed-5 run, after 90 epochs of training.}
\label{tab:imagenet-random}
\begin{adjustbox}{max width=\linewidth}\begin{tabular}{clrrrrr@{}}\toprule
Architecture & Model & Top-1 & AUROC & F1 & Precision & Recall \\\midrule
ResNet50 & Baseline & 70.4 & 99.7 & 70.1 & 70.7 & 70.4 \\
 & $\text{B-cos}$ & 72.2 & 99.0 & 71.8 & 72.4 & 72.2 \\
 & $\text{B-cos}_{\mathrm{BP}}$ & 71.3 & 98.8 & 70.9 & 71.5 & 71.3 \\
 & $\text{B-cos}_{\mathrm{ASAP}}$ & 70.7 & 98.8 & 70.3 & 70.9 & 70.7 \\
\midrule
DenseNet-121 & Baseline & 67.8 & 99.6 & 67.4 & 68.0 & 67.8 \\
 & $\text{B-cos}$ & 66.8 & 99.1 & 66.2 & 66.8 & 66.8 \\
 & $\text{B-cos}_{\mathrm{BP}}$ & 65.8 & 99.1 & 65.2 & 65.8 & 65.8 \\
 & $\text{B-cos}_{\mathrm{ASAP}}$ & 65.8 & 99.0 & 65.2 & 65.8 & 65.8 \\
\bottomrule\end{tabular}\end{adjustbox}
\end{table}

\clearpage
\section{Training Configurations}\label{sec:training-configurations}
In the following we report the training configuration behind every quantitative
table so that each reported score can be traced to its exact recipe.
Table~\ref{tab:training-configurations} lists the optimizer, learning rate,
batch size, warm-up length and loss of every model. All medical experiments
share the following settings. Images are resized to $224\times224$ pixels, and
B-cos models receive the six-channel encoding $[x,\,1-x]$ of the input.
Backbones start from ImageNet-pretrained weights; the B-cos variants are
B-cosified from the same pretrained backbone with $B=2$, without bias terms
and with uncentred batch normalization, and $\text{B-cos}_{\mathrm{BP}}$ and
$\text{B-cos}_{\mathrm{ASAP}}$ replace the stride-two downsampling operations
as described in the main paper. Training runs for 30 epochs with cosine
annealing, a linear warm-up whose initial factor is $0.01$, and adaptive
gradient clipping; the final checkpoint (epoch 29) is evaluated. Every
configuration is trained for seeds 5, 420 and 1337 on folds 0-4, and the
tables report the equally weighted mean and sample standard deviation of these
15 runs. Validation and test images are only resized.
\begin{table}[!htbp]
\centering\footnotesize
\setlength{\tabcolsep}{3.5pt}
\renewcommand{\arraystretch}{1.12}
\caption{Training configurations of the reported models. WD is the weight-decay
coefficient, warm-up is the number of linear warm-up epochs of the cosine
schedule, and the loss column names the training criterion. Medical models
train for 30 epochs; ImageNet models train for 90 epochs. UOL-BCE denotes binary cross-entropy with logits and uniform off-label targets; validation losses use plain binary cross-entropy. CE is cross-entropy. The ImageNet batch is 64 per GPU on four GPUs. $^{\ast}$The archived RSNA ResNet50 configuration sets five warm-up epochs and carries an inline note of an earlier value of three.}\label{tab:training-configurations}
\begin{adjustbox}{max width=\linewidth}\begin{tabular}{cllcccclp{0.27\linewidth}@{}}
\toprule
Dataset & Architecture & Init. & Optimizer & WD & LR & Batch & Warm-up & Loss / reported setting \\
\midrule
RSNA & ResNet50 & pretrained & Adam & 0 & $10^{-4}$ & 16 & 5$^{\ast}$ & UOL-BCE; six settings in Table~\ref{tab:pneumonia-combined}, main paper: heavy / Oversampling \\
RSNA & ConvNeXt-base & pretrained & Adam & 0 & $10^{-3}$ & 16 & 5 & UOL-BCE; heavy / Oversampling \\
VinBig & ResNet50 & pretrained & Adam & 0 & $10^{-4}$ & 32 & 3 & UOL-BCE; six settings in Table~\ref{tab:vinbigxray-combined}, main paper: light / No Sampling \\
VinBig & ConvNeXt-tiny & pretrained & AdamW & 0.01 & $10^{-3}$ & 16 & 5 & UOL-BCE; light / No Sampling \\
VinBig & ConvNeXt-base & pretrained & Adam & 0 & $10^{-3}$ & 16 & 3 & UOL-BCE; light / No Sampling \\
VinBig & DenseNet-121 & pretrained & Adam & 0 & $10^{-4}$ & 64 & 3 & UOL-BCE; heavy / No Sampling \\
ImageNet & ResNet50, DenseNet-121 & random & Adam & 0 & $10^{-4}$ & $64\times4$ & 0 & UOL-BCE (B-cos), CE (Baseline); random-resized crop and flip, seed 5 \\
\bottomrule
\end{tabular}\end{adjustbox}
\end{table}

\subsection{Augmentation Pipelines}
The three Augmentation levels are identical for both chest X-ray datasets and
are applied on the fly to the resized $224\times224$ training image.
\emph{None} applies no transformation beyond resizing.
\emph{Light} applies a random translation of up to $\pm32$ pixels along each
axis, an isotropic rescaling by the factor $2^{\mathcal{N}(0,0.1)}$, a rotation
of $\mathcal{N}(0,5^{\circ})$ combined with a shear of $\mathcal{N}(0,2.5^{\circ})$,
a random perspective distortion of scale $0.1$ with probability $0.5$, and a
gamma adjustment by the factor $2^{\mathcal{N}(0,0.2)}$.
\emph{Heavy} applies a horizontal flip with probability $0.5$, the same
translation, a rescaling by $2^{\mathcal{N}(0,0.15)}$, a rotation of
$\mathcal{N}(0,6^{\circ})$ with a shear of $\mathcal{N}(0,4^{\circ})$, a
perspective distortion of scale $0.15$ with probability $0.5$, a gamma
adjustment by $2^{\mathcal{N}(0,0.25)}$, and then, independently, coarse
salt-and-pepper noise with probability $0.1$ (one percent of pixels each, on
patches covering 10-20\% of the image), Gaussian blur with probability $0.5$
and $\sigma\sim\mathcal{U}(0,2)$, and additive Gaussian pixel noise with
probability $0.5$ and $\sigma\sim\mathcal{U}(0,\,0.04\cdot255)$.

\subsection{Sampling}
Without sampling, training batches are drawn by uniform shuffling. With
Oversampling, batches are drawn with replacement by a weighted random sampler
over one training-set length per epoch. On RSNA Pneumonia the weight of an image
is the inverse frequency of its class. On VinBig each disease label receives the
weight $1/(\text{count}+\varepsilon)$, clipped to $[1/3,\,3]$; the weight of an
image is the sum of the weights of its positive labels, normalized by the
maximum over the training set.

\clearpage
\section{Qualitative explanation examples}
In the following, we provide more qualitative examples across architectures and datasets. 

\subsection{RSNA Pneumonia: ResNet50}

\begin{figure}[!ht]

\centering

\includegraphics[width=\linewidth,height=0.79\textheight,keepaspectratio]{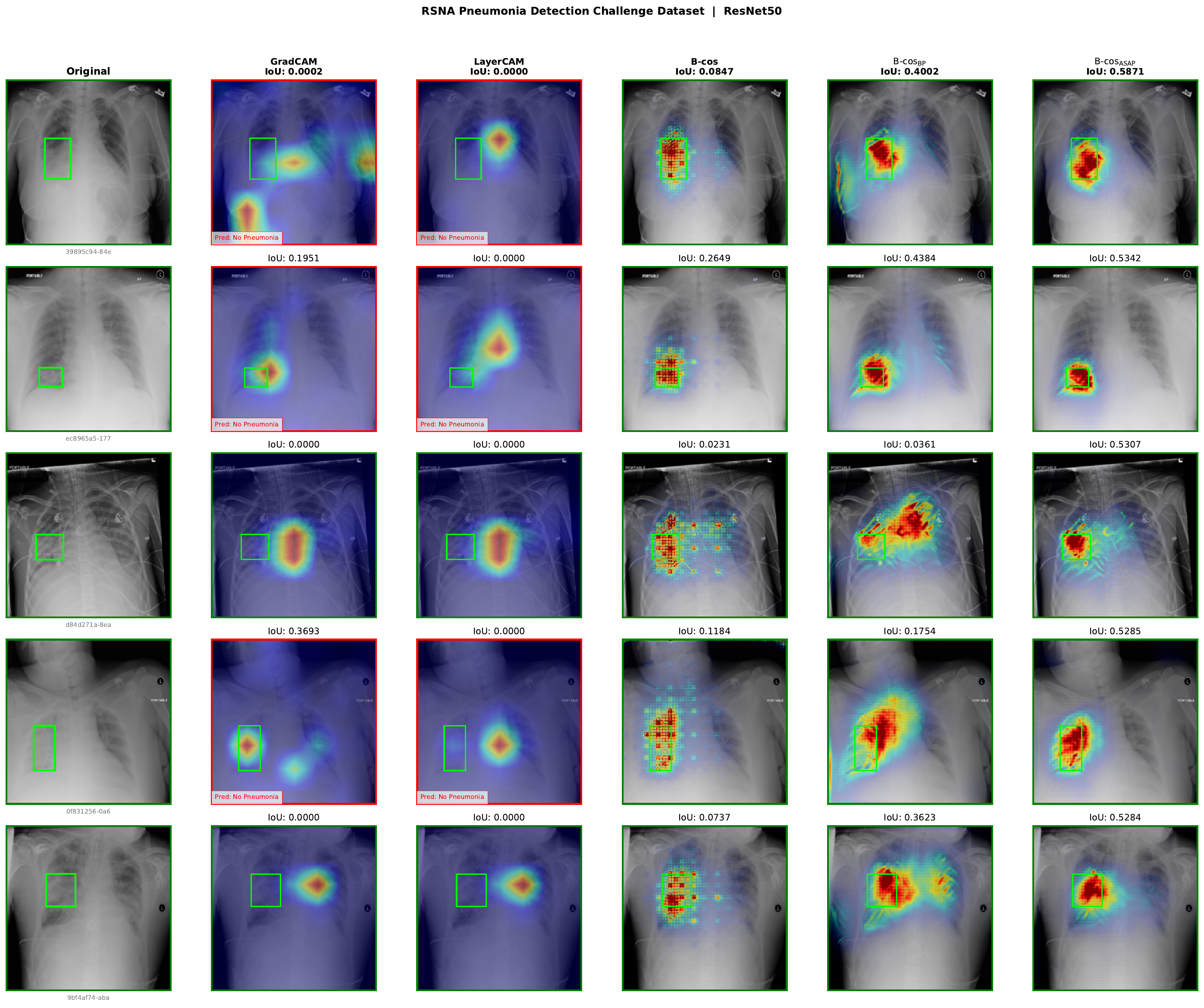}

\caption{RSNA Pneumonia, ResNet50. Five illustrative cases using heavy augmentation and oversampling (fold 3, seed 420).}

\label{fig:final-1}

\end{figure}

\clearpage

\subsection{RSNA Pneumonia: ConvNeXt-base}

\begin{figure}[!ht]

\centering

\includegraphics[width=\linewidth,height=0.79\textheight,keepaspectratio]{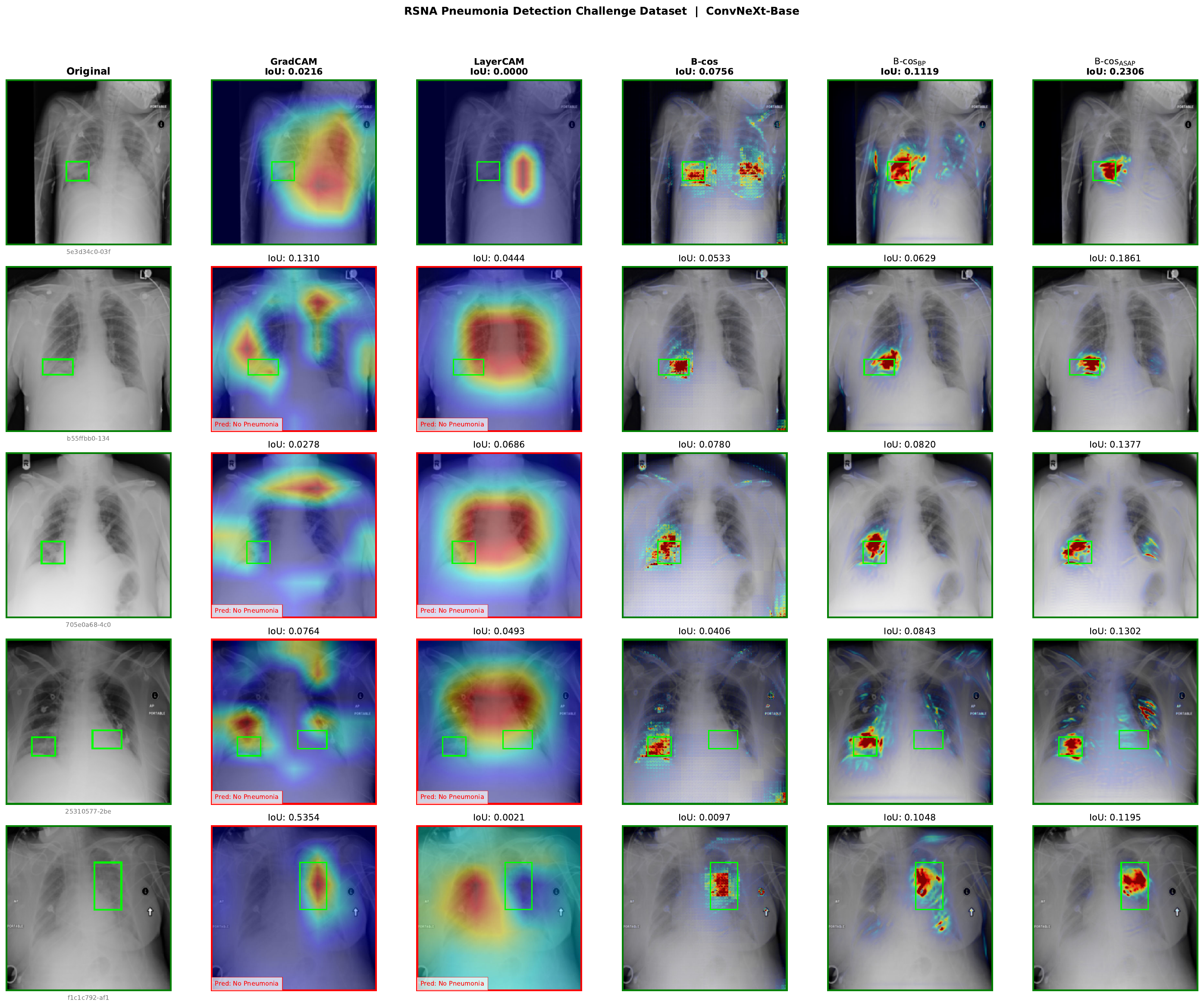}

\caption{RSNA Pneumonia, ConvNeXt-base. Five illustrative cases using heavy augmentation and oversampling (fold 0, seed 1337).}

\label{fig:final-2}

\end{figure}

\clearpage

\subsection{VinBig: ResNet50}

\begin{figure}[!ht]

\centering

\includegraphics[width=\linewidth,height=0.79\textheight,keepaspectratio]{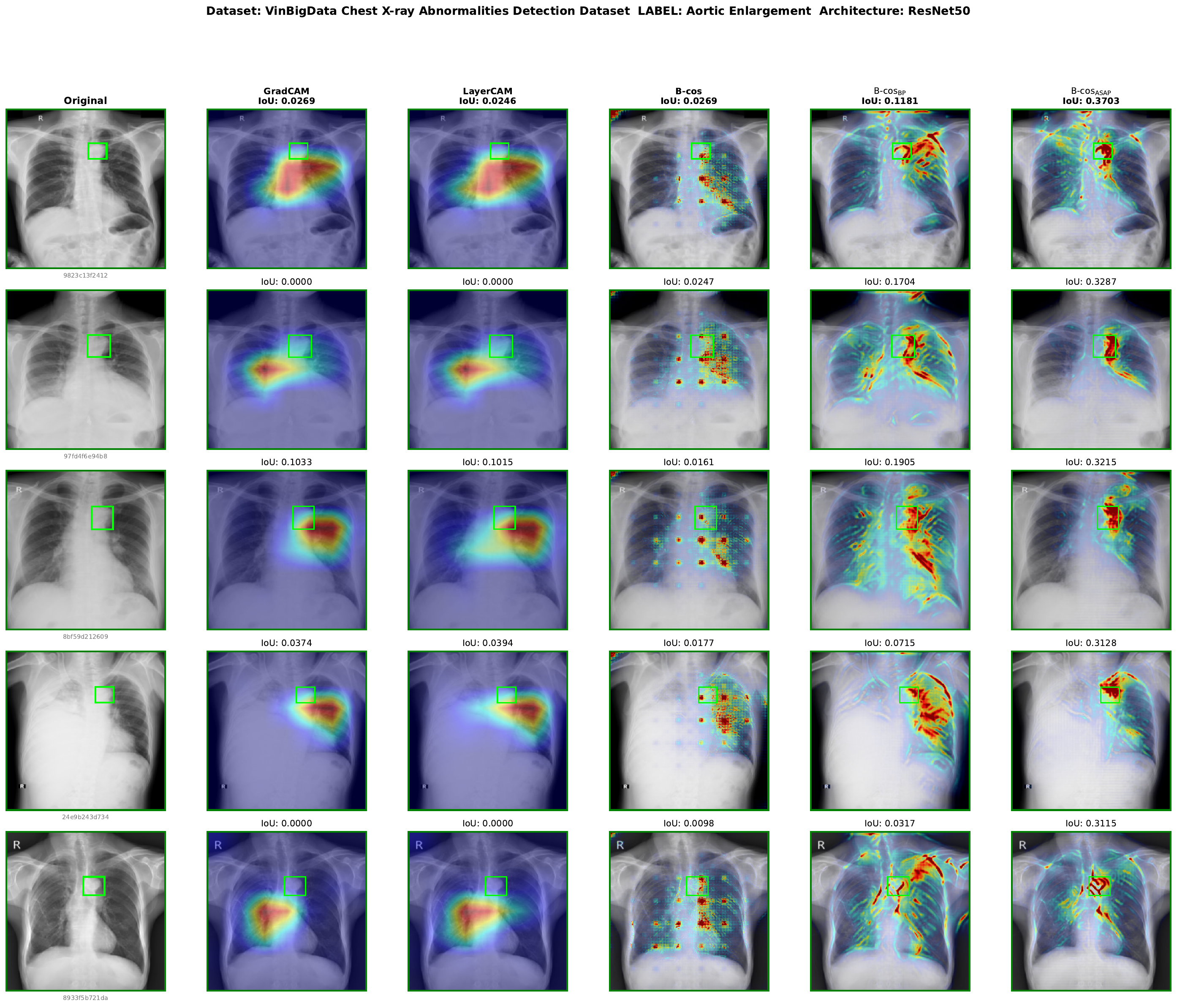}

\caption{VinBig: Aortic Enlargement, ResNet50. Five illustrative cases using light augmentation without sampling (fold 4, seed 1337).}

\label{fig:final-3}

\end{figure}

\clearpage

\begin{figure}[!ht]

\centering

\includegraphics[width=\linewidth,height=0.79\textheight,keepaspectratio]{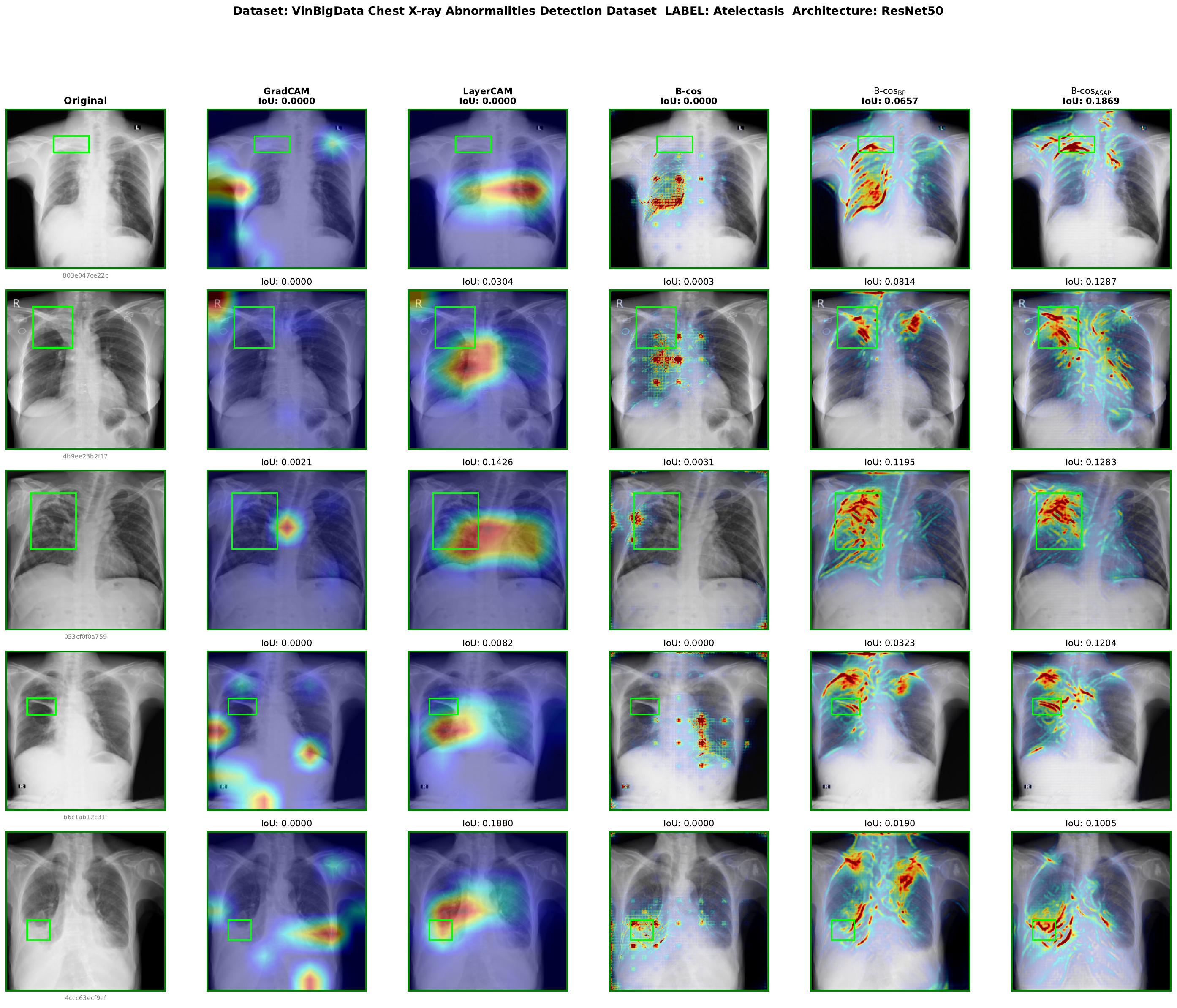}

\caption{VinBig: Atelectasis, ResNet50. Five illustrative cases using light augmentation without sampling (fold 4, seed 1337).}

\label{fig:final-4}

\end{figure}

\clearpage

\begin{figure}[!ht]

\centering

\includegraphics[width=\linewidth,height=0.79\textheight,keepaspectratio]{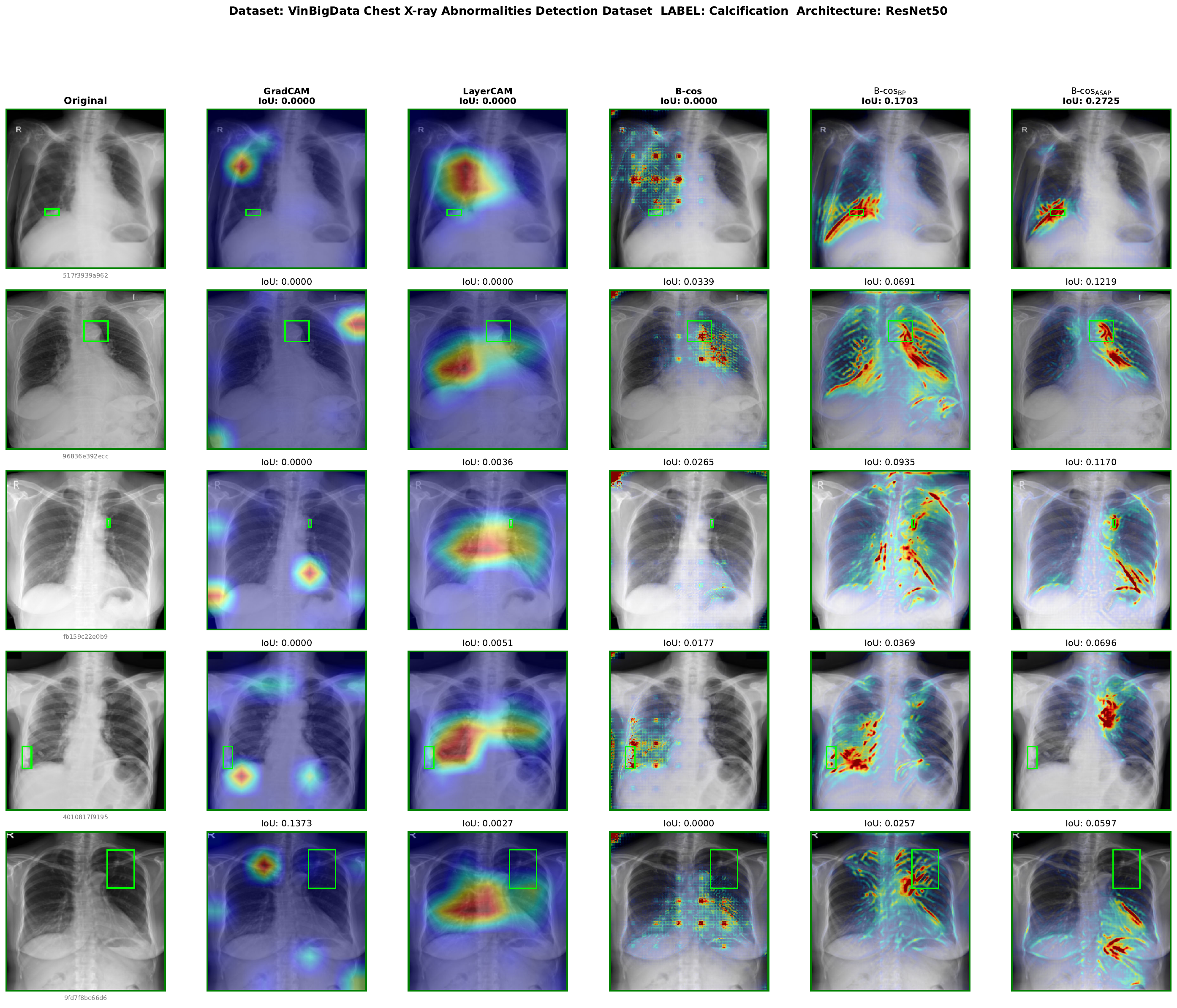}

\caption{VinBig: Calcification, ResNet50. Five illustrative cases using light augmentation without sampling (fold 4, seed 1337).}

\label{fig:final-5}

\end{figure}

\clearpage

\begin{figure}[!ht]

\centering

\includegraphics[width=\linewidth,height=0.79\textheight,keepaspectratio]{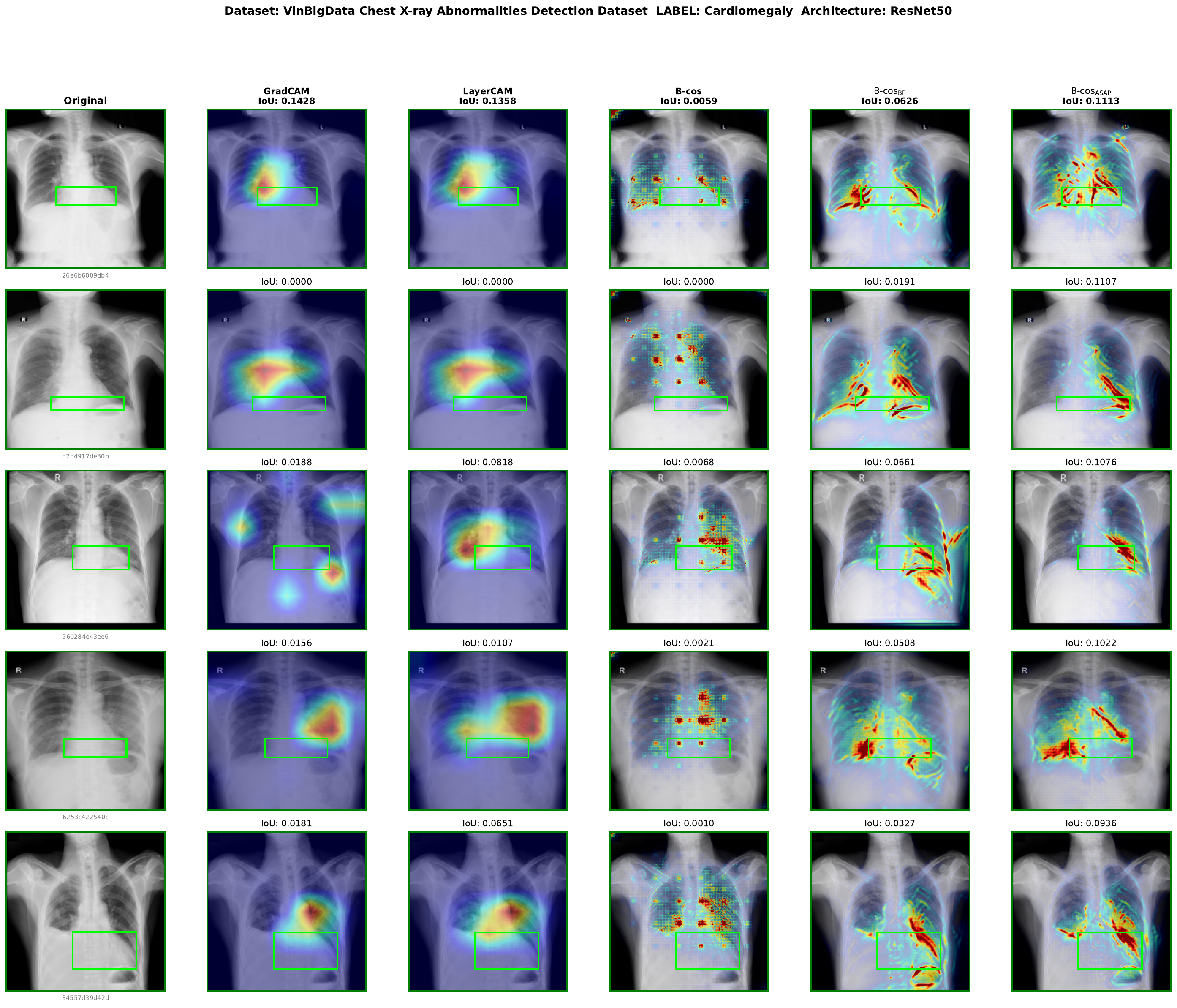}

\caption{VinBig: Cardiomegaly, ResNet50. Five illustrative cases using light augmentation without sampling (fold 4, seed 1337).}

\label{fig:final-6}

\end{figure}

\clearpage

\begin{figure}[!ht]

\centering

\includegraphics[width=\linewidth,height=0.79\textheight,keepaspectratio]{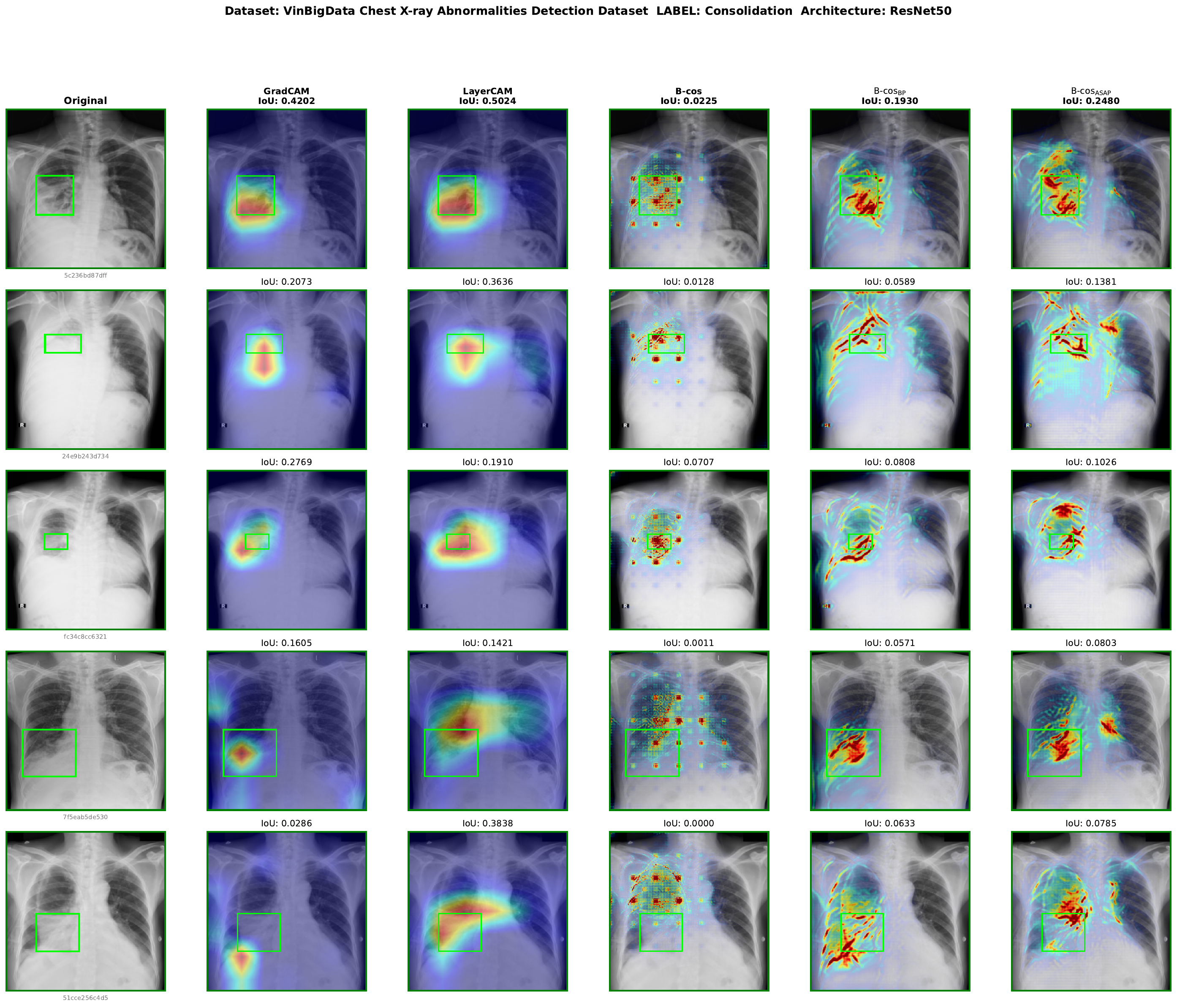}

\caption{VinBig: Consolidation, ResNet50. Five illustrative cases using light augmentation without sampling (fold 4, seed 1337).}

\label{fig:final-7}

\end{figure}

\clearpage

\begin{figure}[!ht]

\centering

\includegraphics[width=\linewidth,height=0.79\textheight,keepaspectratio]{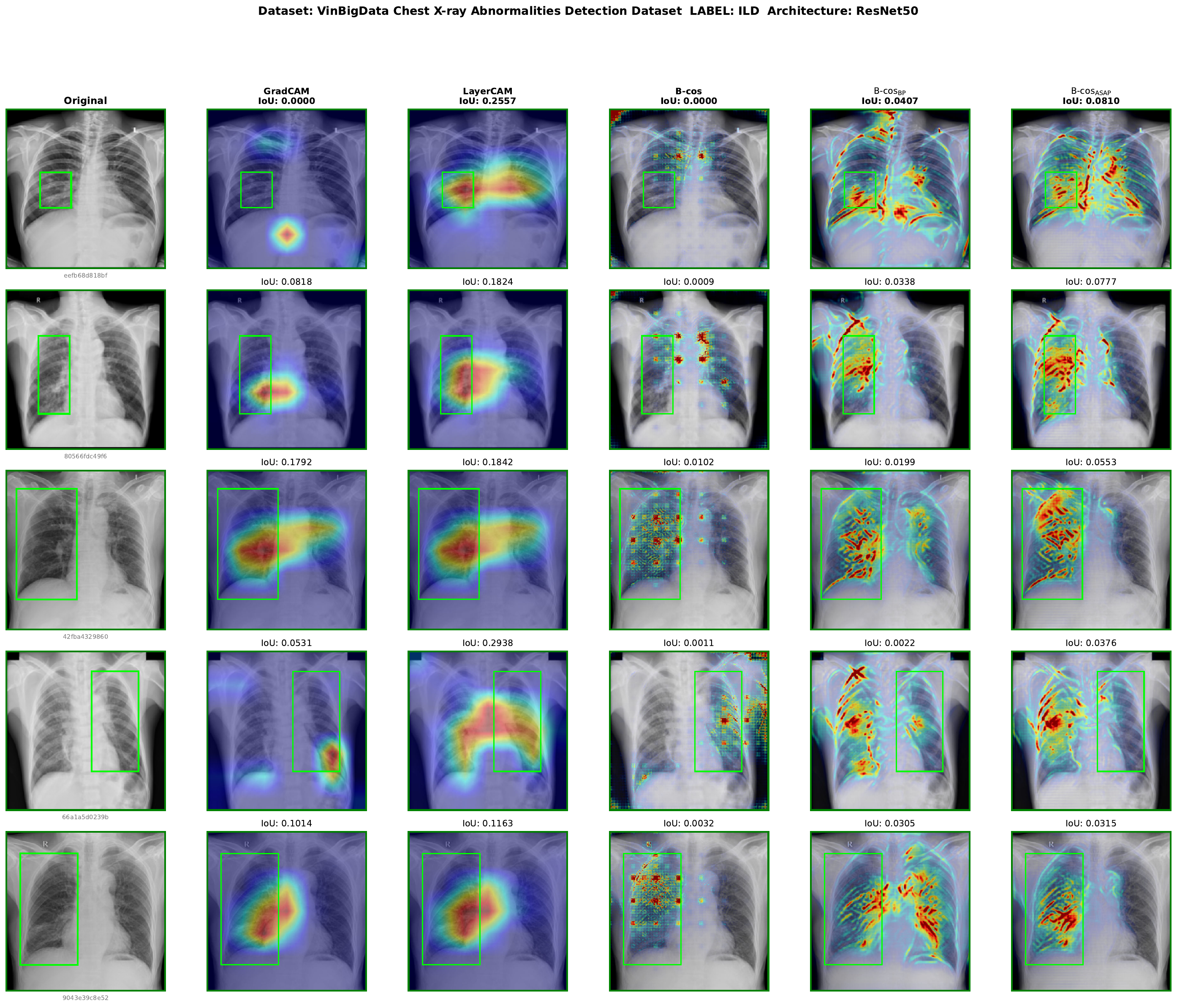}

\caption{VinBig: ILD, ResNet50. Five illustrative cases using light augmentation without sampling (fold 4, seed 1337).}

\label{fig:final-8}

\end{figure}

\clearpage

\begin{figure}[!ht]

\centering

\includegraphics[width=\linewidth,height=0.79\textheight,keepaspectratio]{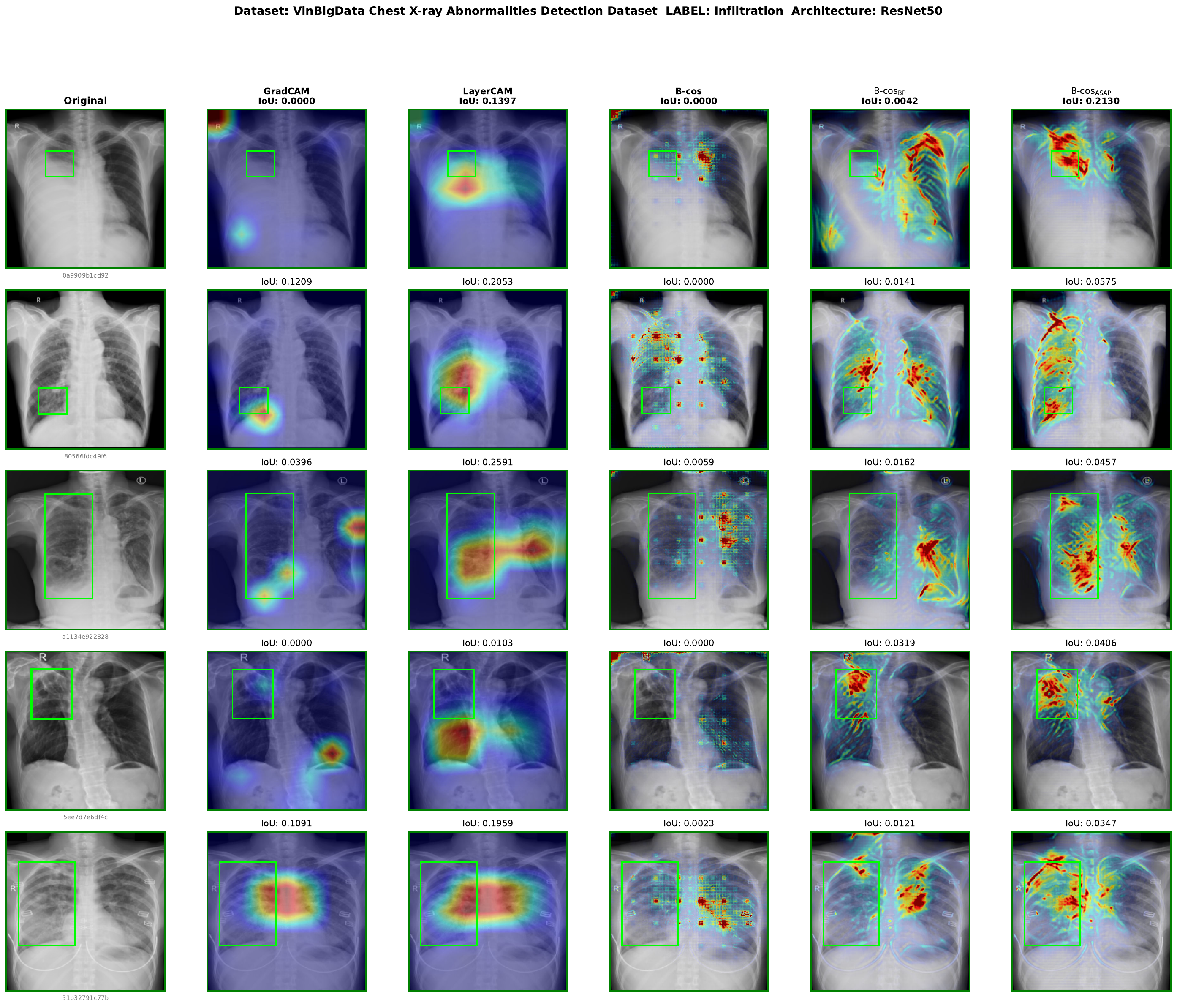}

\caption{VinBig: Infiltration, ResNet50. Five illustrative cases using light augmentation without sampling (fold 4, seed 1337).}

\label{fig:final-9}

\end{figure}

\clearpage

\begin{figure}[!ht]

\centering

\includegraphics[width=\linewidth,height=0.79\textheight,keepaspectratio]{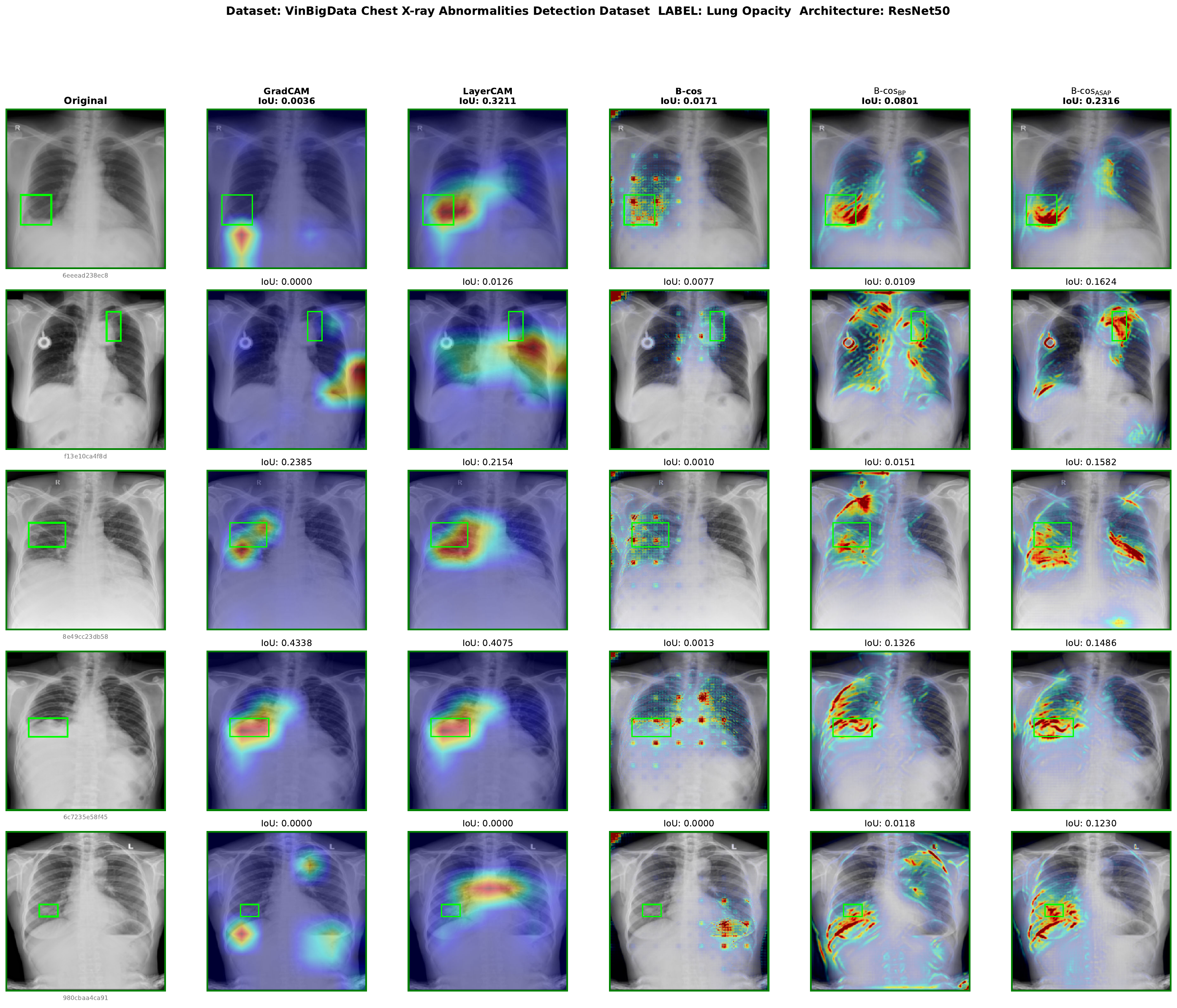}

\caption{VinBig: Lung Opacity, ResNet50. Five illustrative cases using light augmentation without sampling (fold 4, seed 1337).}

\label{fig:final-10}

\end{figure}

\clearpage

\begin{figure}[!ht]

\centering

\includegraphics[width=\linewidth,height=0.79\textheight,keepaspectratio]{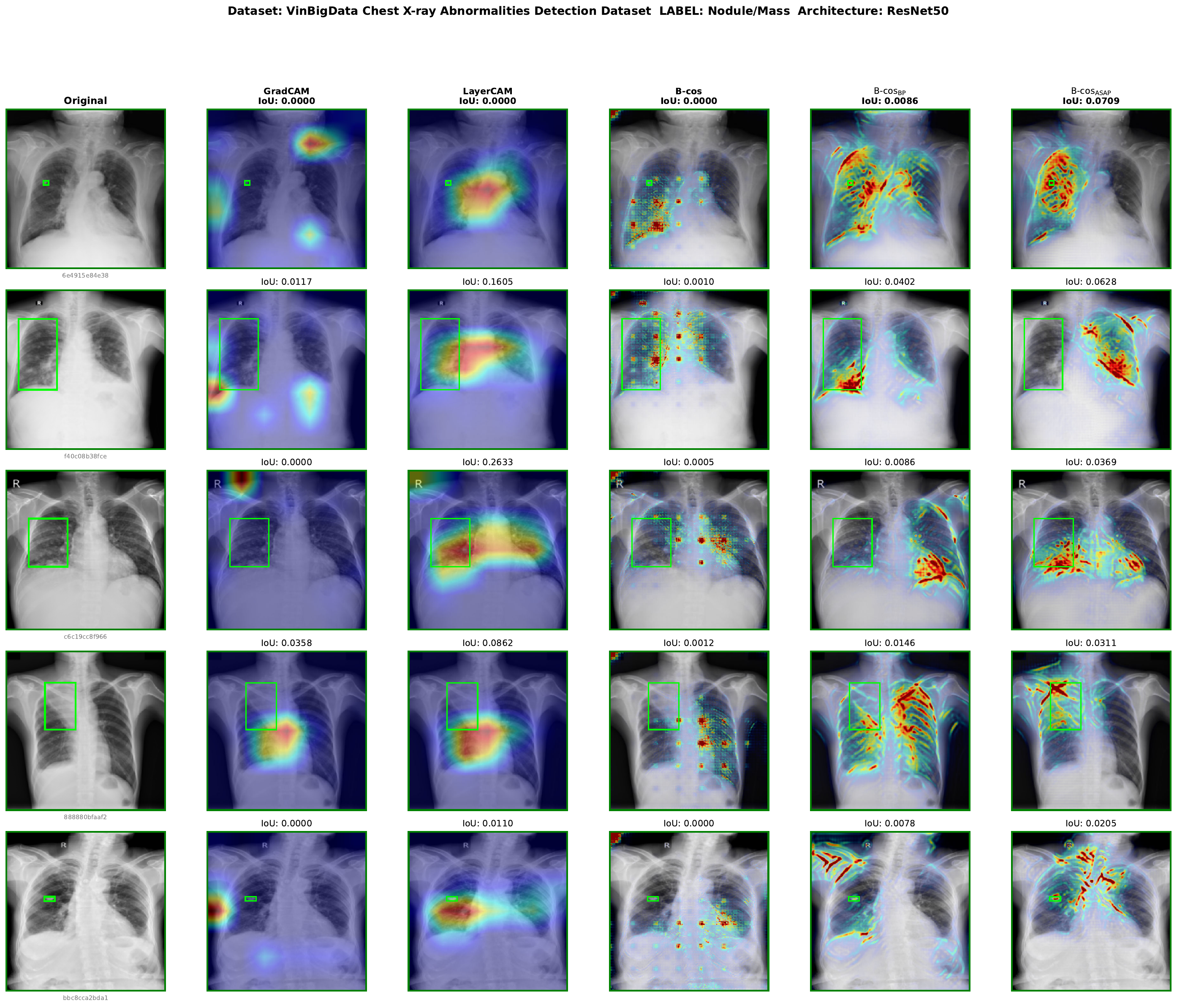}

\caption{VinBig: Nodule/Mass, ResNet50. Five illustrative cases using light augmentation without sampling (fold 4, seed 1337).}

\label{fig:final-11}

\end{figure}

\clearpage

\begin{figure}[!ht]

\centering

\includegraphics[width=\linewidth,height=0.79\textheight,keepaspectratio]{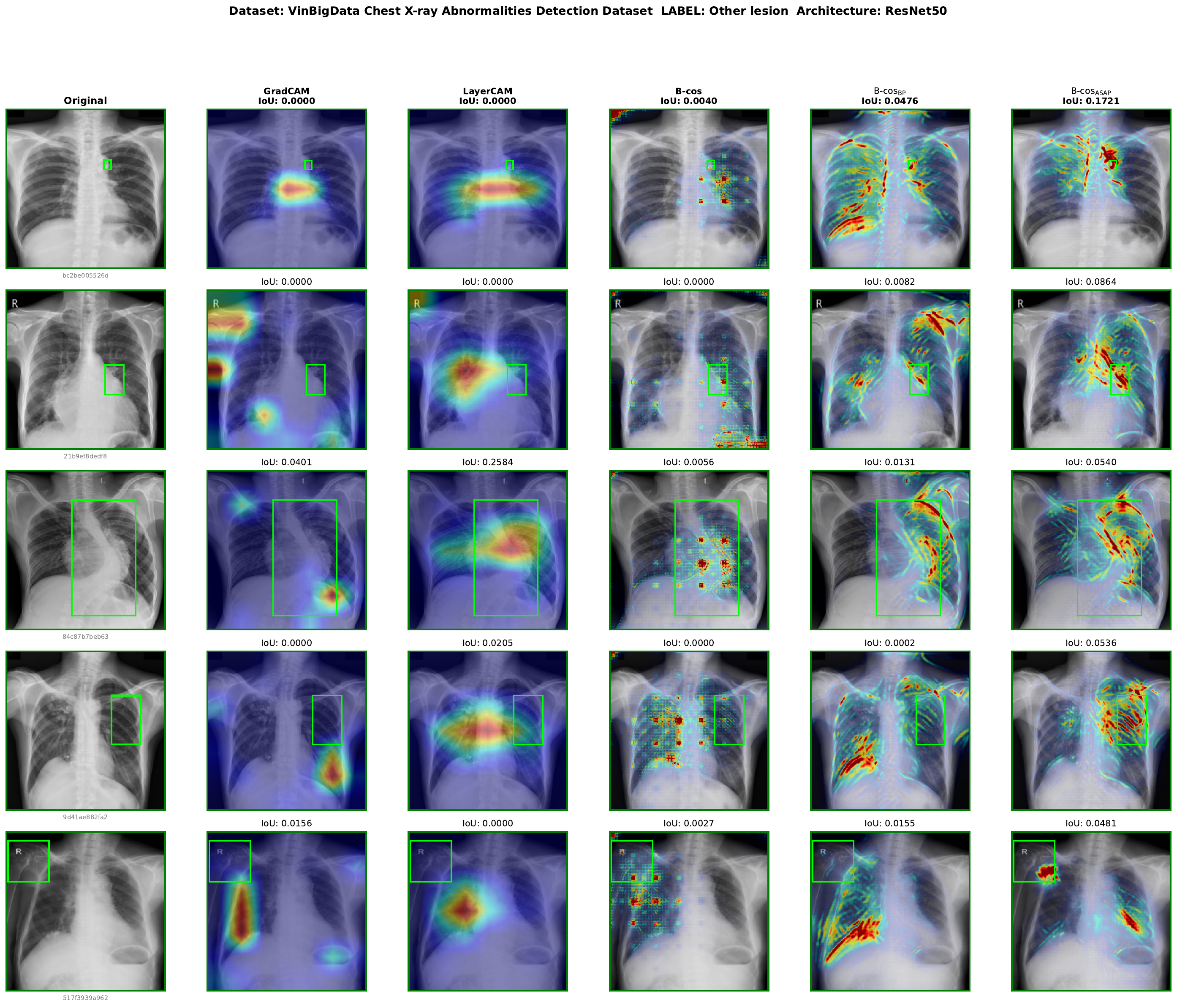}

\caption{VinBig: Other lesion, ResNet50. Five illustrative cases using light augmentation without sampling (fold 4, seed 1337).}

\label{fig:final-12}

\end{figure}

\clearpage

\begin{figure}[!ht]

\centering

\includegraphics[width=\linewidth,height=0.79\textheight,keepaspectratio]{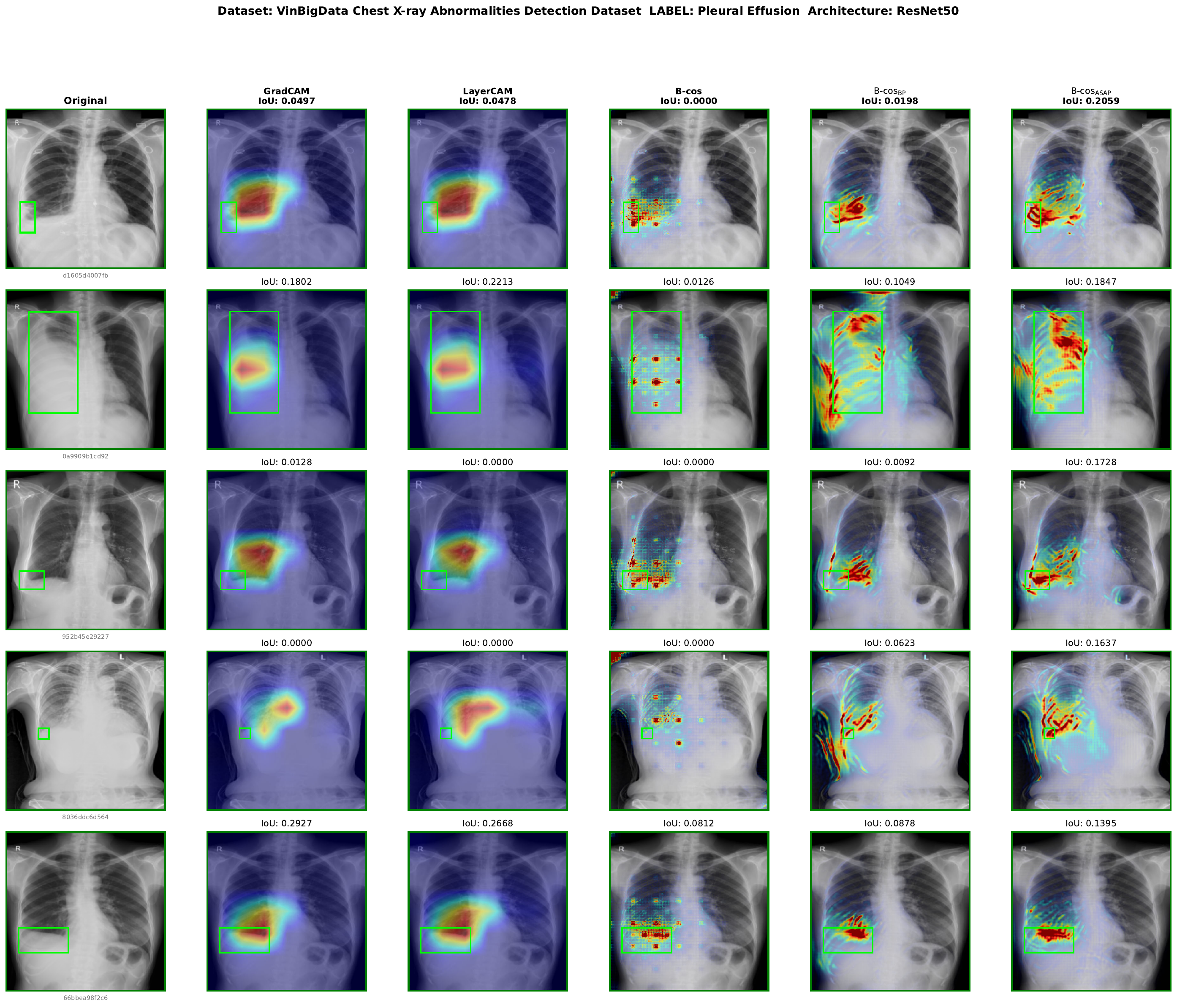}

\caption{VinBig: Pleural Effusion, ResNet50. Five illustrative cases using light augmentation without sampling (fold 4, seed 1337).}

\label{fig:final-13}

\end{figure}

\clearpage

\begin{figure}[!ht]

\centering

\includegraphics[width=\linewidth,height=0.79\textheight,keepaspectratio]{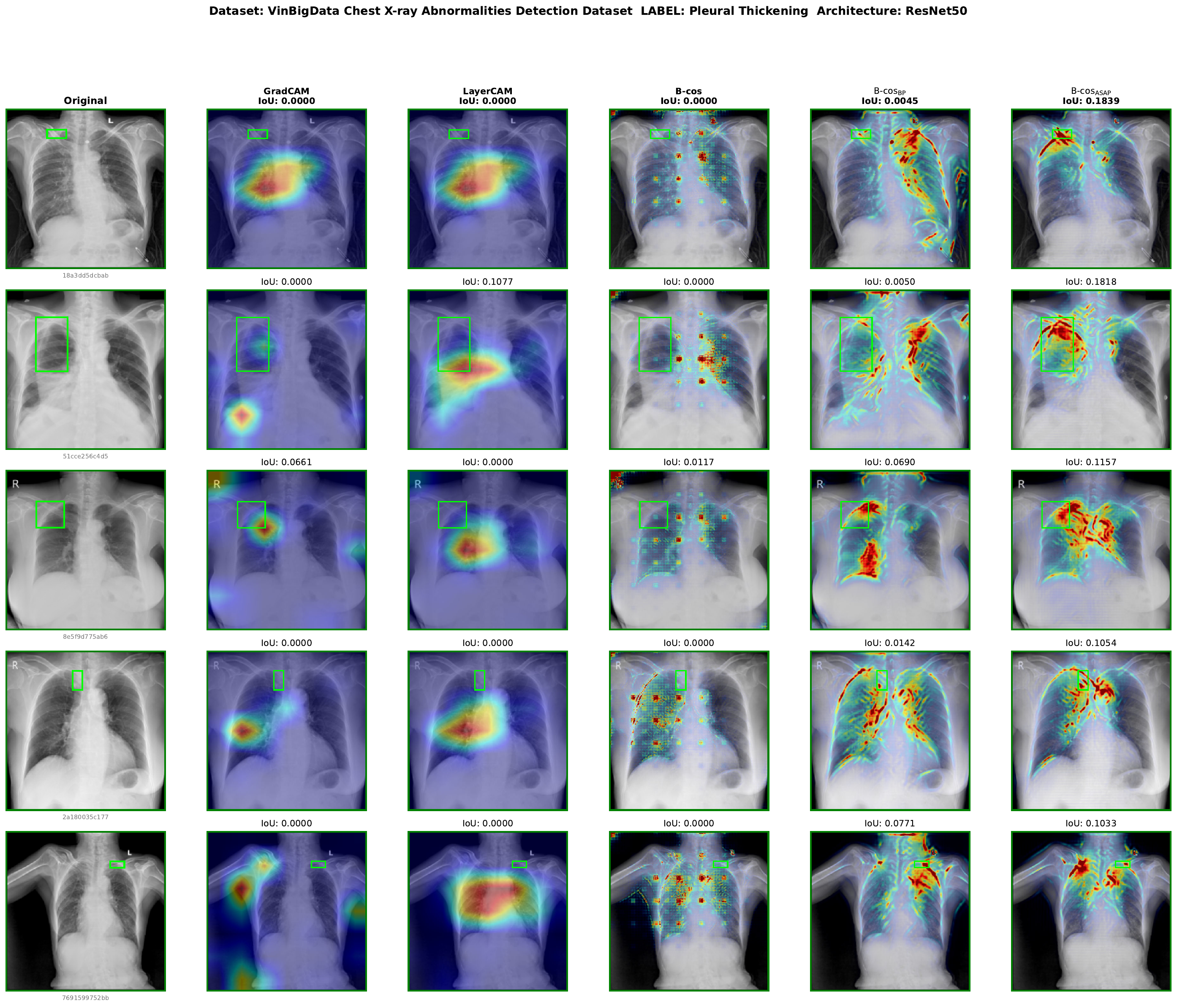}

\caption{VinBig: Pleural Thickening, ResNet50. Five illustrative cases using light augmentation without sampling (fold 4, seed 1337).}

\label{fig:final-14}

\end{figure}

\clearpage

\begin{figure}[!ht]

\centering

\includegraphics[width=\linewidth,height=0.79\textheight,keepaspectratio]{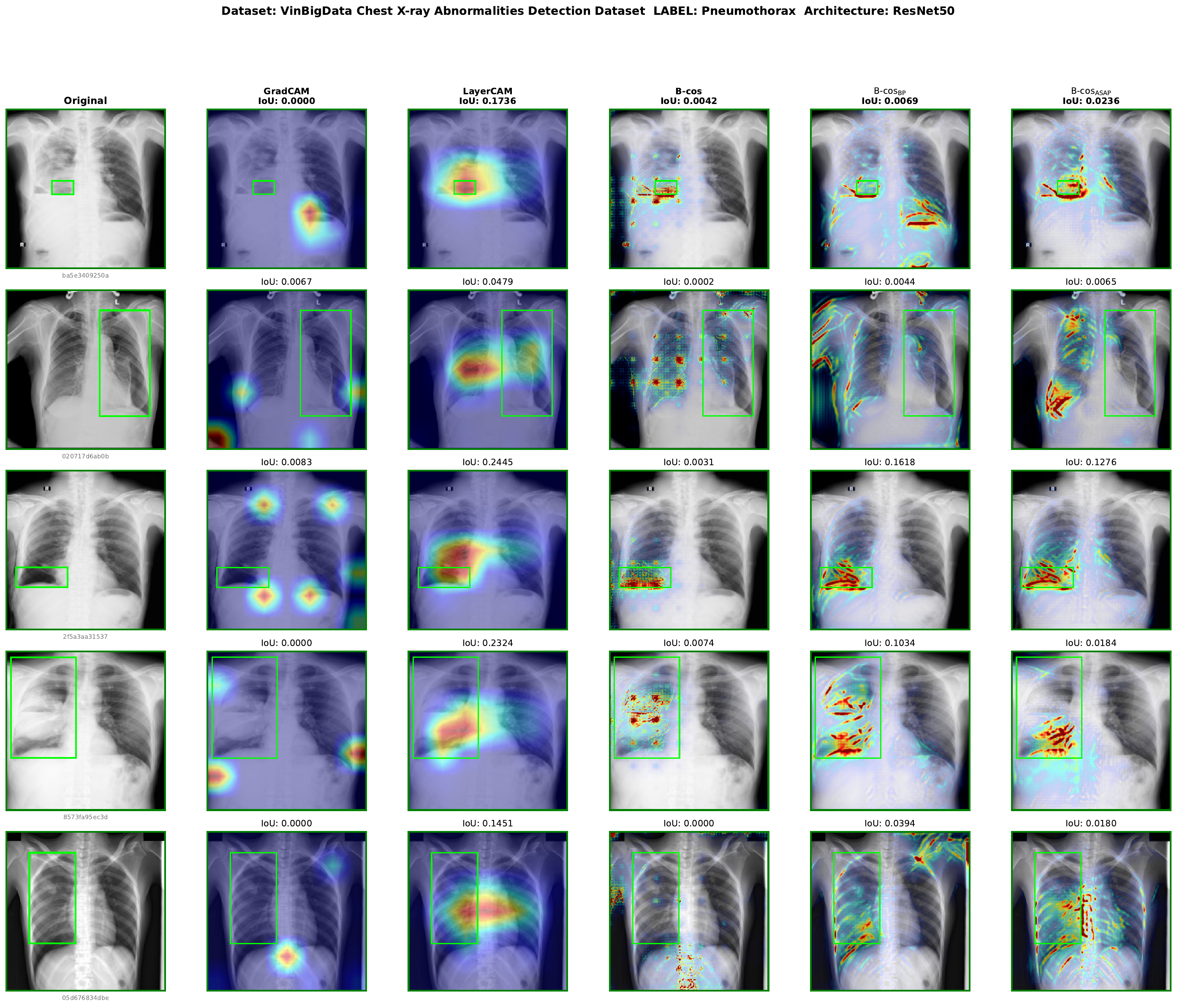}

\caption{VinBig: Pneumothorax, ResNet50. Five illustrative cases using light augmentation without sampling (fold 4, seed 1337).}

\label{fig:final-15}

\end{figure}

\clearpage

\begin{figure}[!ht]

\centering

\includegraphics[width=\linewidth,height=0.79\textheight,keepaspectratio]{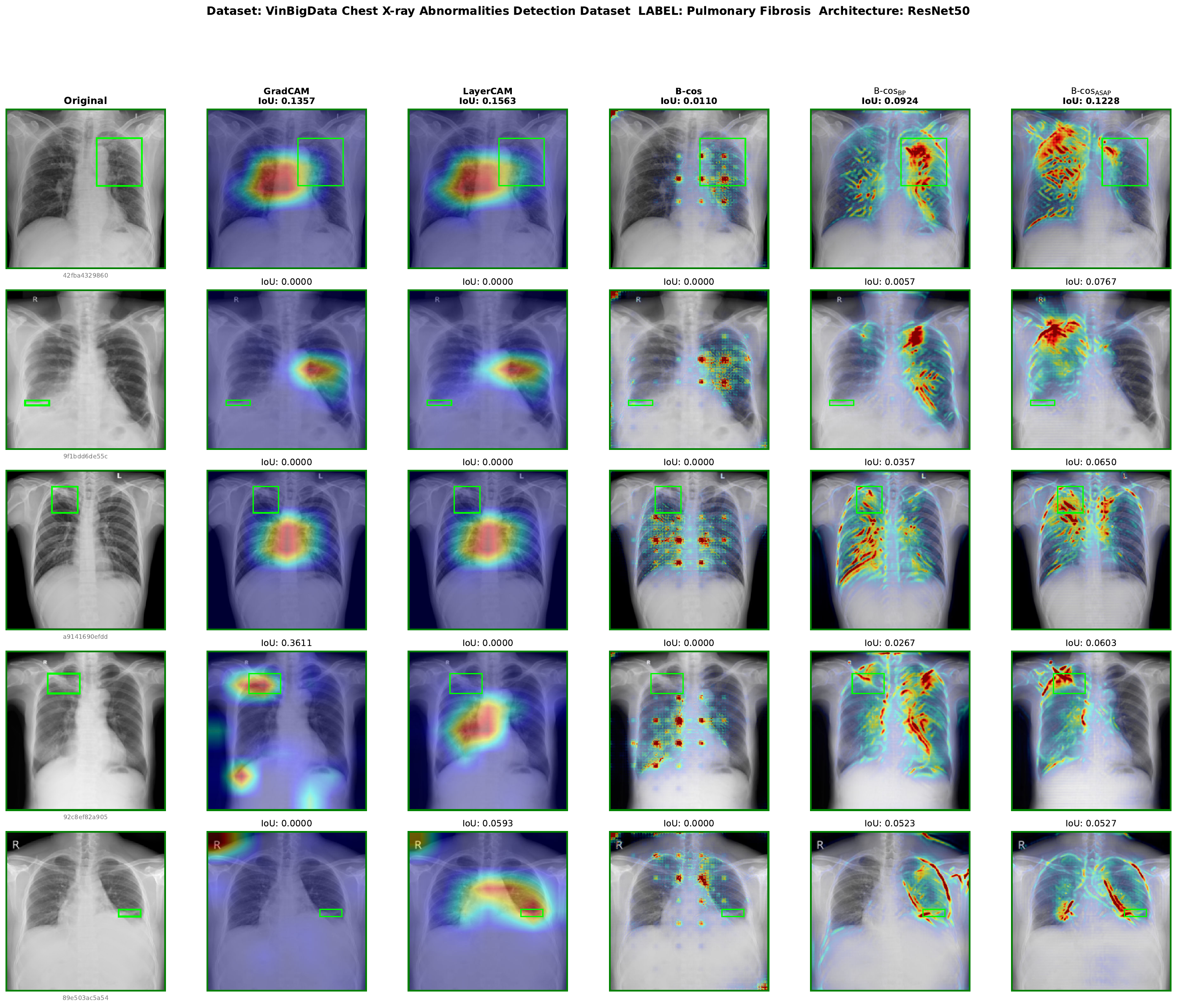}

\caption{VinBig: Pulmonary Fibrosis, ResNet50. Five illustrative cases using light augmentation without sampling (fold 4, seed 1337).}

\label{fig:final-16}

\end{figure}

\clearpage

\subsection{VinBig: ConvNeXt-base}

\begin{figure}[!ht]

\centering

\includegraphics[width=\linewidth,height=0.79\textheight,keepaspectratio]{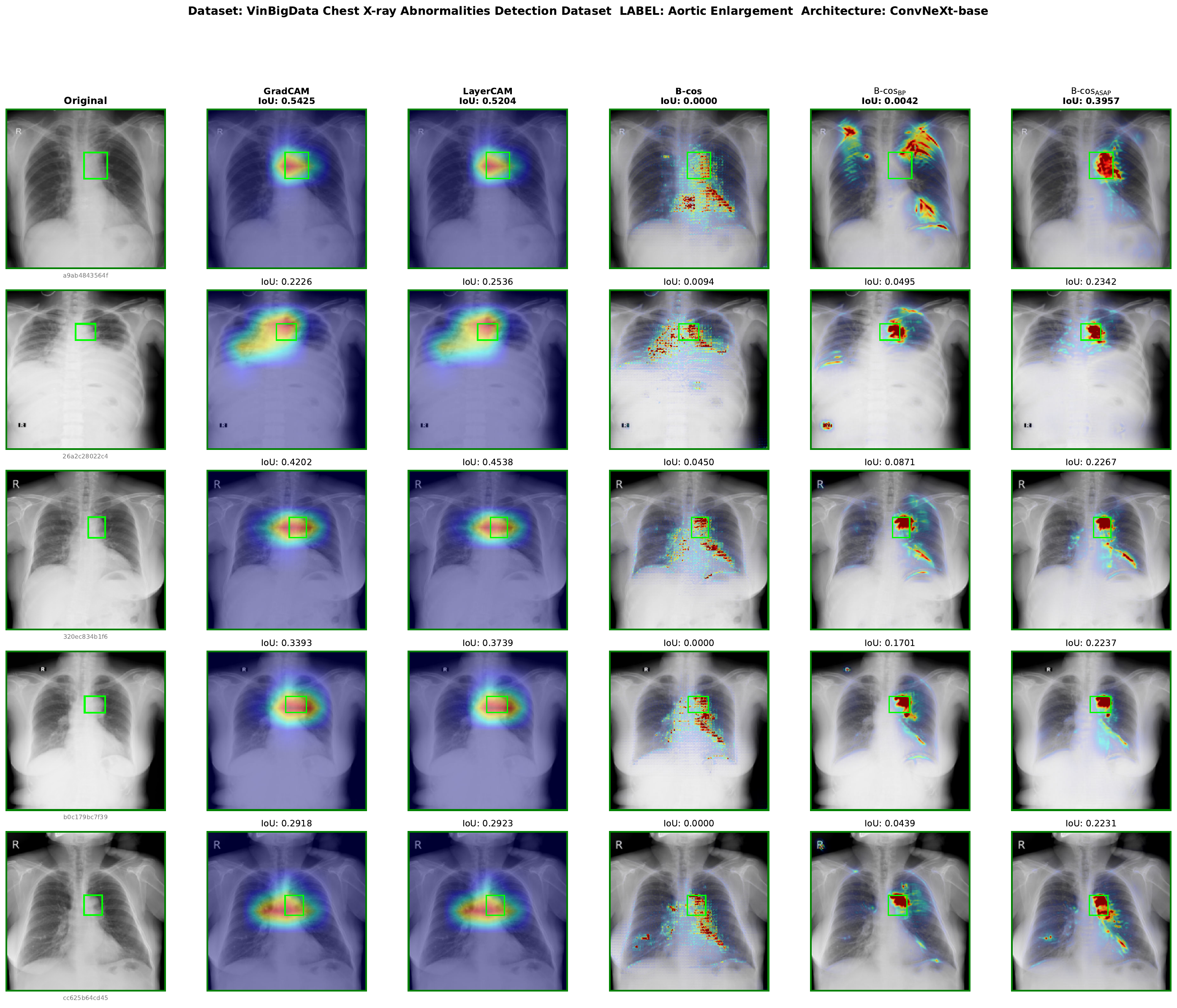}

\caption{VinBig: Aortic Enlargement, ConvNeXt-base. Five illustrative cases using light augmentation without sampling (fold 1, seed 420).}

\label{fig:final-17}

\end{figure}

\clearpage

\begin{figure}[!ht]

\centering

\includegraphics[width=\linewidth,height=0.79\textheight,keepaspectratio]{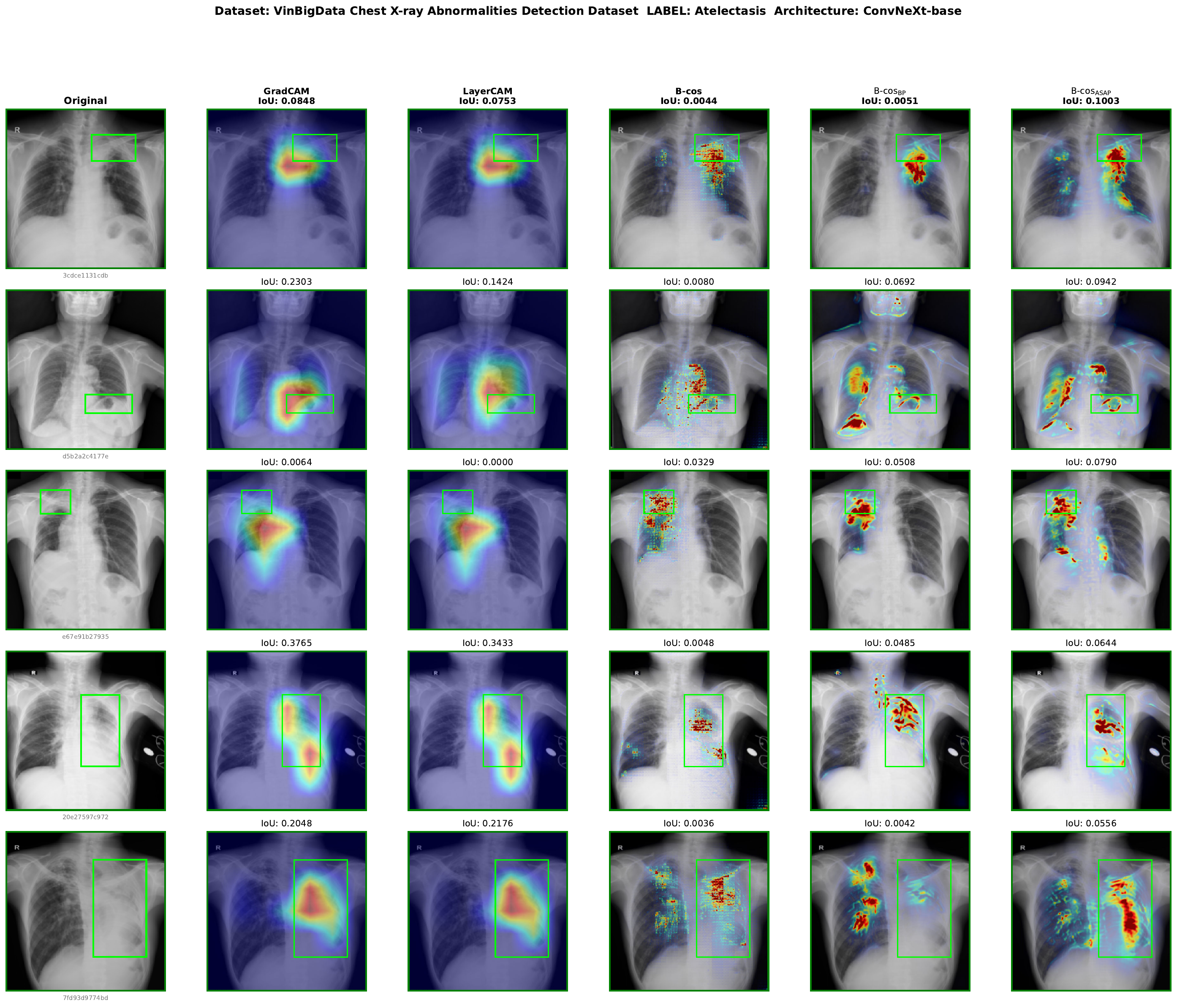}

\caption{VinBig: Atelectasis, ConvNeXt-base. Five illustrative cases using light augmentation without sampling (fold 1, seed 420).}

\label{fig:final-18}

\end{figure}

\clearpage

\begin{figure}[!ht]

\centering

\includegraphics[width=\linewidth,height=0.79\textheight,keepaspectratio]{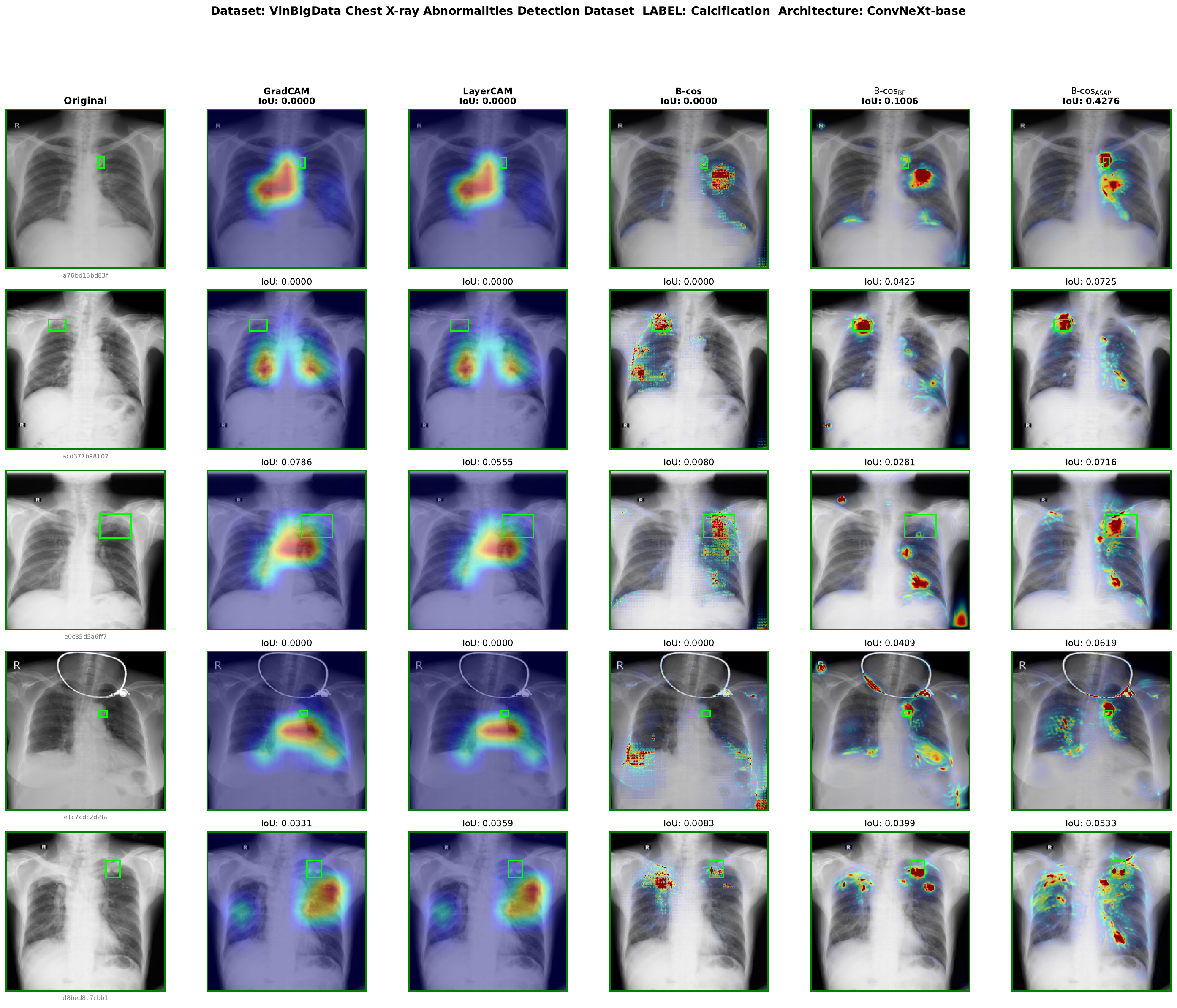}

\caption{VinBig: Calcification, ConvNeXt-base. Five illustrative cases using light augmentation without sampling (fold 1, seed 420).}

\label{fig:final-19}

\end{figure}

\clearpage

\begin{figure}[!ht]

\centering

\includegraphics[width=\linewidth,height=0.79\textheight,keepaspectratio]{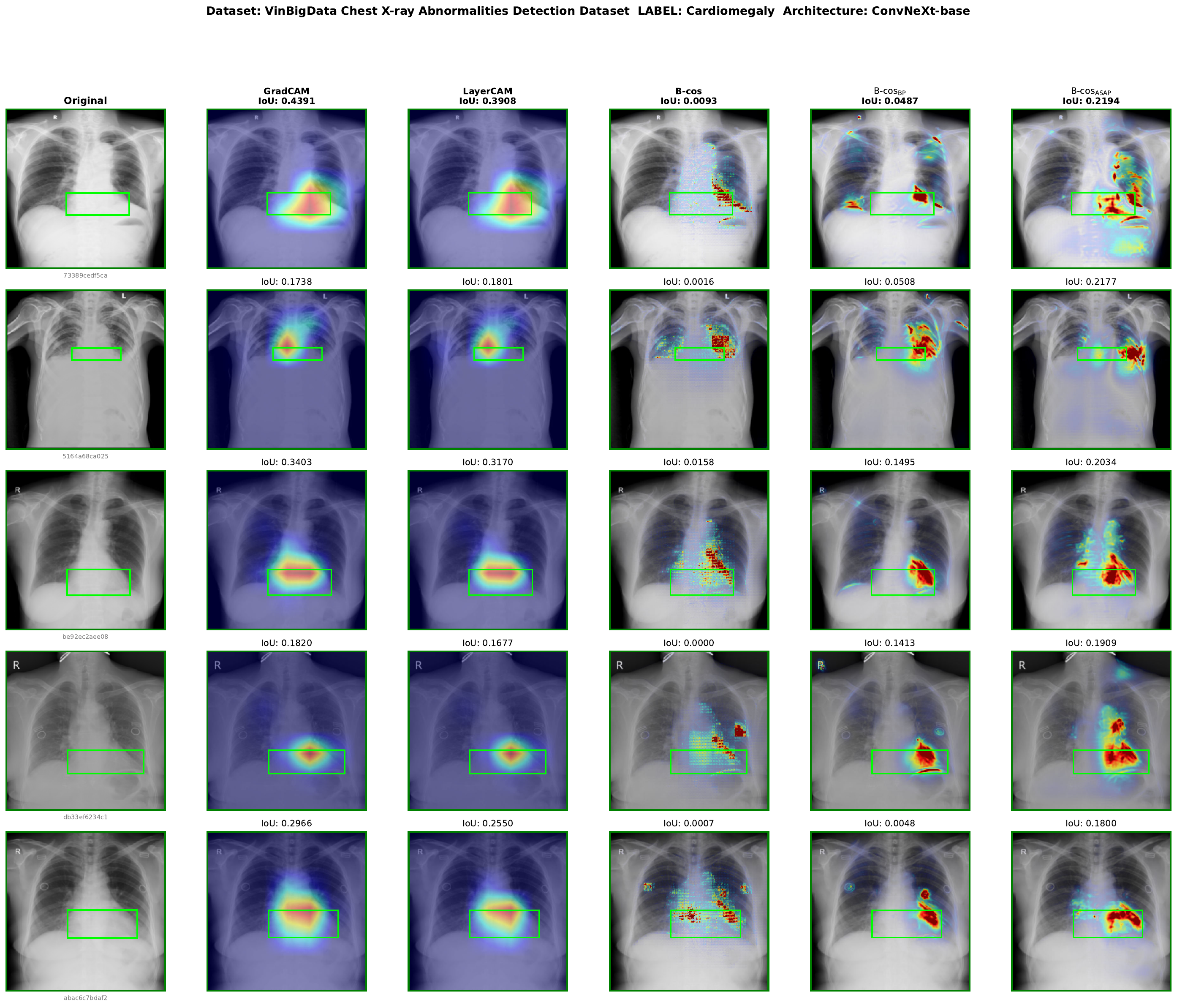}

\caption{VinBig: Cardiomegaly, ConvNeXt-base. Five illustrative cases using light augmentation without sampling (fold 1, seed 420).}

\label{fig:final-20}

\end{figure}

\clearpage

\begin{figure}[!ht]

\centering

\includegraphics[width=\linewidth,height=0.79\textheight,keepaspectratio]{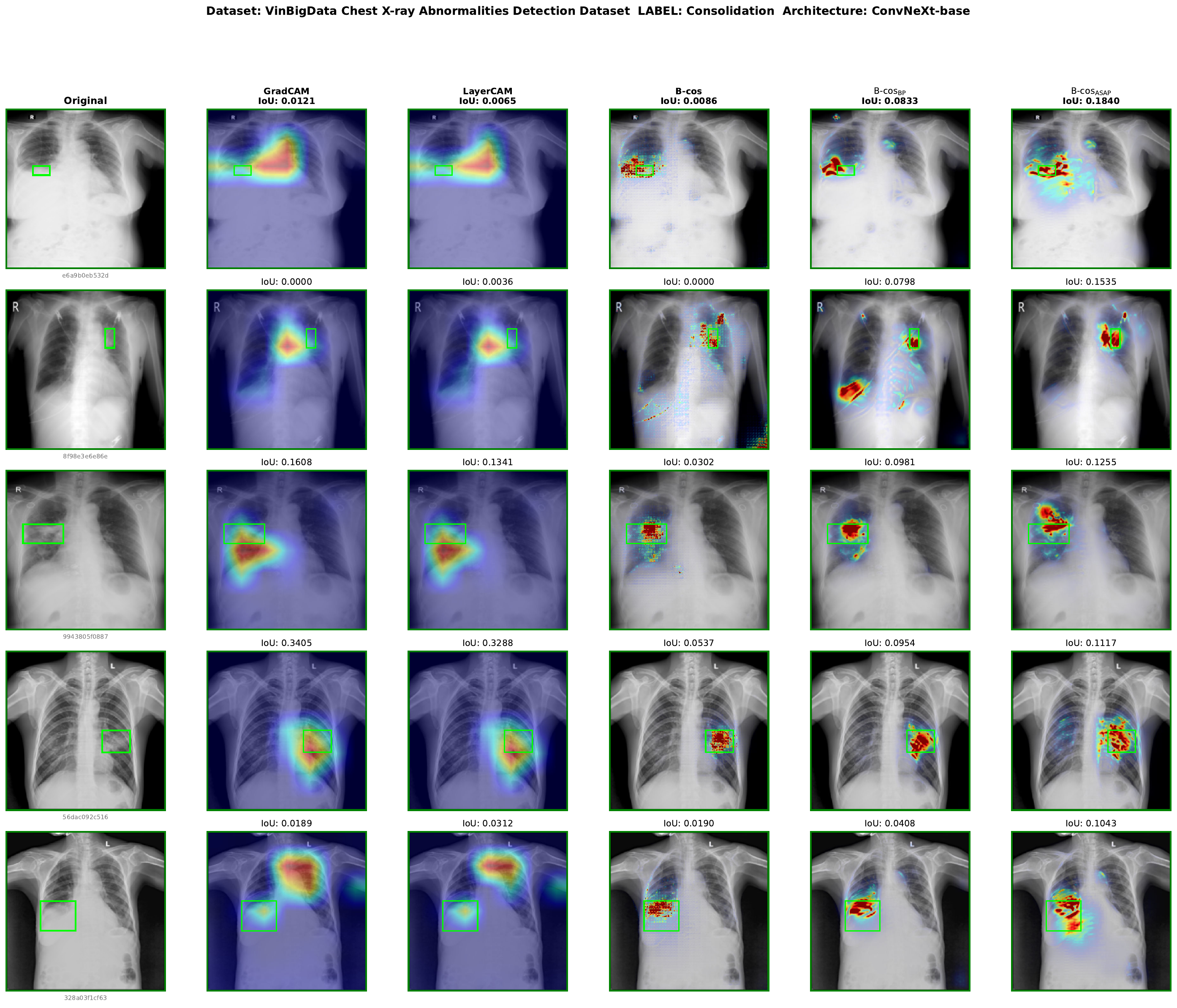}

\caption{VinBig: Consolidation, ConvNeXt-base. Five illustrative cases using light augmentation without sampling (fold 1, seed 420).}

\label{fig:final-21}

\end{figure}

\clearpage

\begin{figure}[!ht]

\centering

\includegraphics[width=\linewidth,height=0.79\textheight,keepaspectratio]{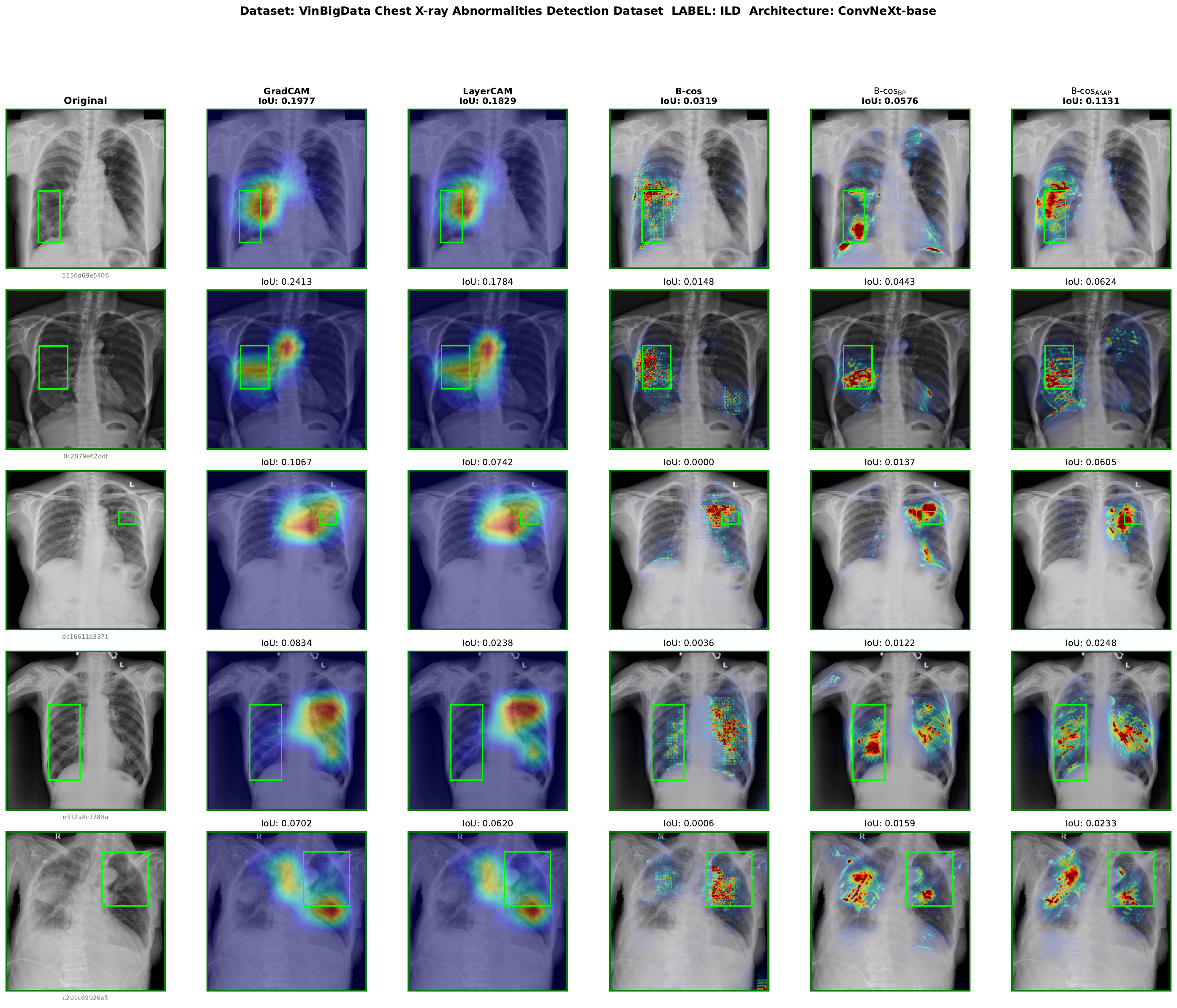}

\caption{VinBig: ILD, ConvNeXt-base. Five illustrative cases using light augmentation without sampling (fold 1, seed 420).}

\label{fig:final-22}

\end{figure}

\clearpage

\begin{figure}[!ht]

\centering

\includegraphics[width=\linewidth,height=0.79\textheight,keepaspectratio]{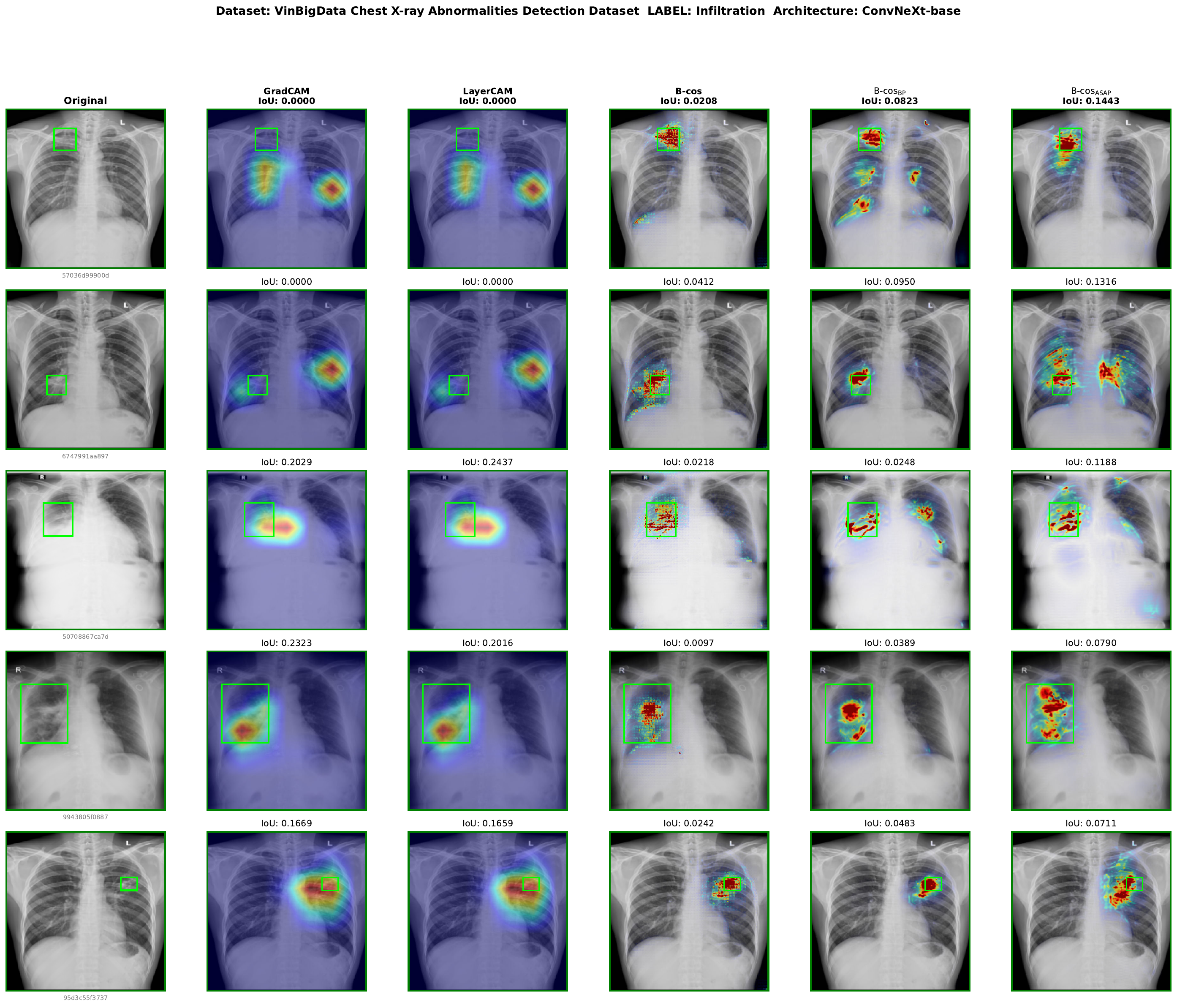}

\caption{VinBig: Infiltration, ConvNeXt-base. Five illustrative cases using light augmentation without sampling (fold 1, seed 420).}

\label{fig:final-23}

\end{figure}

\clearpage

\begin{figure}[!ht]

\centering

\includegraphics[width=\linewidth,height=0.79\textheight,keepaspectratio]{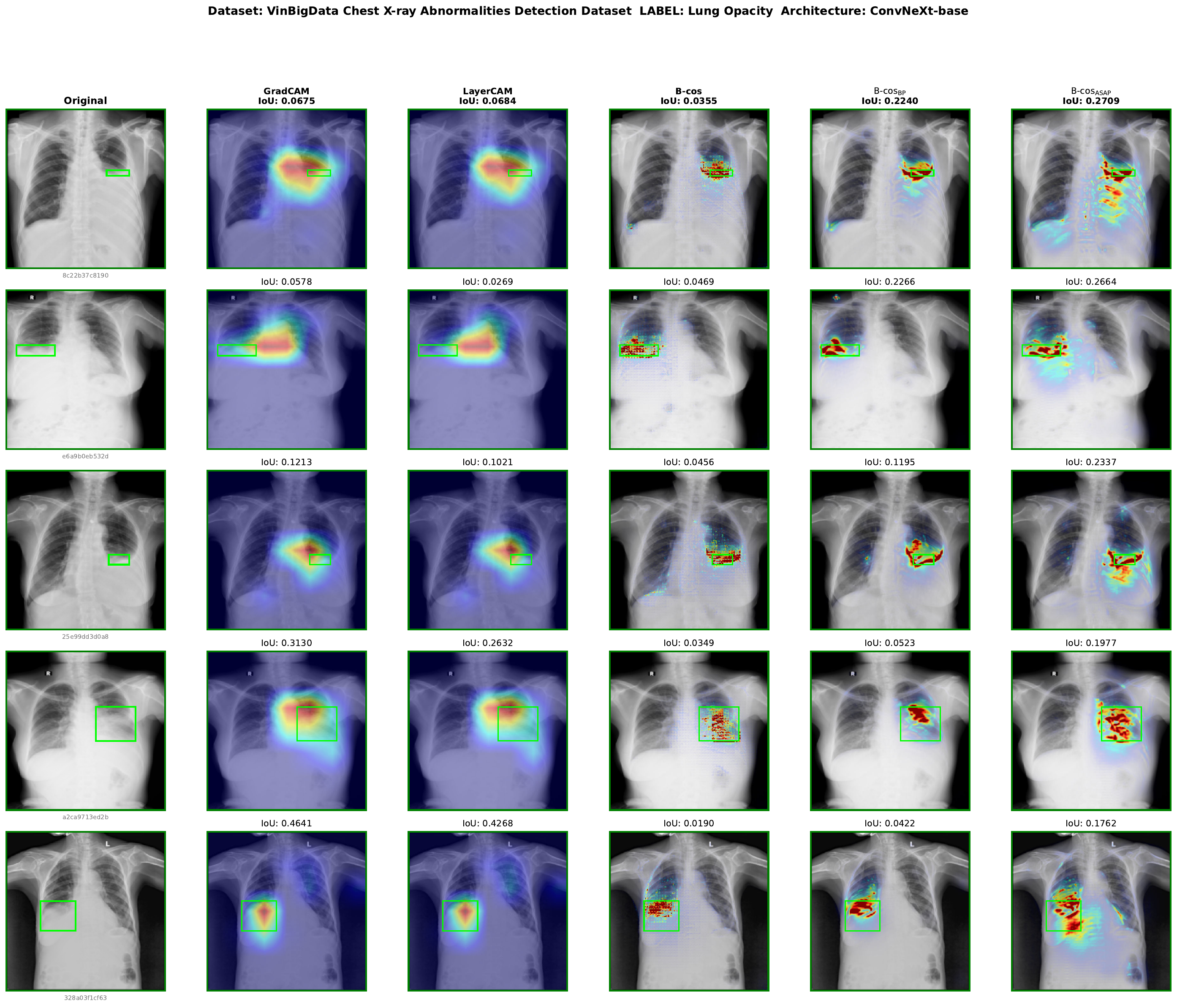}

\caption{VinBig: Lung Opacity, ConvNeXt-base. Five illustrative cases using light augmentation without sampling (fold 1, seed 420).}

\label{fig:final-24}

\end{figure}

\clearpage

\begin{figure}[!ht]

\centering

\includegraphics[width=\linewidth,height=0.79\textheight,keepaspectratio]{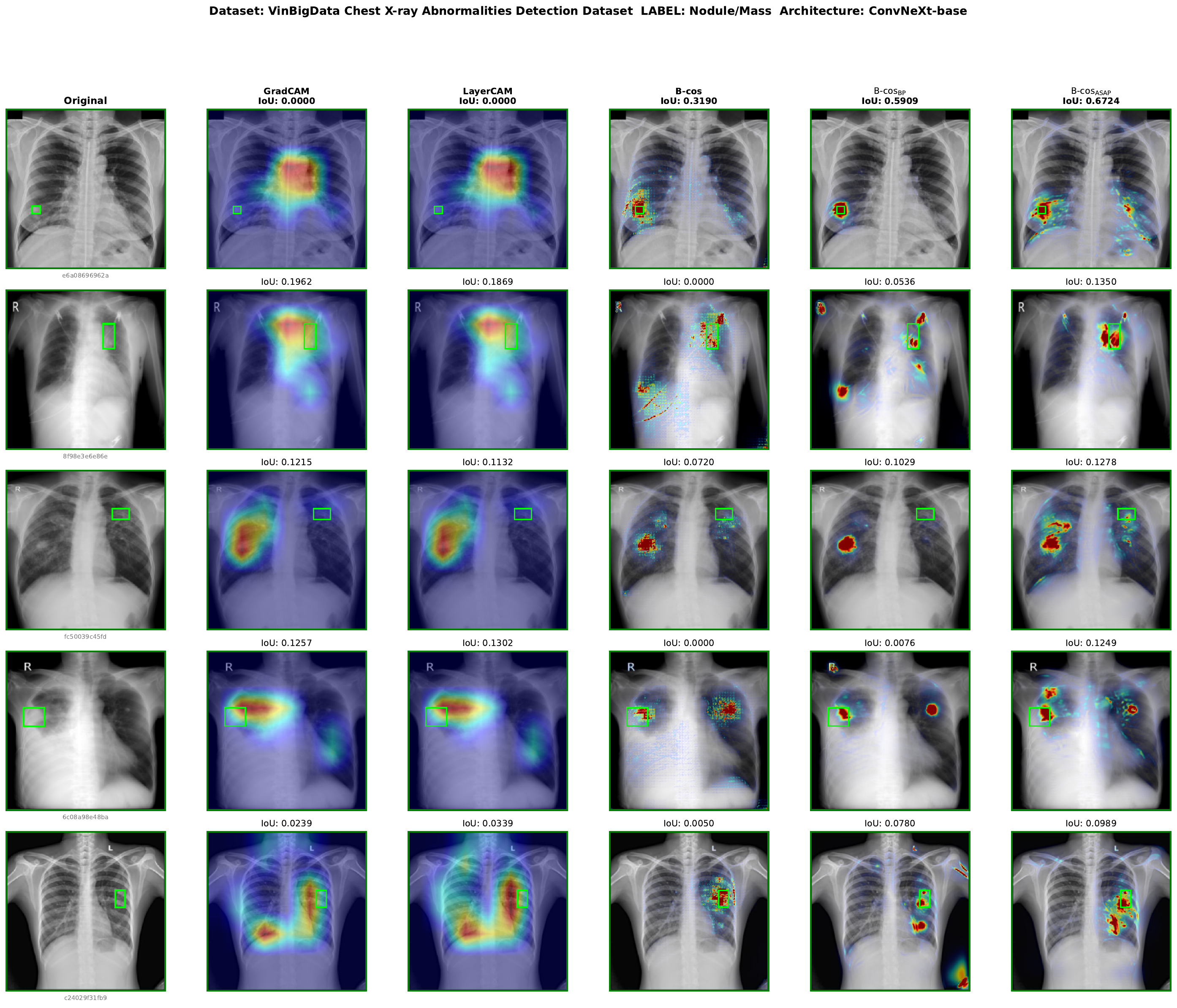}

\caption{VinBig: Nodule/Mass, ConvNeXt-base. Five illustrative cases using light augmentation without sampling (fold 1, seed 420).}

\label{fig:final-25}

\end{figure}

\clearpage

\begin{figure}[!ht]

\centering

\includegraphics[width=\linewidth,height=0.79\textheight,keepaspectratio]{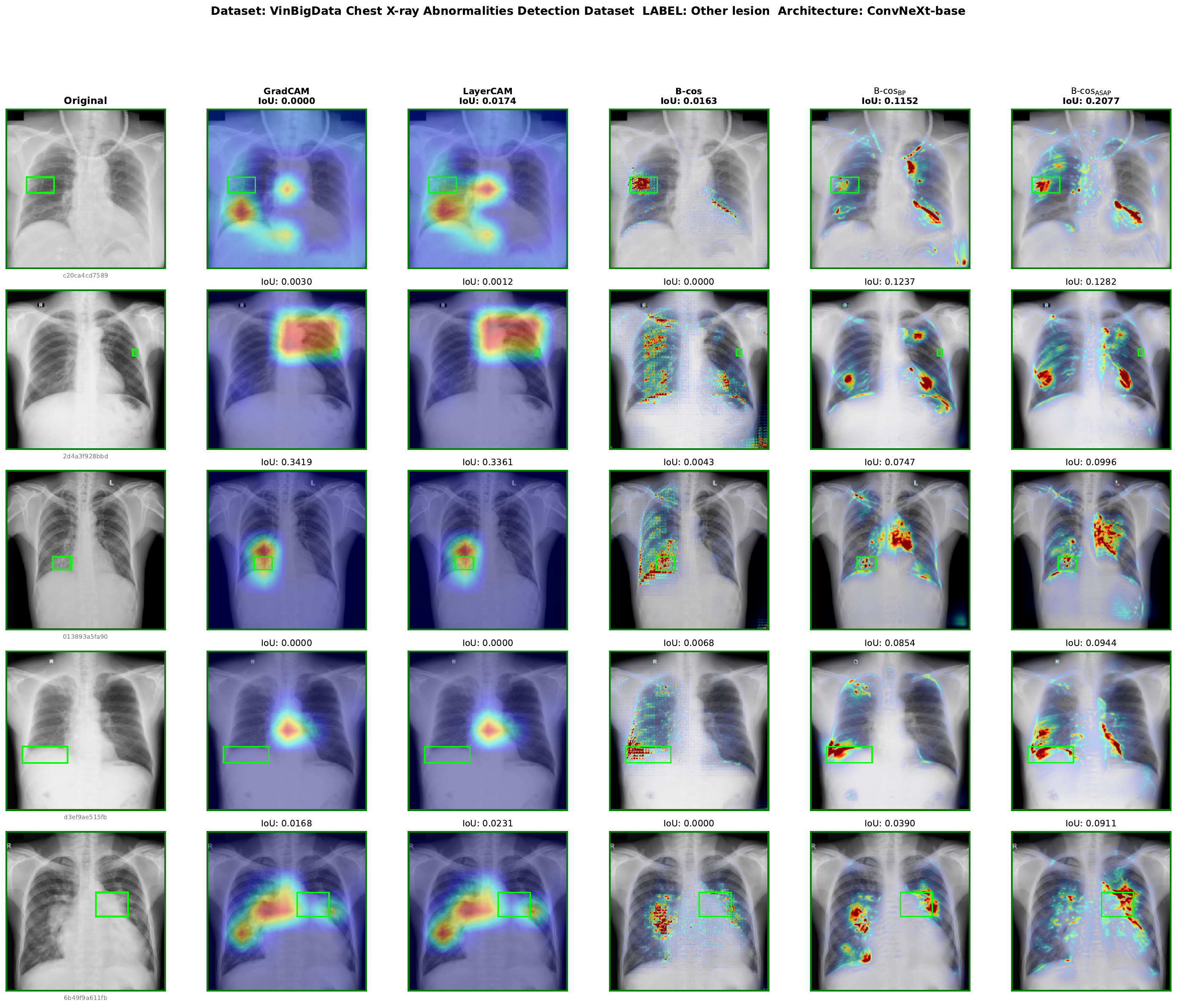}

\caption{VinBig: Other lesion, ConvNeXt-base. Five illustrative cases using light augmentation without sampling (fold 1, seed 420).}

\label{fig:final-26}

\end{figure}

\clearpage

\begin{figure}[!ht]

\centering

\includegraphics[width=\linewidth,height=0.79\textheight,keepaspectratio]{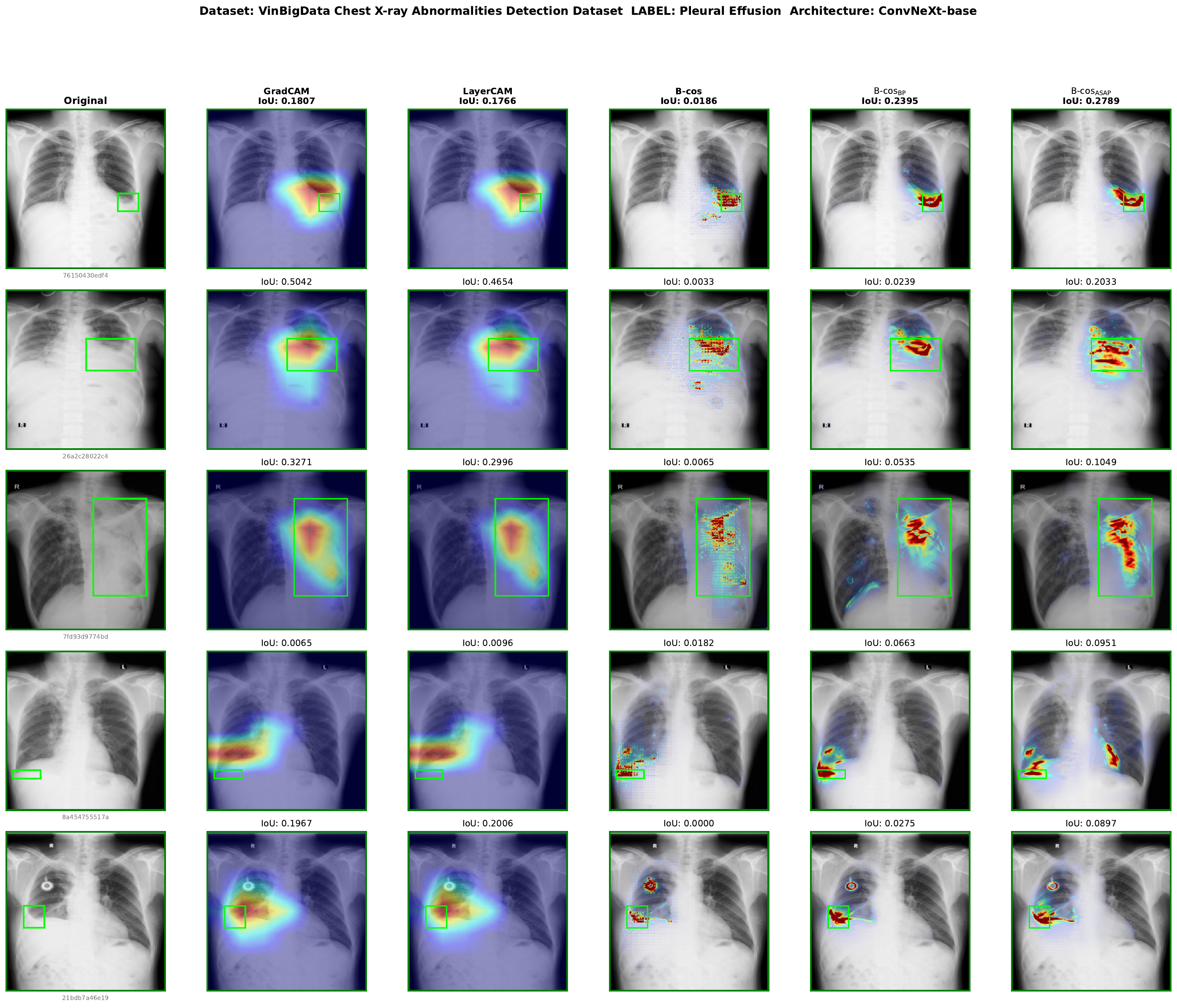}

\caption{VinBig: Pleural Effusion, ConvNeXt-base. Five illustrative cases using light augmentation without sampling (fold 1, seed 420).}

\label{fig:final-27}

\end{figure}

\clearpage

\begin{figure}[!ht]

\centering

\includegraphics[width=\linewidth,height=0.79\textheight,keepaspectratio]{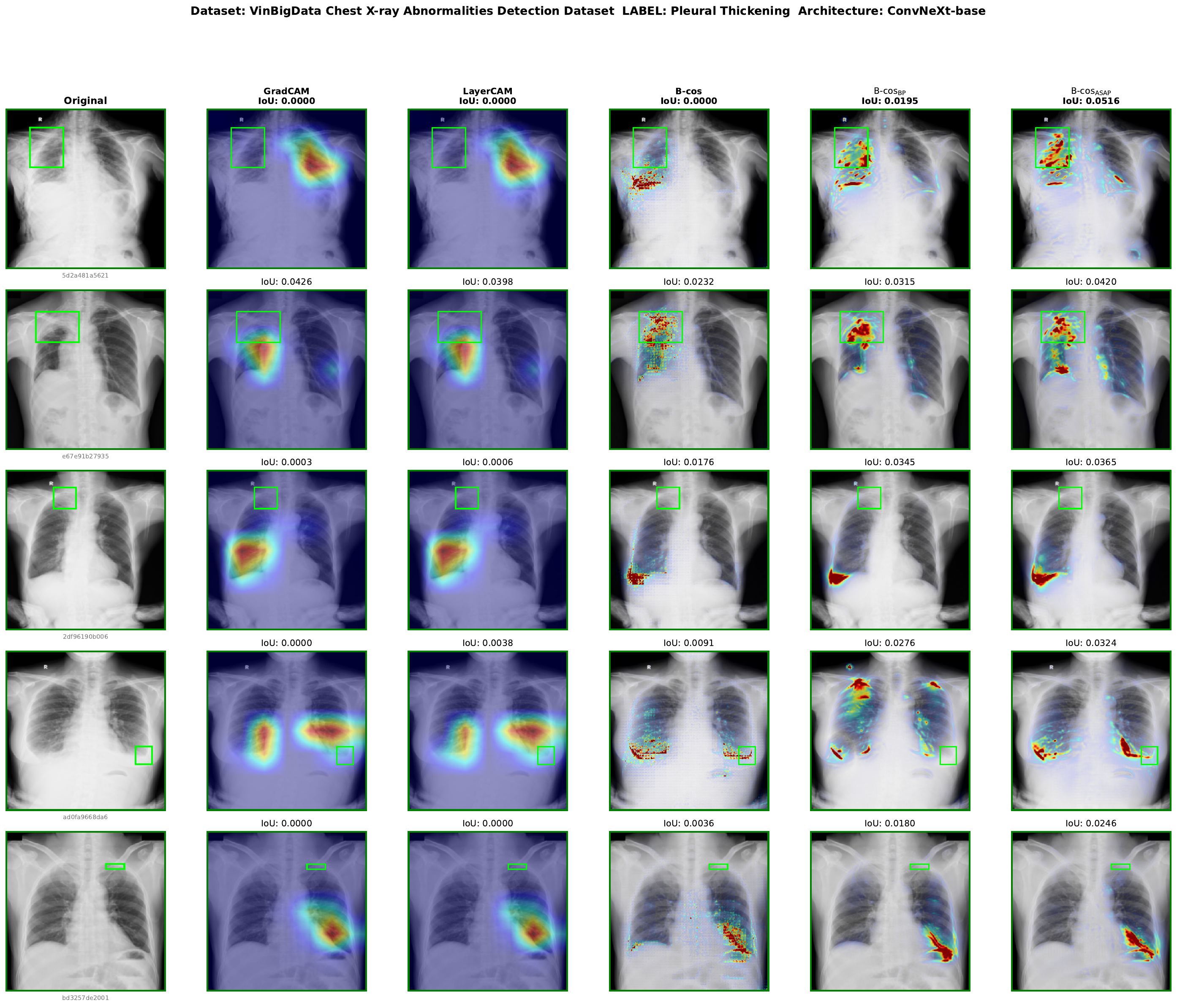}

\caption{VinBig: Pleural Thickening, ConvNeXt-base. Five illustrative cases using light augmentation without sampling (fold 1, seed 420).}

\label{fig:final-28}

\end{figure}

\clearpage

\begin{figure}[!ht]

\centering

\includegraphics[width=\linewidth,height=0.79\textheight,keepaspectratio]{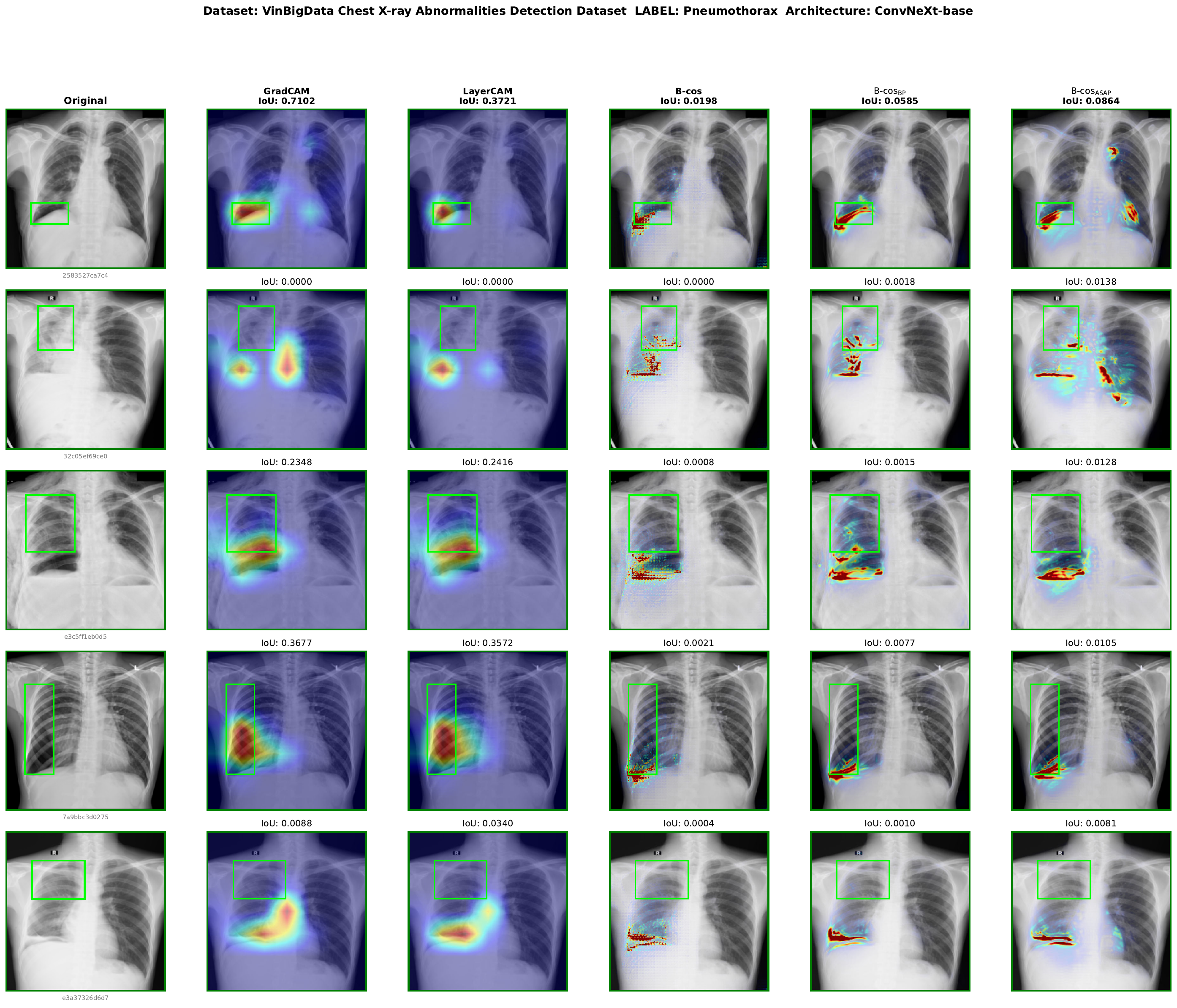}

\caption{VinBig: Pneumothorax, ConvNeXt-base. Five illustrative cases using light augmentation without sampling (fold 1, seed 420).}

\label{fig:final-29}

\end{figure}

\clearpage

\begin{figure}[!ht]

\centering

\includegraphics[width=\linewidth,height=0.79\textheight,keepaspectratio]{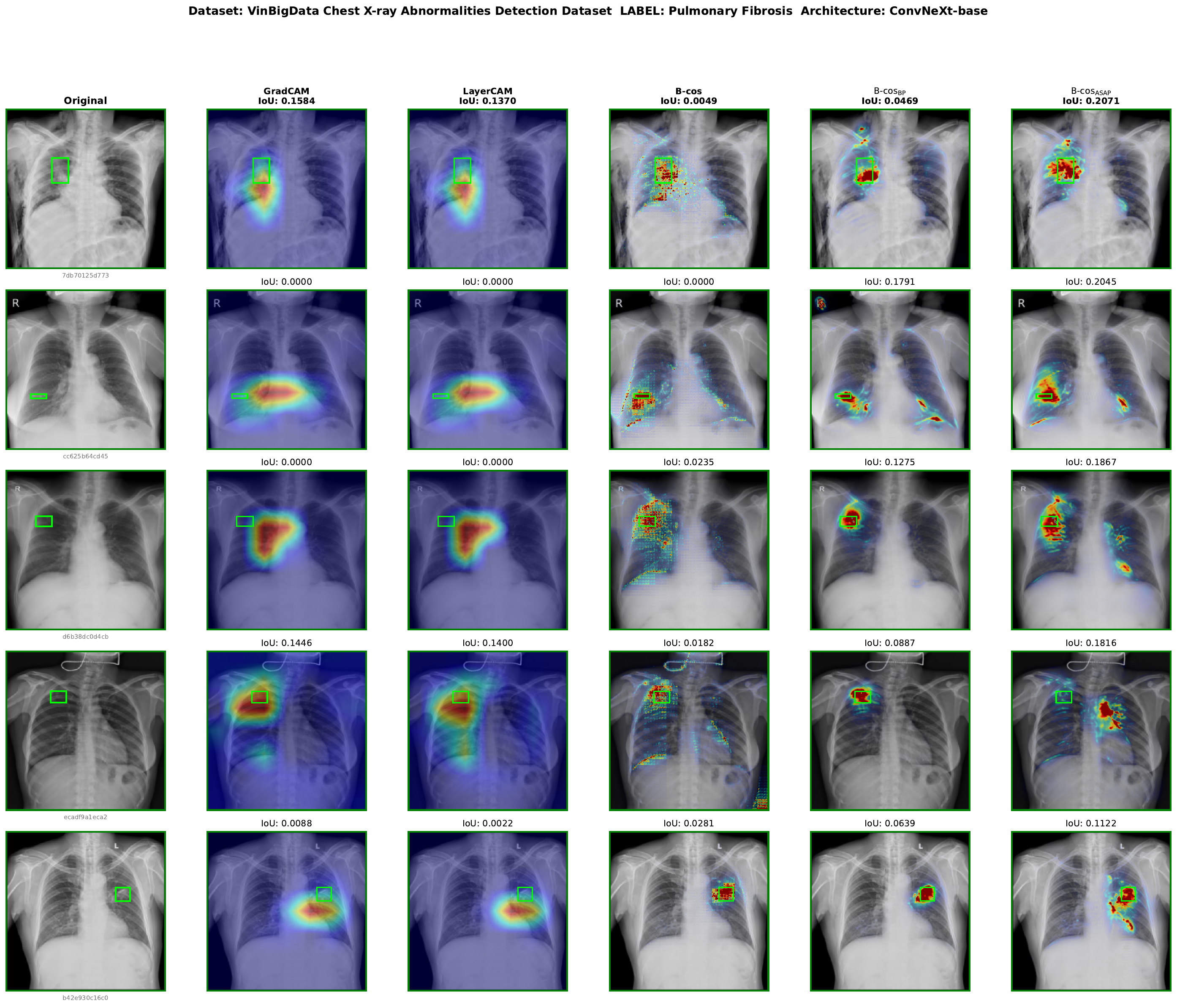}

\caption{VinBig: Pulmonary Fibrosis, ConvNeXt-base. Five illustrative cases using light augmentation without sampling (fold 1, seed 420).}

\label{fig:final-30}

\end{figure}

\clearpage

\subsection{VinBig: ConvNeXt-tiny}

\begin{figure}[!ht]

\centering

\includegraphics[width=\linewidth,height=0.79\textheight,keepaspectratio]{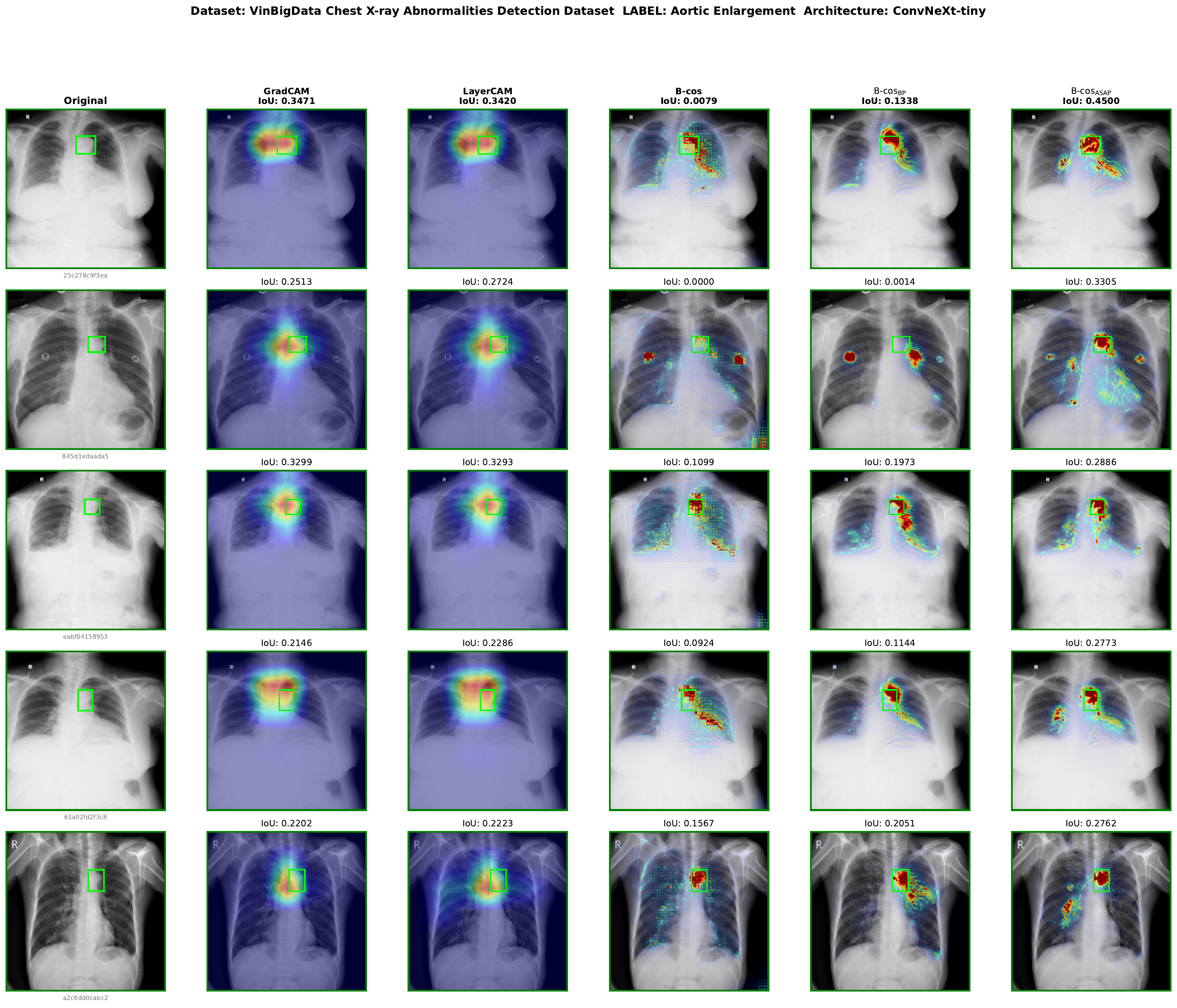}

\caption{VinBig: Aortic Enlargement, ConvNeXt-tiny. Five illustrative cases using light augmentation without sampling (fold 3, seed 1337).}

\label{fig:final-31}

\end{figure}

\clearpage

\begin{figure}[!ht]

\centering

\includegraphics[width=\linewidth,height=0.79\textheight,keepaspectratio]{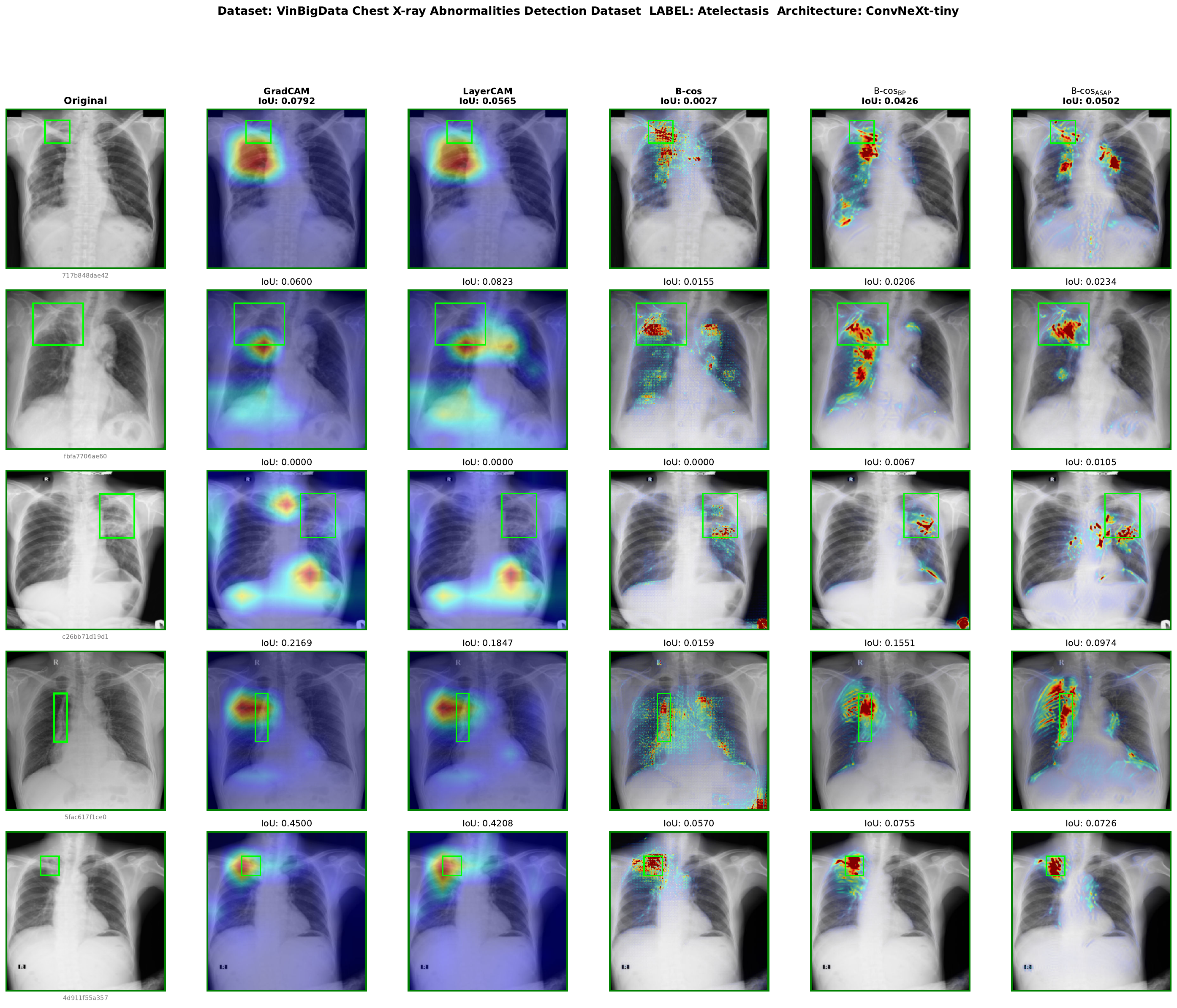}

\caption{VinBig: Atelectasis, ConvNeXt-tiny. Five illustrative cases using light augmentation without sampling (fold 3, seed 1337).}

\label{fig:final-32}

\end{figure}

\clearpage

\begin{figure}[!ht]

\centering

\includegraphics[width=\linewidth,height=0.79\textheight,keepaspectratio]{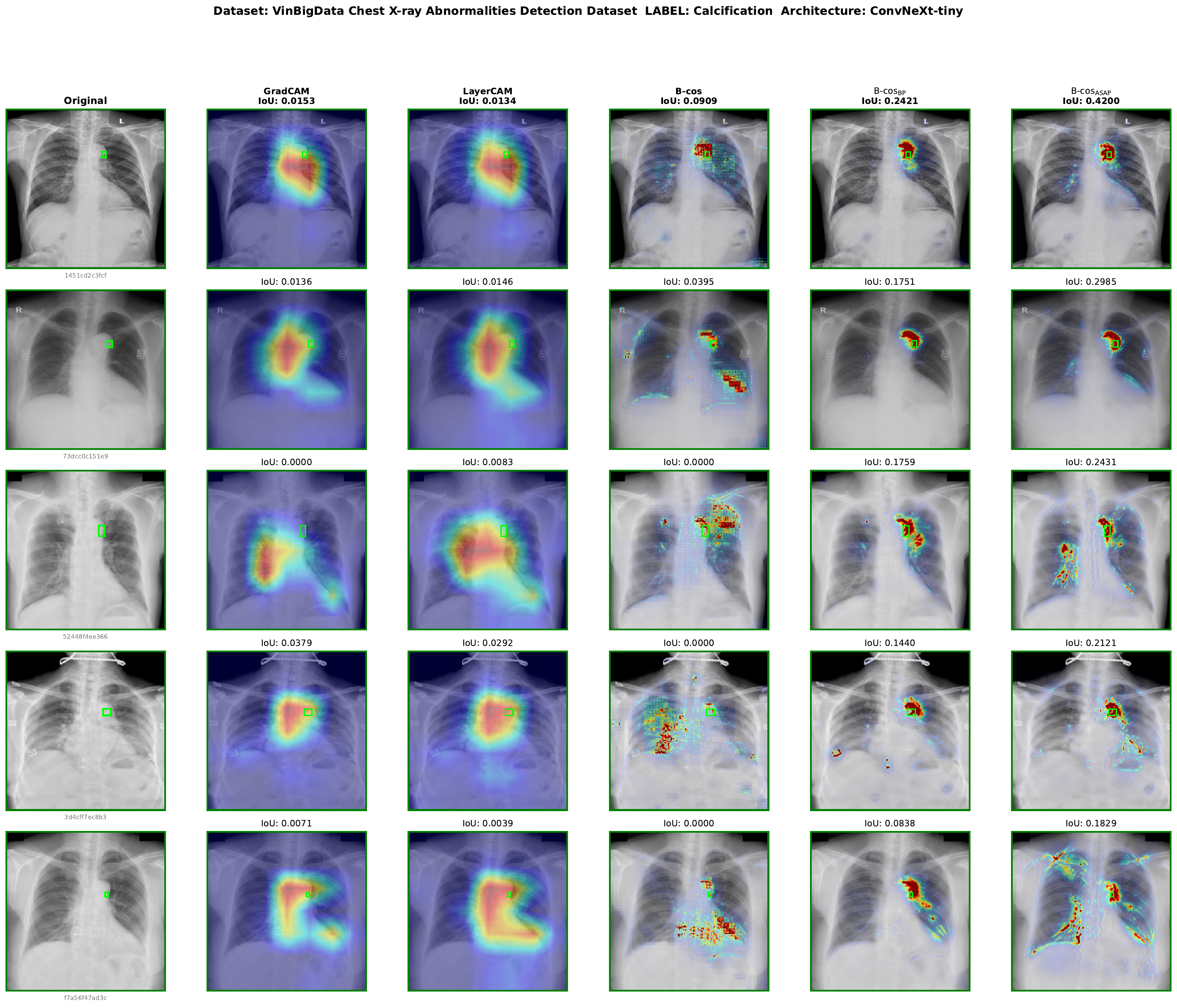}

\caption{VinBig: Calcification, ConvNeXt-tiny. Five illustrative cases using light augmentation without sampling (fold 3, seed 1337).}

\label{fig:final-33}

\end{figure}

\clearpage

\begin{figure}[!ht]

\centering

\includegraphics[width=\linewidth,height=0.79\textheight,keepaspectratio]{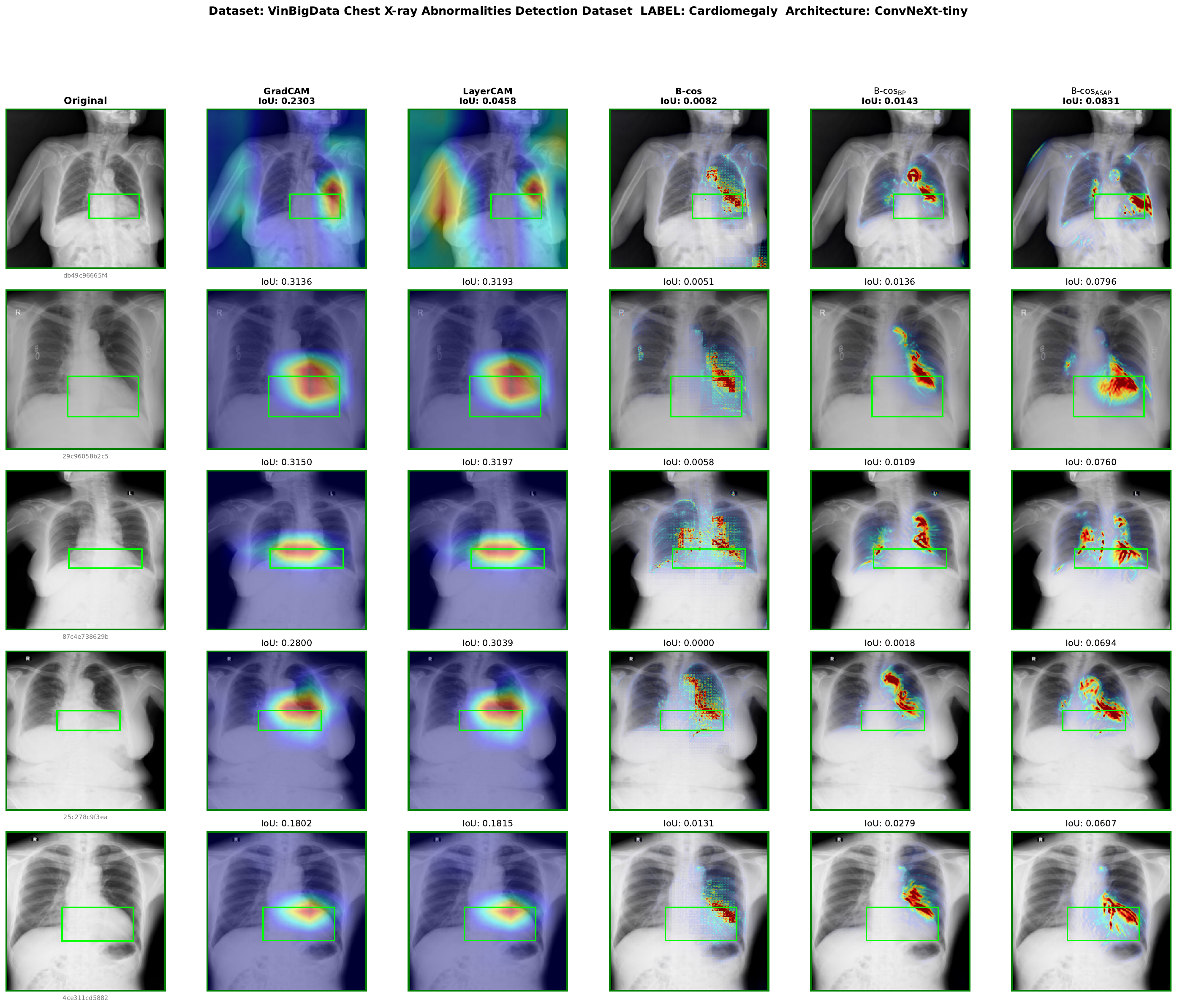}

\caption{VinBig: Cardiomegaly, ConvNeXt-tiny. Five illustrative cases using light augmentation without sampling (fold 3, seed 1337).}

\label{fig:final-34}

\end{figure}

\clearpage

\begin{figure}[!ht]

\centering

\includegraphics[width=\linewidth,height=0.79\textheight,keepaspectratio]{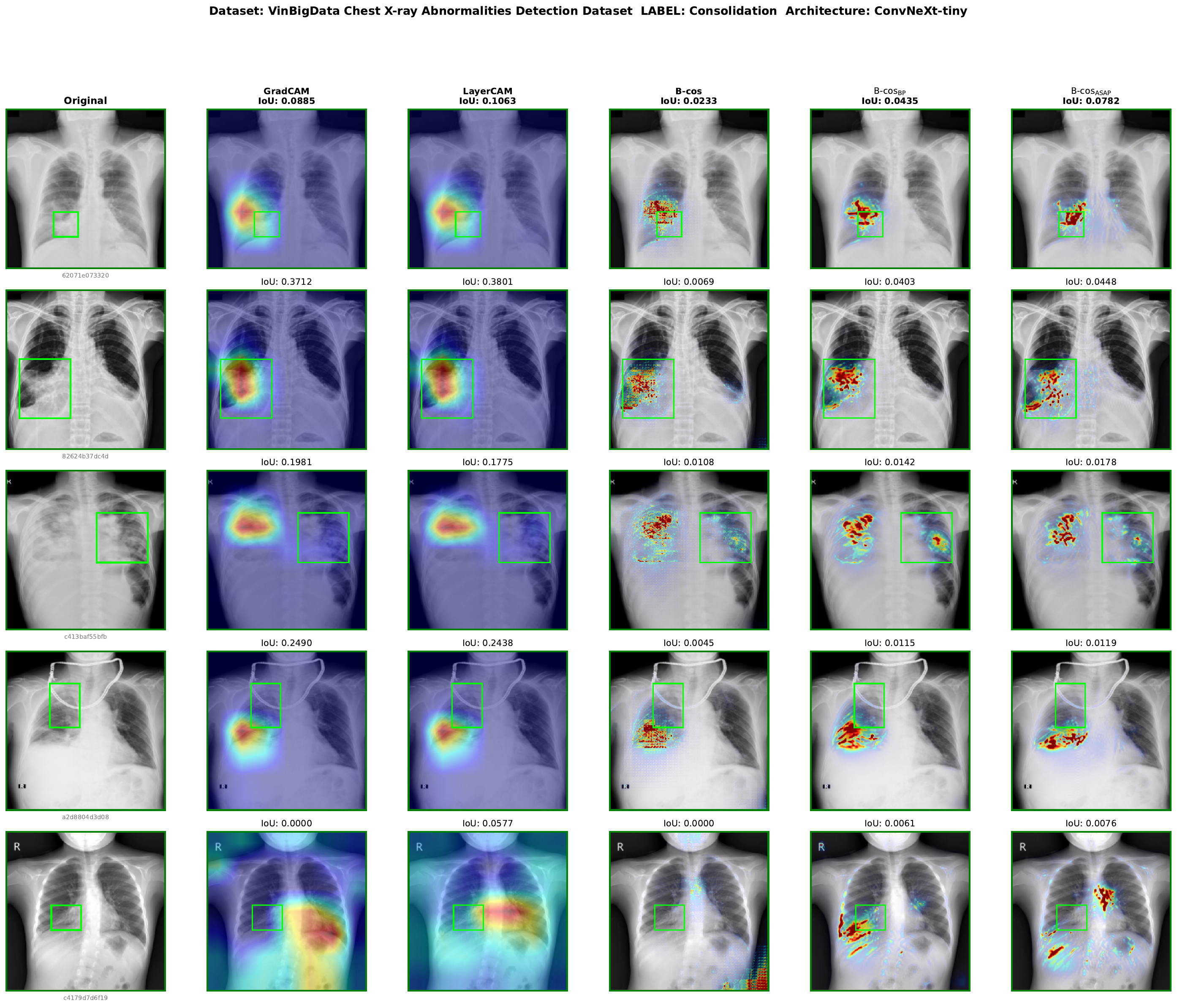}

\caption{VinBig: Consolidation, ConvNeXt-tiny. Five illustrative cases using light augmentation without sampling (fold 3, seed 1337).}

\label{fig:final-35}

\end{figure}

\clearpage

\begin{figure}[!ht]

\centering

\includegraphics[width=\linewidth,height=0.79\textheight,keepaspectratio]{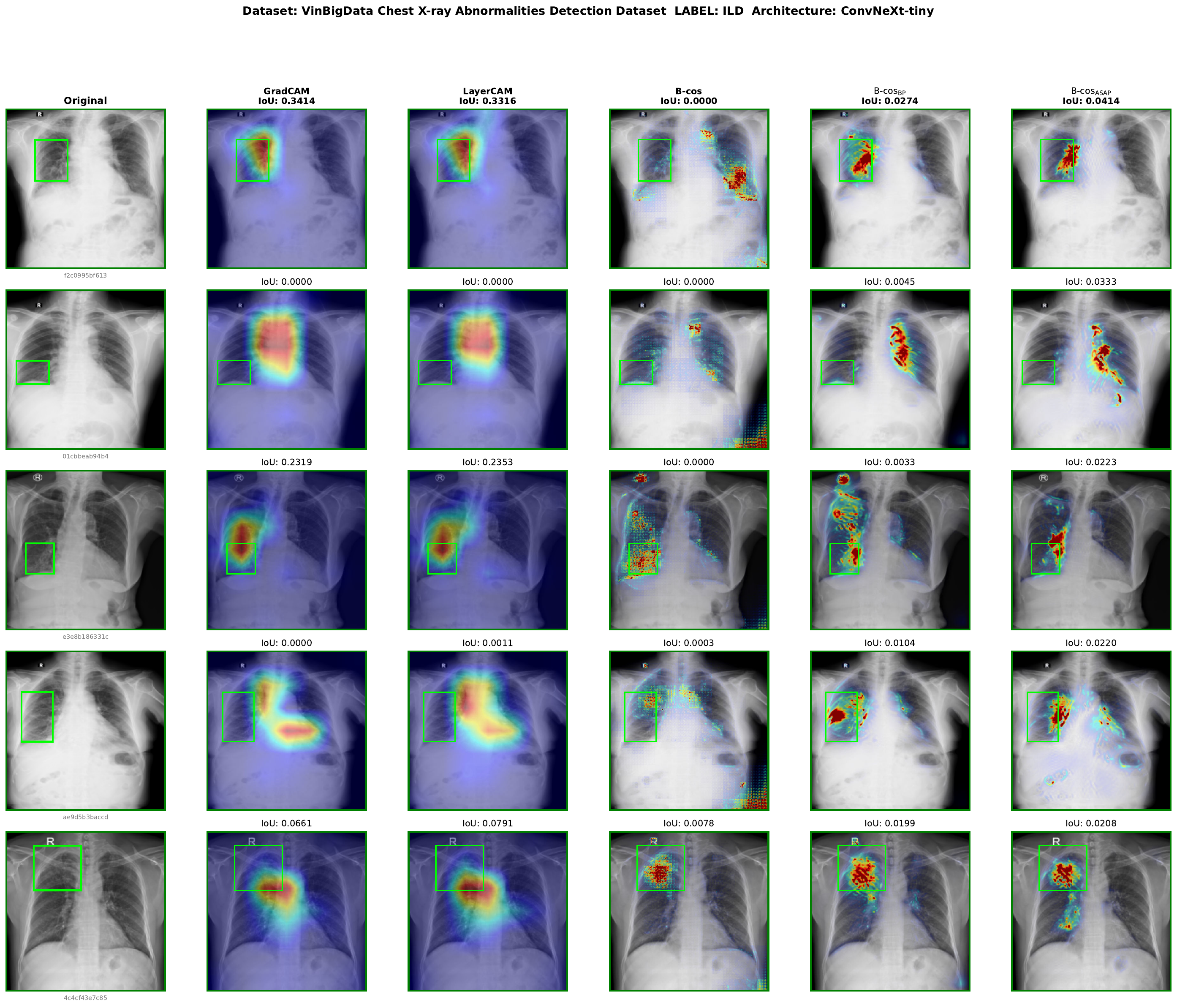}

\caption{VinBig: ILD, ConvNeXt-tiny. Five illustrative cases using light augmentation without sampling (fold 3, seed 1337).}

\label{fig:final-36}

\end{figure}

\clearpage

\begin{figure}[!ht]

\centering

\includegraphics[width=\linewidth,height=0.79\textheight,keepaspectratio]{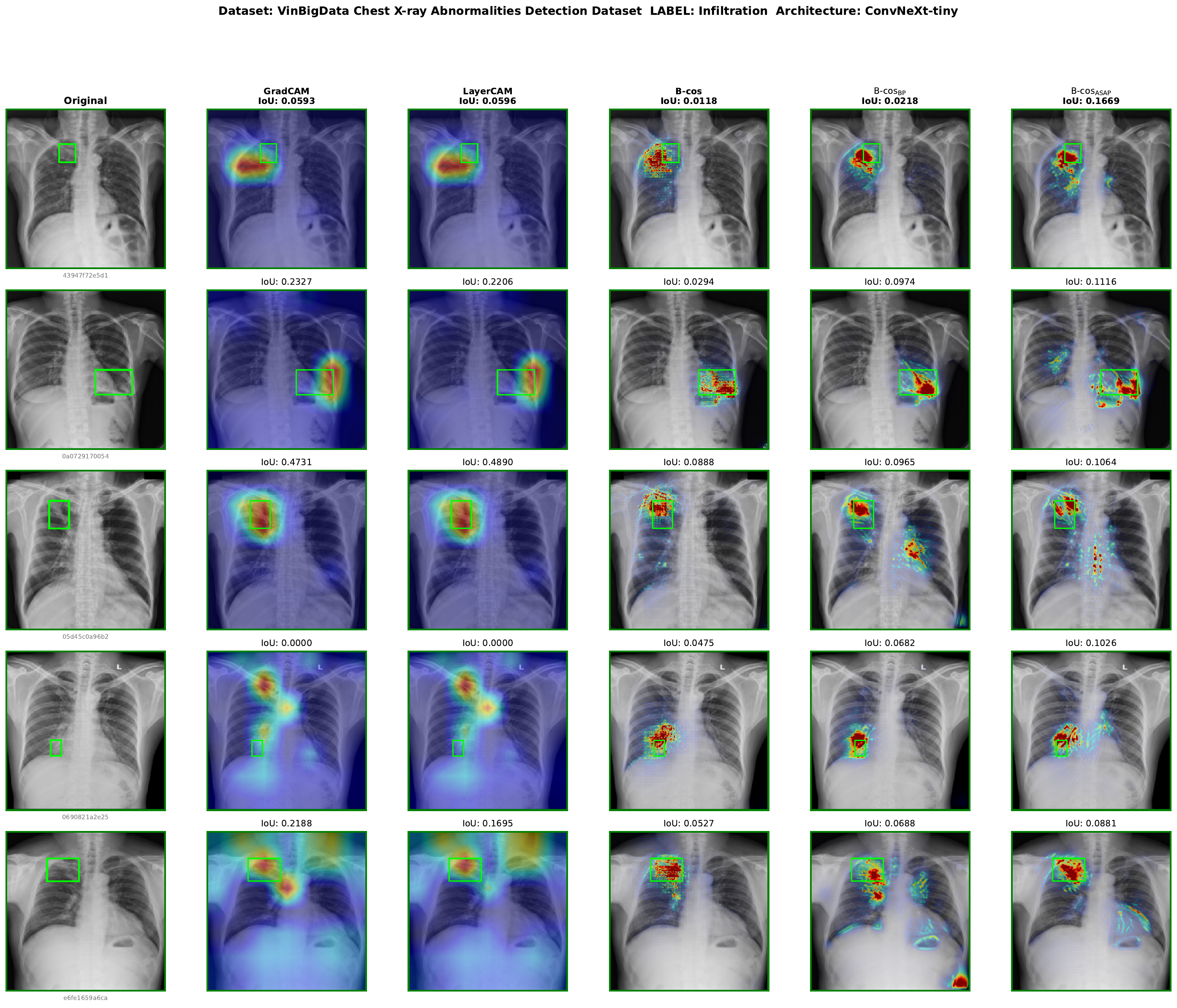}

\caption{VinBig: Infiltration, ConvNeXt-tiny. Five illustrative cases using light augmentation without sampling (fold 3, seed 1337).}

\label{fig:final-37}

\end{figure}

\clearpage

\begin{figure}[!ht]

\centering

\includegraphics[width=\linewidth,height=0.79\textheight,keepaspectratio]{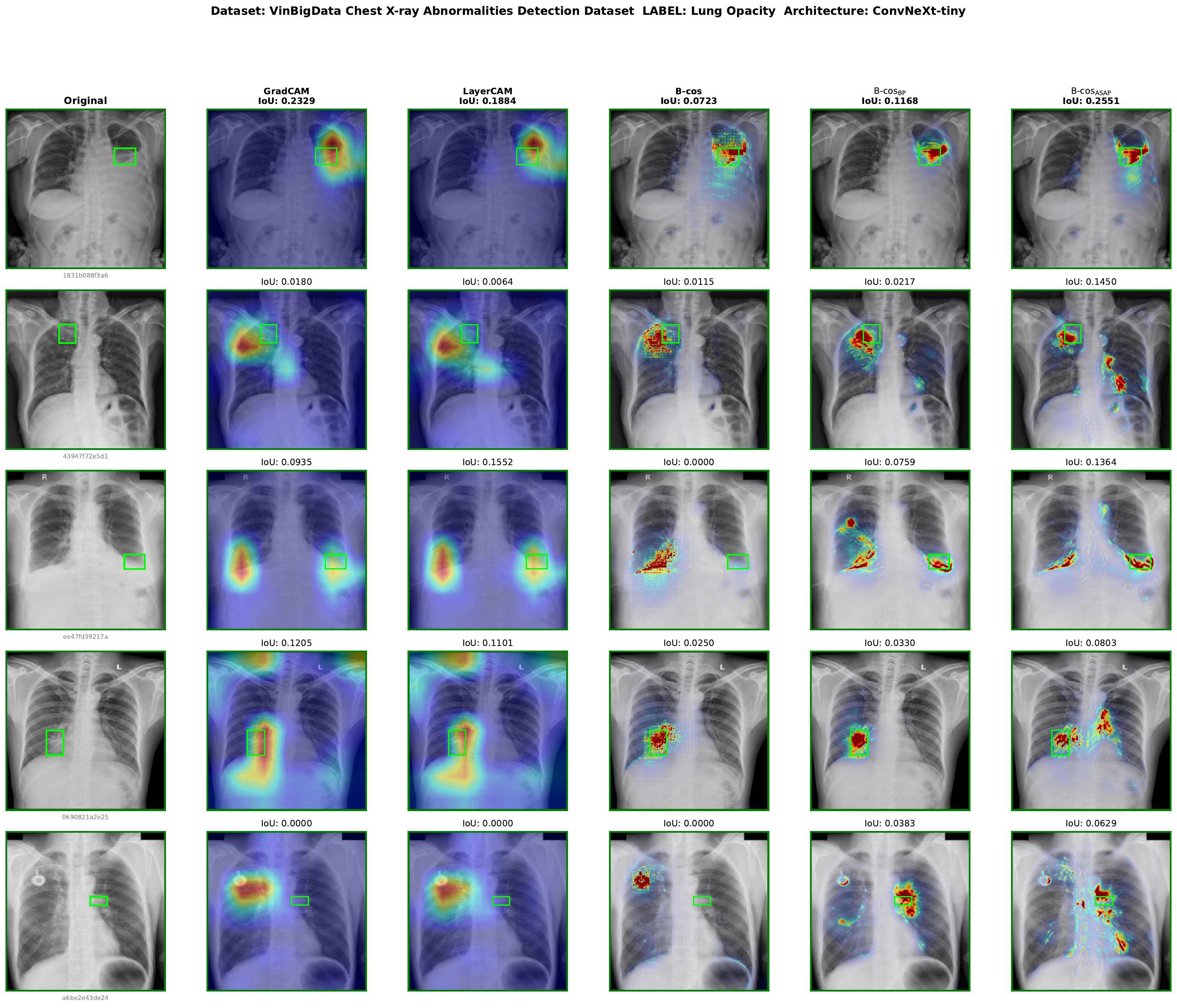}

\caption{VinBig: Lung Opacity, ConvNeXt-tiny. Five illustrative cases using light augmentation without sampling (fold 3, seed 1337).}

\label{fig:final-38}

\end{figure}

\end{document}